\documentclass[9pt,twocolumn,letterpaper]{article}
\usepackage{graphicx,epsfig,fullpage,svg}
\usepackage[hyphens]{url}
\usepackage{caption}
\usepackage{subfig}
\usepackage{float}
\usepackage{comment}
\usepackage{caption}
\usepackage[utf8x]{inputenc}
\usepackage[T1]{fontenc}
\usepackage{mathtools} 
\usepackage{adjustbox}
\usepackage{tabularx}
\usepackage{multirow}
\usepackage{booktabs} 
\usepackage{hyperref}

\title{TRECVID 2019: An evaluation campaign to benchmark Video Activity Detection, Video Captioning and Matching, and Video Search \& retrieval}
\author{
George Awad
  \{gawad@nist.gov\}\\
  Georgetown University; NIST, USA\\\\
Asad A. Butt
  \{asad.butt@nist.gov\}\\
  Johns Hopkins University; NIST, USA\\\\
Keith Curtis
  \{keith.curtis@nist.gov\}\\
  Guest Researcher, NIST, USA\\\\
Yooyoung Lee
  \{yooyoung@nist.gov\}\
Jonathan Fiscus
  \{jfiscus@nist.gov\}\\
  Afzal Godil
   \{godil@nist.gov\}\
Andrew Delgado
  \{andrew.delgado@nist.gov\}\\
Jesse Zhang
  \{jesse.zhang@nist.gov\}\\
Information Access Division\\
National Institute of Standards and Technology\\
Gaithersburg, MD 20899-8940, USA\\\\
Eliot Godard
  \{eliot.godard@nist.gov\}\\
  Guest Researcher, NIST, USA\\\\
Lukas Diduch
  \{lukas.diduch@nist.gov\}\\
  Dakota-consulting,  USA\\\\
Alan F. Smeaton
  \{alan.smeaton@dcu.ie\}\\
 Insight Research Centre, Dublin City University, Glasnevin, Dublin 9, Ireland\\\\
Yvette Graham
  \{graham.yvette@gmail.com\}\\
 ADAPT Research Centre, Dublin City University, Glasnevin, Dublin 9, Ireland\\\\
Wessel Kraaij
  \{w.kraaij@liacs.leidenuniv.nl\}\\
  Leiden University; TNO, Netherlands\\\\
Georges Qu\'{e}not
  \{Georges.Quenot@imag.fr\}\\
  Laboratoire d'Informatique de Grenoble, France\\\\
}

\newlength{\spacelen}
\begin{document}
\graphicspath{{tv18.figures/}}
\maketitle

\section{Introduction}
The TREC Video Retrieval Evaluation (TRECVID) 2019 was a TREC-style
video analysis and retrieval evaluation, the goal of which remains to
promote progress in research and development of content-based exploitation and retrieval of 
information from digital video via open, metrics-based evaluation. 

Over the last nineteen years this effort has yielded a better understanding of how systems can effectively
accomplish such processing and how one can reliably benchmark their
performance. TRECVID has been funded by NIST (National Institute of Standards and Technology) and other US government agencies. 
In addition, many organizations and individuals worldwide contribute significant time and effort.

TRECVID 2019 represented a continuation of four tasks from TRECVID 2018. In total, 27 teams (see Table \ref{participants}) from various research organizations worldwide completed one or more of the following four tasks:

\begin{enumerate} \itemsep0pt \parskip0pt
\item Ad-hoc Video Search (AVS)
\item Instance Search (INS)
\item Activities in Extended Video (ActEV)
\item Video to Text Description (VTT)
\end{enumerate}

Table \ref{nonparticipants} represents organizations that registered but did not submit any runs.

This year TRECVID used a new Vimeo Creative Commons collection dataset (V3C1) \cite{rossetto2019v3c} of about 1000 hours in total and segmented into 1 million short video shots. The dataset is drawn from the Vimeo video sharing website under the Creative Common licenses and reflects a wide variety of content, style, and source device determined only by the self-selected donors. 

The Instance Search task used again the 464 hours of the BBC (British Broadcasting Corporation) EastEnders video as used before since 2013, while the Video to Text description task used a combination of 1044 Twitter social media Vine videos collected through the online Twitter API public stream and another 1010 short Flickr videos.

For the Activities in Extended Video task, about 10 hours of the VIRAT (Video and Image Retrieval and Analysis Tool) dataset was used which was designed to be realistic, natural and challenging for video surveillance domains in terms of its resolution, background clutter, diversity in scenes, and human activity/event categories.

The Ad-hoc search, Instance Search results were judged by NIST human assessors, while the Video to Text task was annotated by NIST human assessors and scored automatically later on using Machine Translation (MT) metrics and Direct Assessment (DA) by Amazon Mechanical Turk workers on sampled runs.

The systems submitted for the ActEV (Activities in Extended Video) evaluations were scored by NIST using reference annotations created by Kitware, Inc.

\begin{table*} 
\caption{Participants and tasks}
\label{participants}
  \vspace{1.0cm}
  \centering{
   \scriptsize{
    \begin{tabular}{|c|c|c|c|l|l|l|}
      \hline 
\multicolumn{4}{|c|}{Task}&Location&TeamID&Participants \\
      \hline 
$INS$ & $VTT$ & $ActEv$ & $AV$ &         &               &    \\
\hline

$---$ & $VTT$ & $-----$ & $AVS$ & $Eur$ & $EURECOM$ & EURECOM\\
$---$ & $VTT$ & $-----$ & $---$ & $Asia$ & $FDU$ & Fudan University\\
$---$ & $VTT$ & $-----$ & $---$ & $Asia$ & $KU\_ISPL$ & Korea University\\
$---$ & $***$ & $ActEv$ & $***$ & $Aus$ & $MUDSML$ & Monash University\\
$---$ & $VTT$ & $-----$ & $---$ & $Eur$ & $PicSOM$ & Aalto University\\
$INS$ & $***$ & $-----$ & $---$ & $Asia$ & $PKU\_ICST$ & Peking University\\
$---$ & $---$ & $-----$ & $AVS$ & $Eur$ & $SIRET$ & Charles University\\
$INS$ & $---$ & $ActEv$ & $---$ & $Eur$ & $HSMW\_TUC$ & University of Applied Sciences Mittweida\\&&&&&& Chemnitz University of Technology\\
$---$ & $VTT$ & $-----$ & $---$ & $Aus$ & $UTS\_ISA$ & Centre for Artificial Intelligence, \\&&&&&& University of Technology Sydney\\
$---$ & $VTT$ & $-----$ & $---$ & $Eur$ & $Insight\_DCU$ & Insight Dublin City University\\
$---$ & $VTT$ & $-----$ & $AVS$ & $NAm+SAm$ & $IMFD\_IMPRESEE$ & Millennium Institute Foundational Research \\&&&&&& on Data (IMFD) Chile;\\&&&&&& Impresee Inc ORAND S.A. Chile\\
$***$ & $---$ & $ActEv$ & $***$ & $Eur$ & $ITI\_CERTH$ & Information Technologies Institute, \\&&&&&& Centre for Research and Technology Hellas\\
$---$ & $---$ & $*****$ & $AVS$ & $Asia$ & $kindai\_kobe$ & Dept. of Informatics, Kindai University \\&&&&&& Graduate School of System Informatics,\\&&&&&& Kobe University\\
$---$ & $---$ & $ActEv$ & $---$ & $Asia$ & $NTT\_CQUPT$ & NTT Media Intelligence Laboratories \\&&&&&& Chongqing University of Posts and \\&&&&&& Telecommunications\\
$---$ & $***$ & $-----$ & $AVS$ & $Asia$ & $WasedaMeiseiSoftbank$ & Waseda University; Meisei University;\\&&&&&& SoftBank Corporation\\
$INS$ & $---$ & $ActEv$ & $---$ & $Asia$ & $BUPT\_MCPRL$ & Beijing University of Posts \\&&&&&& and Telecommunications\\
$---$ & $VTT$ & $-----$ & $---$ & $Asia$ & $KsLab$ & Nagaoka University of Technology\\
$INS$ & $***$ & $ActEv$ & $***$ & $Asia$ & $NII\_Hitachi\_UIT$ & National Institute of Informatics; Hitachi, Ltd; \\&&&&&& University of Information Technology, VNU-HCM\\
$---$ & $VTT$ & $-----$ & $---$ & $Asia$ & $RUC\_AIM3$ & Renmin University of China\\
$---$ & $VTT$ & $-----$ & $AVS$ & $Asia$ & $RUCMM$ & Renmin University of China;\\&&&&&& Zhejiang Gongshang University\\
$---$ & $---$ & $ActEv$ & $AVS$ & $Asia$ & $VIREO$ & City University of Hong Kong\\
$INS$ & $---$ & $-----$ & $---$ & $Asia$ & $WHU\_NERCMS$ & National Engineering Research Center\\&&&&&& for Multimedia Software\\
$---$ & $---$ & $-----$ & $AVS$ & $NAm$ & $FIU\_UM$ & Florida Intl. University; University of Miami\\
$---$ & $---$ & $ActEv$ & $---$ & $NAm$ & $UCF$ & University of Central Florida\\
$---$ & $---$ & $ActEv$ & $---$ & $Eur$ & $FraunhoferIOSB$ & Fraunhofer IOSB and Karlsruhe\\&&&&&& Institute of Technology (KIT)\\
$INS$ & $***$ & $ActEv$ & $AVS$ & $NAm+Asia+Aus$ & $Inf$ & Monash University; Renmin University;\\&&&&&& Shandong University\\
$---$ & $***$ & $-----$ & $AVS$ & $Asia$ & $ATL$ & Alibaba group, ZheJiang University\\

      \hline 
   \end{tabular} \\
\vspace{.2cm}
Task legend. INS:Instance Search; VTT:Video to Text; ActEv:Activities in Extended videos; AVS:Ad-hoc search; $--$:no run planned; $***$:planned but not submitted\\
   } 
  } 
\end{table*}

\begin{table*} 
\caption{Participants who did not submit any runs}
\label{nonparticipants}
  \vspace{1.0cm}
  \centering{
   \scriptsize{
    \begin{tabular}{|c|c|c|c|l|l|l|}
      \hline 
\multicolumn{4}{|c|}{Task}&Location&TeamID&Participants \\
      \hline 
        $INS$ & $VTT$ & $ActEv$ & $AVS$ &         &               &    \\
        \hline
$***$ & $---$ & $-----$ & $***$ & $Eur$ & $JRS$ & JOANNEUM RESEARCH\\
$---$ & $***$ & $*****$ & $***$ & $Eur$ & $MediaMill$ & University of Amsterdam\\
$***$ & $---$ & $-----$ & $---$ & $Asia$ & $IOACAS$ & University of Chinese Academy of Sciences\\
$***$ & $---$ & $-----$ & $***$ & $Asia$ & $D\_A777$ & Malla Reddy College of Engineering Technology, \\&&&&&& Department of Electronics and communication Engineering\\
$---$ & $***$ & $*****$ & $---$ & $NAm$ & $Arete$ & Scientific Computing Data Analytics \\&&&&&& Image Processing and Computer Vision\\
$---$ & $***$ & $-----$ & $---$ & $Asia$ & $GDGCV$ & G D Goenka University\\
$---$ & $***$ & $-----$ & $---$ & $Asia$ & $MAGUS\_ITAI.Wing$ & Nanjing University ITAI\\
$---$ & $---$ & $*****$ & $***$ & $Asia$ & $TokyoTech\_AIST$ & Tokyo Institute of Technology, National Institute \\&&&&&& of Advanced Industrial Science and Technology\\
$***$ & $---$ & $*****$ & $***$ & $NAm+Asia$ & $TeamCRN$ & Microsoft Research; Singapore Management University;\\&&&&&& University of Washington\\
$---$ & $---$ & $*****$ & $---$ & $NAm$ & $USF$ & University of South Florida, USF\\
$***$ & $---$ & $-----$ & $***$ & $Aus$ & $MIAOTEAM$ & University of Technology Sydney\\
$---$ & $---$ & $-----$ & $***$ & $Asia$ & $MET$ & Sun Yet-sen University\\

       \hline 
    \end{tabular} \\
\vspace{.2cm}
Task legend. INS:Instance Search; VTT:Video to Text; ActEv:Activities in extended videos; AVS:Ad-hoc search; $--$:no run planned; $**$:planned but not submitted\\
   } 
  } 
\end{table*}

This paper is an introduction to the evaluation framework, tasks, data, and measures used in the workshop. For detailed information about the approaches and results, the reader should see the various site reports and the results pages available at the workshop proceeding online page \cite{tv19pubs}.
Finally we would like to acknowledge that all work presented here has been cleared by HSPO (Human Subject Protection Office) under HSPO number: \#ITL-17-0025

\emph{Disclaimer: Certain commercial entities, equipment, or
materials may be identified in this document in order to describe an
experimental procedure or concept adequately. Such identification is
not intended to imply recommendation or endorsement by the National
Institute of Standards and Technology, nor is it intended to imply 
that the entities, materials, or equipment are necessarily the best 
available for the purpose. The views and conclusions contained herein 
are those of the authors and~should not be interpreted as necessarily
representing the official policies or endorsements, either expressed or 
implied, of IARPA (Intelligence Advanced Research Projects Activity), NIST, or the U.S. Government.}

\section{Datasets}
\subsection{BBC EastEnders Instance Search Dataset}
The BBC in collaboration the European Union's AXES project made
464 h of the popular and long-running soap opera EastEnders available to TRECVID for research since 2013. 
The data comprise 244 weekly ``omnibus'' broadcast files (divided into 471\,527 shots), transcripts, and 
a small amount of additional metadata. This dataset was adopted to test systems on retrieving target persons (characters)
doing specific actions.

\subsection{Vimeo Creative Commons Collection (V3C) Dataset}
The V3C1 dataset (drawn from a larger V3C video dataset \cite{rossetto2019v3c}) is composed of 7475 Vimeo videos (1.3 TB, 1000 h) with Creative Commons licenses and mean duration of 8 min. All videos have some metadata available such as title, keywords, and description in json files. The dataset has been segmented into 1\,082\,657 short video segments according to the provided master shot boundary files. In addition, keyframes and thumbnails per video segment have been extracted and made available. While the V3C1 dataset was adopted for testing, the previous Internet Archive datasets (IACC.1-3) of about 1800 h were available for development and training.

\subsection{Activity Detection VIRAT Dataset}
The VIRAT Video Dataset \cite{oh2011large} is a large-scale surveillance video dataset designed to assess the performance of activity detection algorithms in realistic scenes. The dataset was collected to facilitate both detection of activities and to localize the corresponding spatio-temporal location of objects associated with activities from a large continuous video. The stage for the data collection data was a group of buildings, and grounds and roads surrounding the area. The VIRAT dataset are closely aligned with real-world video surveillance analytics. In addition, we are also building a series of even larger multi-camera datasets, to be used in the future to organize a series of Activities in Extended Video (ActEV) challenges. The main purpose of the data is to stimulate the computer vision community to develop advanced algorithms with improved performance and robustness of human activity detection of multi-camera systems that cover a large area.

\subsection{Twitter Vine Videos}
A dataset of about 50\,000 video URL using the public Twitter stream API have been collected by NIST. Each video duration is about 6 sec. A list of 1044 URLs was distributed to participants of the video-to-text task. The previous years' testing data from 2016-2018 were also available for training (a set of about 5700 Vine URLs and their ground truth descriptions).

\subsection{Flickr Videos}
University of Twente\footnote{Thanks to Robin Aly} worked in consultation with NIST to collect Flickr video dataset available
under a Creative Commons license for research. The videos were then divided into segments of about 10s
in duration. A set of 91 videos divided into 74\,958 files was chosen independently by NIST. This year a set of about 1000 segmented video clips were selected randomly to complement the Twitter vine videos for the video-to-text task testing dataset.

\section{Ad-hoc Video Search}
This year we continued the Ad-hoc video search task that had resumed again in 2016 but adopted a new dataset (V3C1). The task models the end user video search use-case, who is looking for segments of video containing people, objects, activities, locations, etc. and combinations of the former. It was coordinated by NIST and by the Laboratoire d'Informatique de Grenoble\footnote{Thanks to Georges Qu\'{e}not}.

The Ad-hoc video search task was as follows. Given a standard set of shot boundaries for the V3C1 test collection and a list of 30 ad-hoc queries, participants were asked to return for each query, at most the top 1\,000 video clips from the standard master shot boundary reference set, ranked according to the highest probability of containing the target query. The presence of each query was assumed to be binary, i.e., it was either present or absent in the given standard video shot. 

Judges at NIST followed several rules in evaluating system output. If the query was true for some frame (sequence) within the shot, then it was true for the shot. This is a simplification adopted for the benefits it afforded in pooling of results and approximating the basis for calculating recall. In query definitions, ``contains x" or words to that effect are short for ``contains x to a degree sufficient for x to be recognizable as x by a human". This means among other things that unless explicitly stated, partial visibility or audibility may suffice. 
The fact that a segment contains video of a physical object representing the query target, such as photos, paintings, models, or toy versions of the target (e.g picture of Barack Obama vs Barack Obama himself), was NOT grounds for judging the query to be true for the segment. Containing video of the target within video may be grounds for doing so.

Like it's predecessor, in 2019 the task again supported experiments using the ``no annotation" version of the tasks: the idea is to promote the development of methods that permit the indexing of concepts in video clips using only data from the web or archives without the need of additional annotations. The training data could for instance consist of images or videos retrieved by a general purpose search engine (e.g. Google) using only the query definition with only automatic processing of the returned images or videos. This was implemented by adding the categories of ``E'' and ``F'' for the training types besides A and D:
In general, runs submitted were allowed to choose any of the below four training types:

\begin{itemize}
\item{A - used only IACC training data}
\item{D - used any other training data}
\item{E - used only training data collected automatically using only the official query textual description}
\item{F - used only training data collected automatically using a query built manually from the given official query textual description}
\end{itemize}

This means that even just the use of something like a face detector that was trained on non-IACC training data would disqualify the run as type A. 
\\\\
Three main submission types were accepted:
\begin{itemize}
\item{Fully automatic runs (no human input in the loop): System takes a query as input and produces result without any human intervention.}
\item{Manually-assisted runs: where a human can formulate the initial query based on topic and query interface, not on knowledge of collection or search results. Then system takes the formulated query as input and produces result without further human intervention.}
\item{Relevance-Feedback: System takes the official query as input and produce initial results, then a human judge can assess the top-5 results and input this information as a feedback to the system to produce a final set of results. This feedback loop is strictly permitted only once.}
\end{itemize}

A new progress subtask was introduced this year with the objective of measuring system progress on a set of 20 fixed topics. As a result, this year systems were allowed to submit results for 30 query topics (see Appendix \ref{appendixA} for the complete list) to be evaluated in 2019 and additional results for 20 common topics (not evaluated in 2019) that will be fixed for three years (2019-2021). Next year in 2020 NIST will evaluate progress runs submitted in 2019 and 2020 so that systems can measure their progress against two years (2019-2020) while in 2021 they can measure their progress against three years. 

A new extra one "Novelty" run type was allowed to be submitted within the main task. The goal of this run is to encourage systems to submit novel and unique relevant shots not easily discovered by other runs.

\subsection{Ad-hoc Data}
The V3C1 dataset (drawn from a larger V3C video dataset \cite{rossetto2019v3c}) was adopted as a testing dataset. It is composed of 7\,475 Vimeo videos (1.3 TB, 1000 h) with Creative Commons licenses and mean duration of 8 min. All videos will have some metadata available e.g., title, keywords, and description in json files. The dataset has been segmented into 1\,082\,657 short video segments according to the provided master shot boundary files. In addition, keyframes and thumbnails per video segment have been extracted and made available.
For training and development, all previous Internet Archive datasets (IACC.1-3) with about 1\,800 h were made available with their ground truth and xml meta-data files. Throughout this report we do not differentiate between a clip and a shot and thus they may be used interchangeably.

\subsection{Evaluation}
Each group was allowed to submit up to 4 prioritized runs per submission type, and per task type (main or progress) and two additional if they were ``no annotation'' runs. In addition, one novelty run type was allowed to be submitted within the main task.

In fact, 10 groups submitted a total of 85 runs with 47 main runs and 38 progress runs. The 47 main runs consisted of 
37 fully automatic, and 10 manually-assisted runs.

For each query topic, pools were created and randomly sampled as follows. The top pool sampled 100 \% of clips ranked 1 to 250 across all submissions after removing duplicates.  The bottom pool sampled 11.1 \% of ranked 251 to 1000 clips and not already included in a pool. 10 Human judges (assessors) were presented with the pools - one assessor per topic - and they judged each shot by watching the associated video and listening to the audio. Once the assessor completed judging for a topic, he or she was asked to rejudge all clips submitted by at least 10 runs at ranks 1 to 200. In all, 181\,649 clips were judged while 256\,753 clips fell into the unjudged part of the overall samples. Total hits across the 30 topics reached 23\,549 with 10\,910 hits at submission ranks from 1 to 100, 8428 hits at submission ranks 101 to 250 and 4211 hits at submission ranks between 251 to 1000.

\subsection{Measures}
Work at Northeastern University \cite{yilmaz06} has resulted in methods for estimating standard system performance measures using relatively small samples of the usual judgment sets so that larger numbers of features can be evaluated using the same amount of judging effort. Tests on past data showed the new measure (inferred average precision) to be a good estimator of average precision \cite{tv6overview}. This year mean extended inferred average precision (mean xinfAP) was used which permits sampling density to vary \cite{yilmaz08}. This allowed the evaluation to be more sensitive to clips returned below the lowest rank ($\approx$250) previously pooled and judged. It also allowed adjustment of the sampling density to be greater among the highest ranked items that contribute more average precision than those ranked lower.
The {\em sample\_eval} software \footnote{http://www-nlpir.nist.gov/projects/trecvid/\\trecvid.tools/sample\_eval/}, a tool implementing xinfAP, was used to calculate inferred recall, inferred precision, inferred average precision, etc., for each result, given the sampling plan and a submitted run. Since all runs provided results for all evaluated topics, runs can be compared in terms of the mean inferred average precision across all evaluated query topics. 

\subsection{Ad-hoc Results}

\begin{figure}[htbp]
\begin{center}
\includegraphics[height=2.0in,width=3in,angle=0]{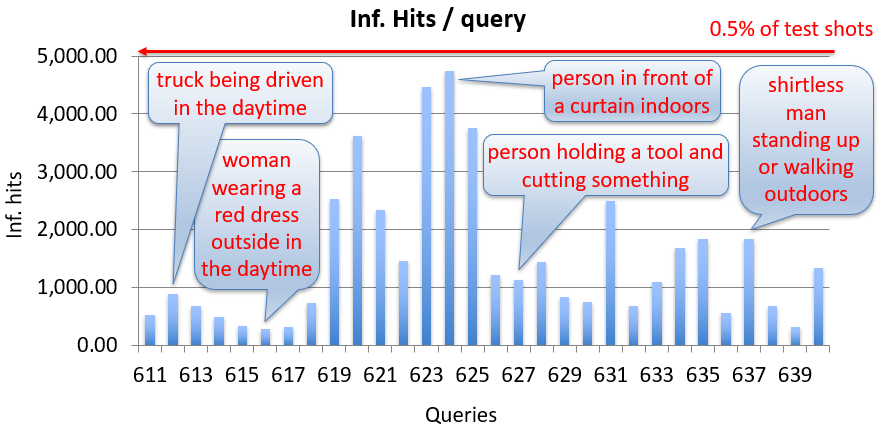}
\caption{AVS: Histogram of shot frequencies by query number}
\label{avs.frequencies}
\end{center}
\end{figure}

The frequency of correctly retrieved results varied greatly by query. 
Figure \ref{avs.frequencies} shows how many unique instances were found to be true for each tested query. The inferred true positives (TPs) of all queries are less than 0.5 \% from the total tested clips. 

Top 5 found queries were "person in front of a curtain indoors", "person wearing shorts outdoors", "person wearing a backpack", "person with a painted face or mask", and "one or more art pieces on a wall". On the other hand, the bottom 5 found queries were "woman wearing a red dress outside in the daytime". "inside views of a small airplane flying", "one or more picnic tables outdoors", "person smoking a cigarette outdoors", and "a drone flying".

The complexity of the queries or the nature of the dataset may be factors in the different frequency of hits across the 30 tested queries. One observation though is that less frequent hits are associated with queries that include more than one condition to be satisfied (i.e person gender, location, action being performed).

\begin{figure}[htbp]
\begin{center}
\includegraphics[height=2.0in,width=3in,angle=0]{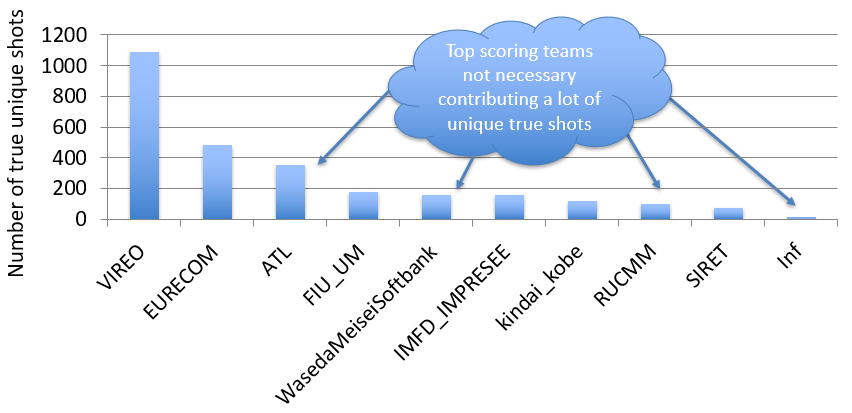}
\caption{AVS: Unique shots contributed by team}
\label{avs.unique.byteam}
\end{center}
\end{figure}

\begin{figure}[htbp]
\begin{center}
\includegraphics[height=2.5in,width=3.0in,angle=0]{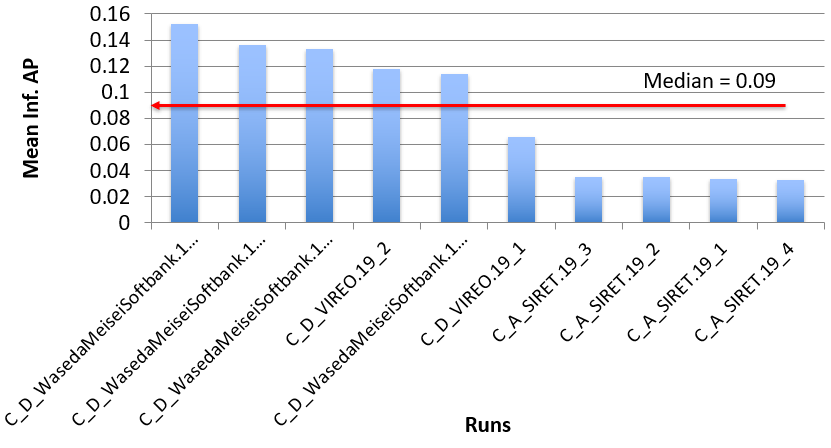}
\caption{AVS: xinfAP by run (manually assisted)}
\label{avs.m.all}
\end{center}
\end{figure}

\begin{figure}
\begin{center}
\includegraphics[height=2.5in,width=3.0in,angle=0]{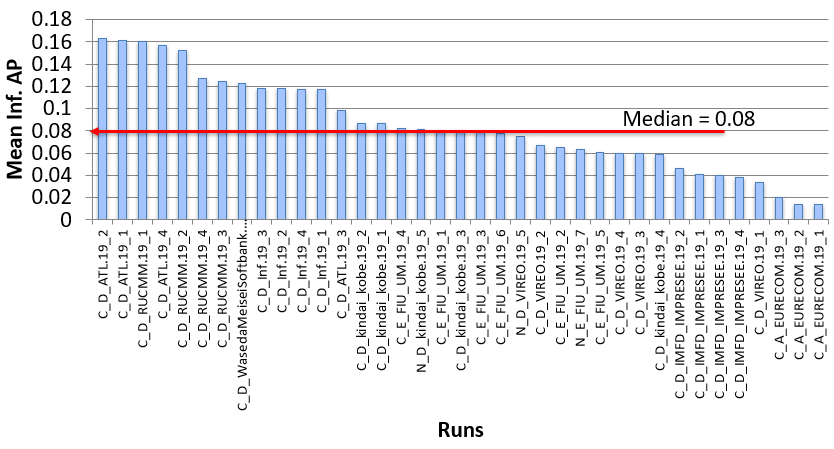}
\caption{AVS: xinfAP by run (fully automatic)}
\label{avs.f.all}
\end{center}
\end{figure}

Figure \ref{avs.unique.byteam} shows the number of unique clips found by the different participating teams. From this figure and the overall scores in figures \ref{avs.m.all} and \ref{avs.f.all} it can be shown that there is no clear relation between teams who found the most unique shots and their total performance. Many of the top performing teams did not contribute a lot of unique relevant shots. While the top two teams contributing the most unique relevant shots are not among the top automatic team runs. This observation is consistent with the past few years. 

Figures \ref{avs.m.all} and \ref{avs.f.all} show the results of all the 10 manually-assisted and 37 fully automatic run submissions respectively. 

As this year the ad-hoc task is adopting a new dataset, we can not compare the performance against previous years. However the max and median scores for both automatic and manually-assisted runs are very near. That may indicate that the best automatic system performance is comparable to manually-assisted runs after
a human in the loop is needed to adjust the official query text before submitting the query to the system.

We should also note here that 7 runs were submitted under the "E" training category, 0 runs using category "F" while the majority of runs (33) were of type "D". While the evaluation supported a relevance feedback run types, this year no submissions were received under this category.

Compared to the semantic indexing task that was running to detect single concepts (e.g airplane, animal, bridge,...etc) from 2010 to 2015 it can be shown from the results that the ad-hoc task is still very hard and systems still have a lot of room to research methods that can deal with unpredictable queries composed of one or more concepts including their interactions.

A new novelty run type was introduced this year to encourage submitting unique (hard to find) relevant shots. Systems were asked to label their runs as either of novelty type or common type runs. A new novelty metric was designed to score runs based on how good are they in detecting unique relevant shots. A weight was given to each topic and shot pairs such as follows:

\[TopicX\_ShotY_{weight} (x) = 1 - \frac{N}{M}\]

Where N is the number of times Shot Y was retrieved for topic X by any run submission, and M is the number of total runs submitted by all teams. For instance, a unique relevant shot weight will be 0.978 (given 47 runs in 2019) while a shot submitted by all runs will be assigned a weight of 0.

\begin{figure}
\begin{center}
\includegraphics[height=2.5in,width=3.0in,angle=0]{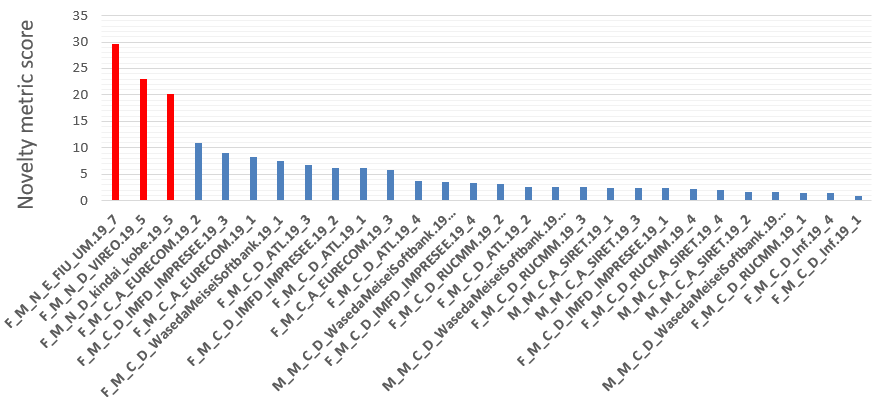}
\caption{AVS: Novelty metric scores}
\label{avs.novelty.scores}
\end{center}
\end{figure}

For Run R and for all topics, we calculate the summation S of all unique shot weights only and the final novelty metric score is the mean score across all evaluated 30 topics. Figure \ref{avs.novelty.scores} shows the novelty metric scores. The red bars indicate the submitted novelty runs. In comparison with all the common runs from other teams, novelty runs achieved the top 3 scores. We should note here that in running this experiment, for a team that submitted a novelty run, we removed all it's other common runs submitted. The reason for doing this was the fact that usually for a given team there will be many overlapping shots within all it's submitted runs. So to accurately judge how novel is their submitted novelty runs we removed their other common runs in this scoring procedure. It was difficult to do the same for other team runs because they did not submit novelty runs.

\begin{figure}
\begin{center}
\includegraphics[height=2.5in,width=3.0in,angle=0]{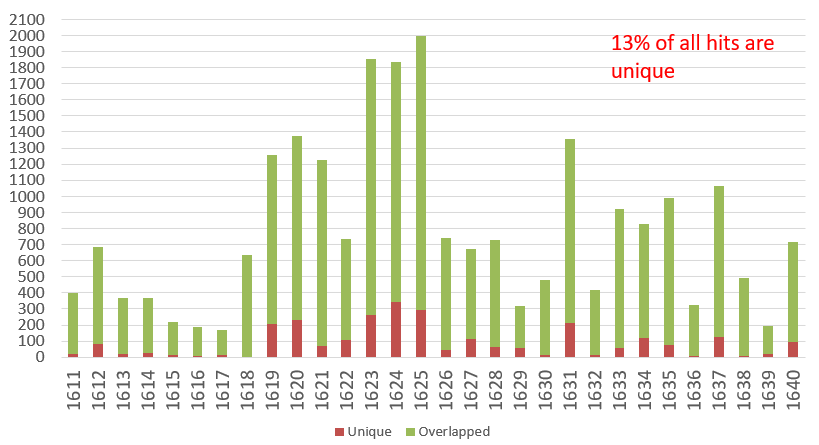}
\caption{AVS: Unique vs overlapping results}
\label{avs.unique.overlapped}
\end{center}
\end{figure}

Figure \ref{avs.unique.overlapped} shows for each topic the number of relevant and unique shots submitted by all teams combined (red color). On the other hand, the green color counts the total non-unique (true) shots (submitted by at least 2 or more teams) per topic. The three topics 1623,1624, and 1625 achieved the most unique and common hits overall.

\begin{figure}[htbp]
\begin{center}
\includegraphics[height=2.0in,width=3.0in,angle=0]{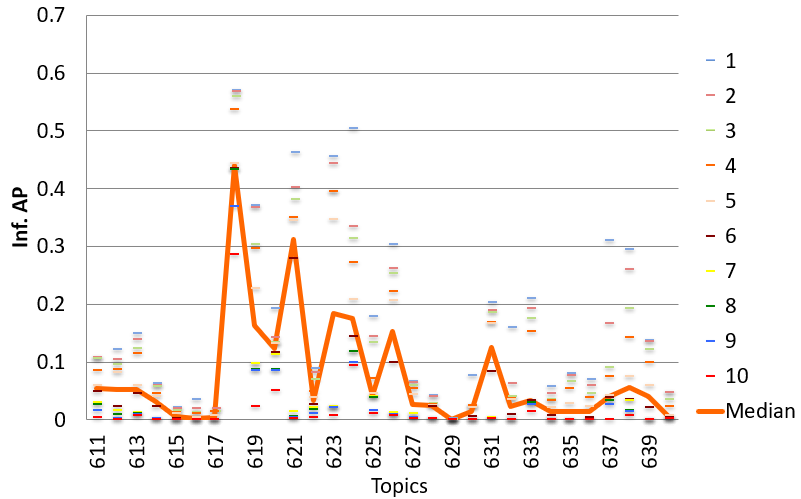}
\caption{AVS: Top 10 runs (xinfAP) by query number (manually assisted)}
\label{avs.top10.m}
\end{center}
\end{figure}

\begin{figure}[htbp]
\begin{center}
\includegraphics[height=2.0in,width=3.0in,angle=0]{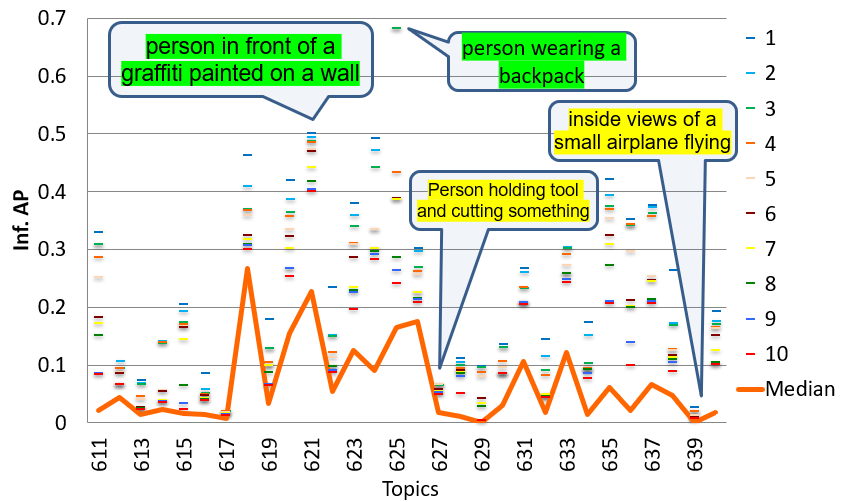}
\caption{AVS: Top 10 runs (xinfAP) by query number (fully automatic)}
\label{avs.top10.f}
\end{center}
\end{figure}

Figures \ref{avs.top10.m} and \ref{avs.top10.f} show the performance of the top 10 teams across the 30 queries. Note that each series in this plot represents a rank (from 1 to 10) of the scores, but not necessary that all scores at given rank belong to a specific team.
A team's scores can rank differently across the 30 queries. Some samples of top queries are highlighted in green while samples of bottom queries are highlighted in yellow.

A main theme among the top performing queries is their composition of more common visual concepts (e.g painted face, backpack, curtain, coral reef, graffiti, etc) compared to the bottom ones which require more temporal analysis for some activities and combination of one or more facets of who,what and where/when (e.g person or object doing certain action or activity in a specific location/time, etc). 

In general there is a noticeable spread in score ranges among the top 10 runs specially with high performing topics which may indicate the variation in the performance of the used techniques and that there is still room for further improvement. However for topics not performing well, usually all top 10 runs are condensed together with low spread between their scores. In addition, there is no clear relation between the performance of automatic runs vs manually-assisted runs. For example, some topics performed well in automatic runs and poor in manually-assisted runs and vice versa.

\begin{figure}
\begin{center}
\includegraphics[height=2.0in,width=3.0in,angle=0]{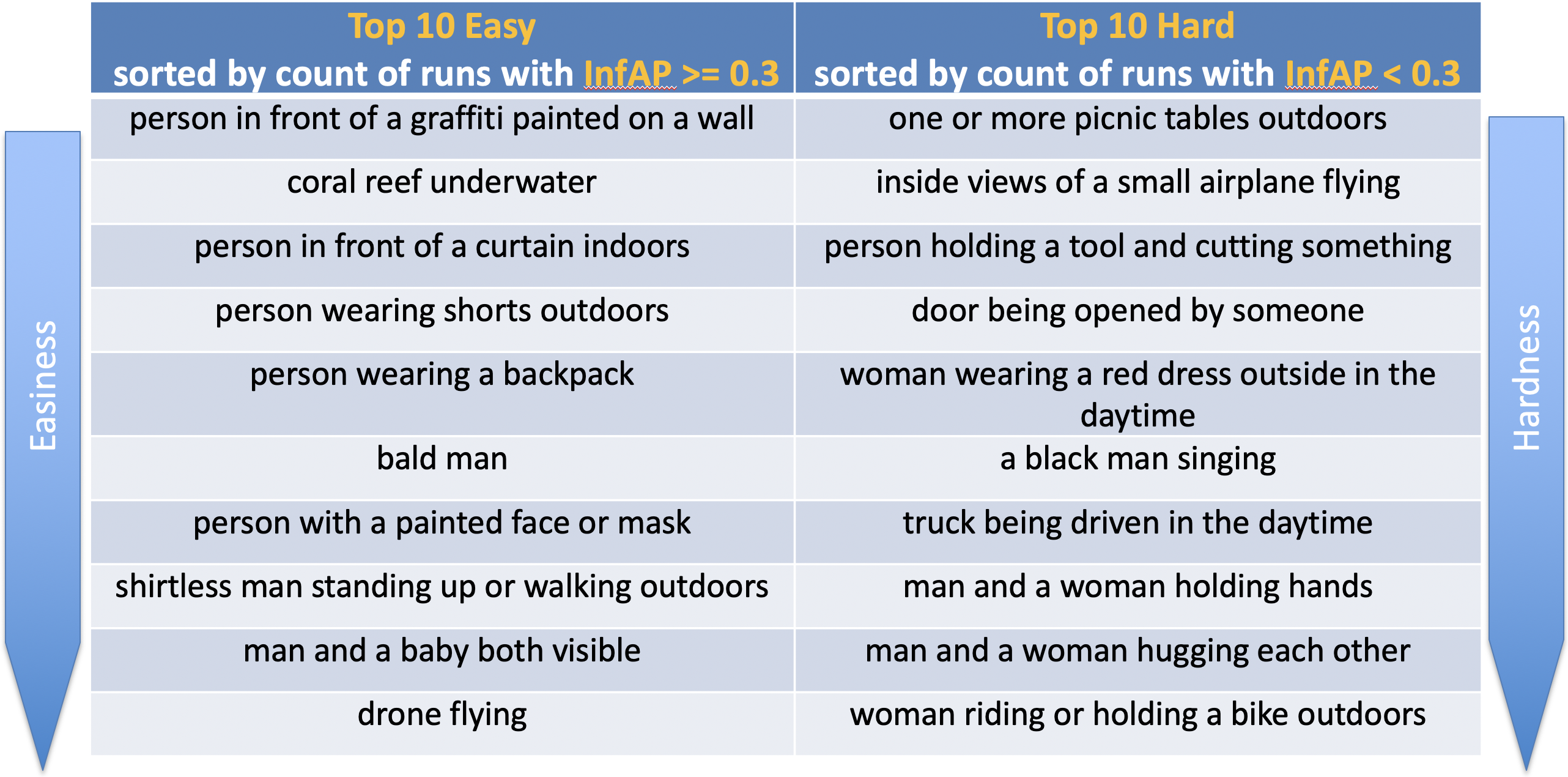}
\caption{AVS: Easy vs Hard topics}
\label{avs.easy.hard.topics}
\end{center}
\end{figure}

\begin{figure}
\begin{center}
\includegraphics[height=2.0in,width=3.0in,angle=0]{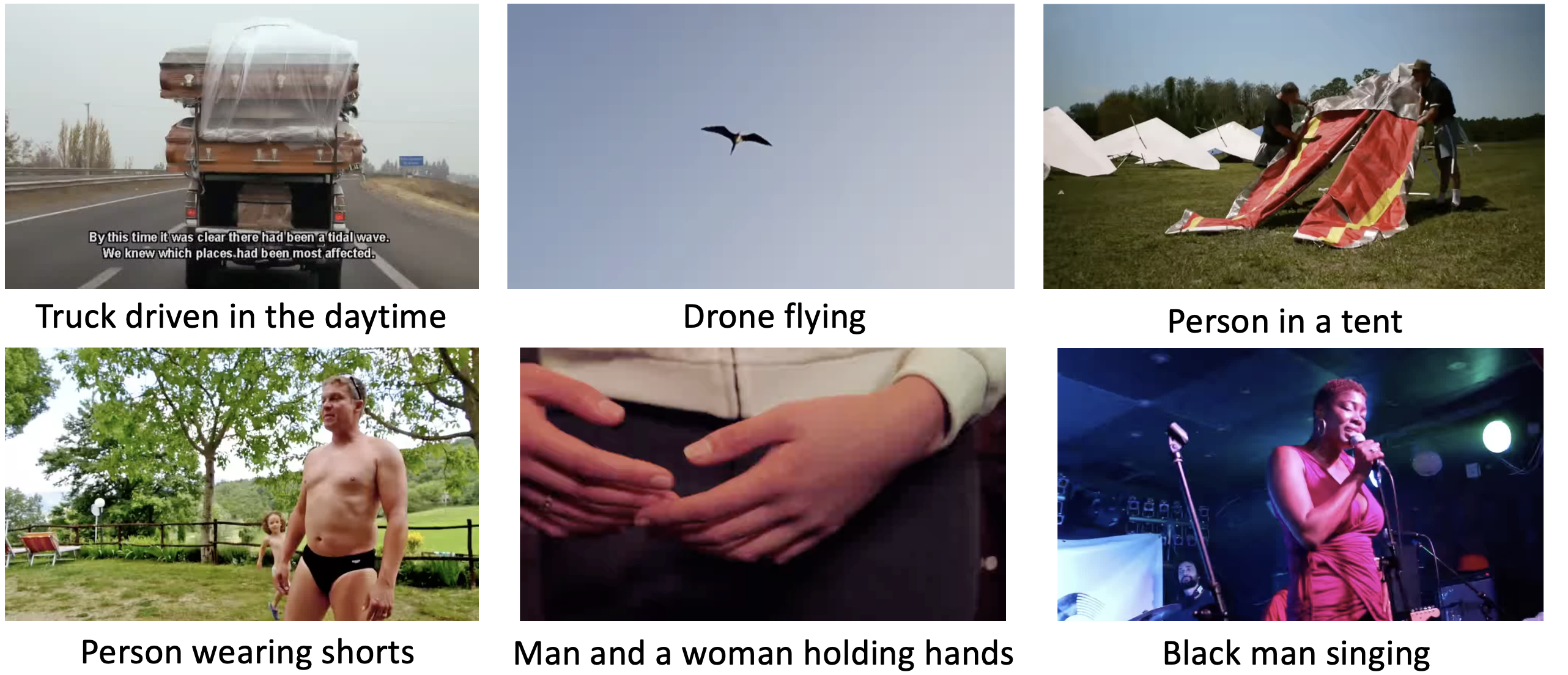}
\caption{AVS: Samples of frequent false positive results}
\label{avs.fp}
\end{center}
\end{figure}

In order to analyze which topics in general were the most easy or difficult we sorted topics by number of runs that scored xInfAP $>$= 0.3 for any given topic and assumed that those were the easiest topics, while xInfAP $<$ 0.3 indicates a hard topic. Using this criteria, Figure \ref{avs.easy.hard.topics} shows a table with the easiest/hardest topics at the top rows. From that table it can be concluded that hard topics are associated with activities, actions and more dynamics or conditions that must be satisfied in the retrieved shots compared to easily identifiable visual concepts within the easy topics. One exception to this observation is the topic "One or more picnic tables outdoors" which is an easy topic for retrieving just tables. However, most likely the type of the tables returned by systems was not picnic or may be was not outdoors. Sample results of frequently submitted false positive shots are demonstrated\footnote{All figures are in the public domain and permissible under HSPO \#ITL-17-0025} in Figure \ref{avs.fp}.

\begin{figure}[htbp]
\begin{center}
\includegraphics[height=2.5in,width=3.0in,angle=0]{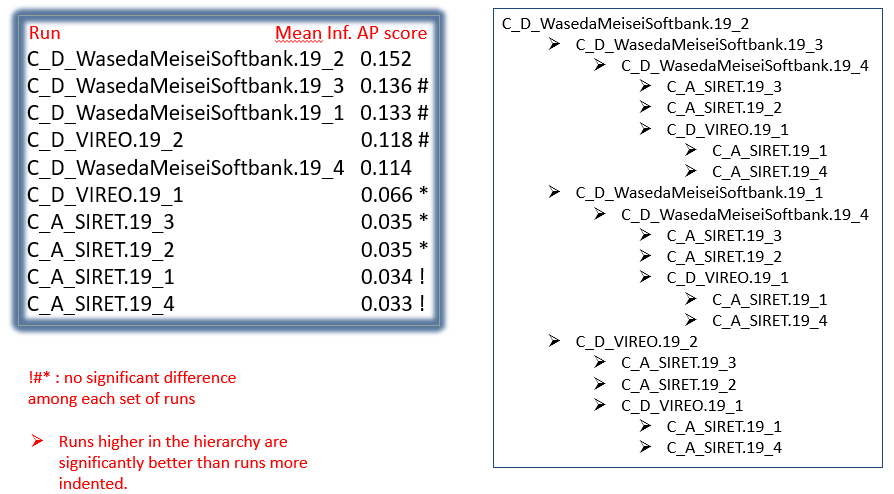}
\caption{AVS: Statistical significant differences (top 10 manually-assisted runs). The symbols \#,! and * denotes that there is no statistical significance between those runs for a given team}
\label{avs.top10.rand.m}
\end{center}
\end{figure}

\begin{figure}[htbp]
\begin{center}
\includegraphics[height=2.5in,width=3.0in,angle=0]{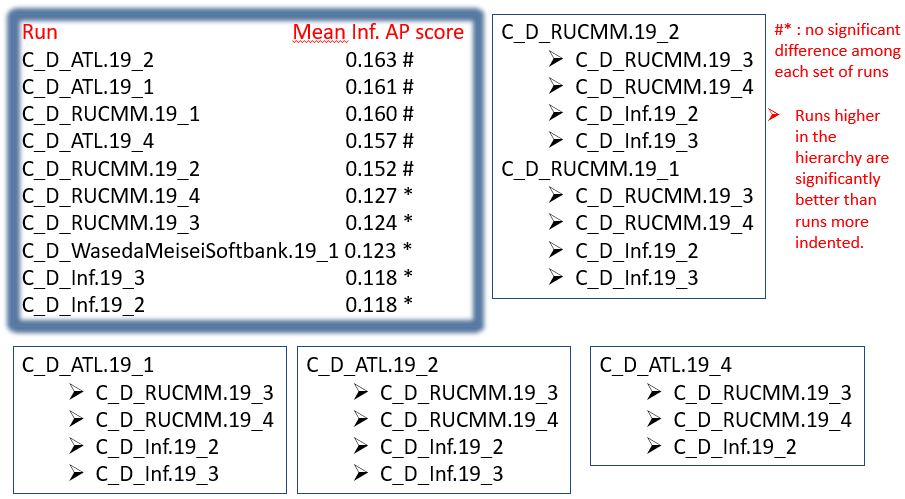}
\caption{AVS: Statistical significant differences (top 10 fully automatic runs). The symbols \#,! and * denotes that there is no statistical significance between those runs for a given team}
\label{avs.top10.rand.f}
\end{center}
\end{figure}

To test if there were significant differences between the systems' performance, we applied a randomization test \cite{manly97} on the top 10 runs for manually-assisted and automatic run submissions as shown in Figures \ref{avs.top10.rand.m} and \ref{avs.top10.rand.f} respectively using significance threshold of p$<$0.05. These figures indicate the order by which the runs are significant according to the randomization test. Different levels of indentation means a significant difference according to the test. Runs at the same level of indentation are indistinguishable in terms of the test and all equivalent runs are marked with the same symbol (e.g. *, \#, !, etc). For example, it can be shown that there is no sufficient evidence for a significant difference between top 5 automatic runs and between runs ranked 6th to 10th as well.

\begin{figure}[htbp]
\begin{center}
\includegraphics[height=2.0in,width=3.0in,angle=0]{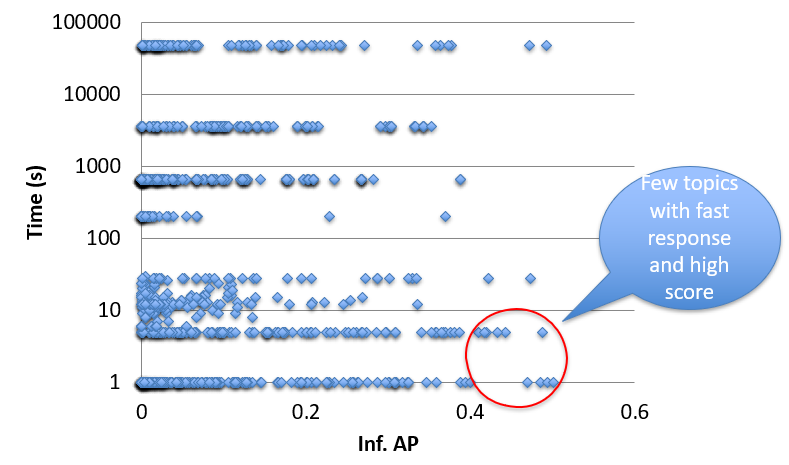}
\caption{AVS: Processing time vs Scores (fully automatic)}
\label{avs.f.time.score}
\end{center}
\end{figure}

\begin{figure}[htbp]
\begin{center}
\includegraphics[height=2.0in,width=3.0in,angle=0]{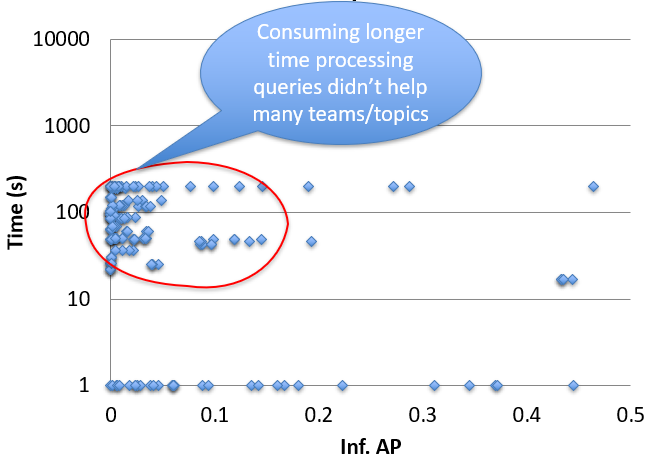}
\caption{AVS: Processing time vs Scores (Manually assisted)}
\label{avs.m.time.score}
\end{center}
\end{figure}

Among the submission requirements, we asked teams to submit the processing time that was consumed to return the result sets for each query. Figures \ref{avs.f.time.score} and \ref{avs.m.time.score} plots the reported processing time vs the InfAP scores among all run queries for automatic and manually-assisted runs respectively. It can be shown that spending more time did not necessarily help in many cases and few queries achieved high scores in less time. There is more work to be done to make systems efficient and effective at the same time.

In order to measure how were the submitted runs diverse, we measured the percentage of common clips returned for all queries between each pair of runs. We found that on average about 8 \% (minimum 3 \%) of submitted clips are common between any pair of runs. 

\subsection{Ad-hoc Observations and Conclusions}

In 2018 we concluded 1-cycle of three years of Ad-hoc task using the internet Archive (IACC.3) dataset \cite{2016trecvidover}. This year, a new dataset ,Vimeo Creative Commons Collection (V3C1), is being used for testing. NIST Developed a set of 90 queries to be used between 2019-2021 including a progress subtask. To summarize major observations in 2019 we can see that most Submitted runs are of training type “D”, no relevance feedback submissions were received, and new “novelty” run type (and metric) was utilized this year. Novelty runs proved to submit unique true shots compared to common run types. Overall, team participation and task completion rate are stable. While manually-assisted runs are decreasing, there is a high participation in the progress subtask. The absolute number of hits are higher than previous years. However, we can’t compare the performance with previous years (2016-2018) due to the new dataset and queries. Fully automatic and Manually-assisted performance are almost similar. Among high scoring topics, there is more room for improvement among systems. Among low scoring topics, most systems scores are collapsed in small narrow range. Dynamic topics (actions, interactions, multi-facets ..etc) are the hardest topics. Most systems are slow. Few systems are efficient and effective retrieving fast and accurate results. Finally, the task is still challenging!

As a general high-level systems overview, we can see that there is two main competing approaches among participating teams: “concept banks” and “(visual-textual) embedding spaces”.
Currently there is a significant advantage for “embedding space” approaches, especially for fully automatic search and even overall. Training data for semantic spaces included MSR and TRECVID VTT, TGIF, IACC.3, Flickr8k, Flickr30k, MS COCO, and Conceptual Captions.

For detailed information about the approaches and results for individual teams' performance and runs, the reader should see the various site reports \cite{tv19pubs} in the online workshop notebook proceedings.

\section{Instance search}

\begin{table*}[t]
\caption{Instance search pooling and judging statistics}
\label{searchstats} 
  \vspace{0.5cm}
  \centering{
   \small{
    \begin{tabular}{|c|c|c|c|c|c|c|c|c|}
      \hline 

\parbox{1.3cm}{Topic \\ number} & 
\parbox{1.5cm}{Total \\ submitted} & 
\parbox{1.5cm}{Unique \\ submitted} & 
\parbox{1.0cm}{total \\ that \\ were \\ unique\\ \%} & 
\parbox{1.0cm}{Max.\\ result\\ depth \\ pooled\\} & 
\parbox{1.2cm}{Number \\ judged} & 
\parbox{1.0cm}{unique \\ that \\ were \\ judged\\ \%} & 
\parbox{1.2cm}{Number \\ relevant} & 
\parbox{1.2cm}{judged \\ that \\ were \\ relevant\\ \%}

 \\ \hline
 
9249&27122&7343&27.07&520&4360&59.38&439&10.07\\ \hline
9250&27225&8100&29.75&520&4827&59.59&367&7.60\\ \hline
9251&27029&7324&27.10&520&4178&57.05&241&5.77\\ \hline
9252&27228&7225&26.54&520&4332&59.96&352&8.13\\ \hline
9253&27031&7144&26.43&520&4086&57.19&575&14.07\\ \hline
9254&27092&7615&28.11&520&4461&58.58&524&11.75\\ \hline
9255&27278&8835&32.39&520&5153&58.32&275&5.34\\ \hline
9256&27220&9359&34.38&520&5309&56.73&250&4.71\\ \hline
9257&27073&8456&31.23&520&4979&58.88&178&3.58\\ \hline
9258&27418&8169&29.79&520&4894&59.91&41&0.84\\ \hline
9259&27344&8483&31.02&520&5322&62.74&91&1.71\\ \hline
9260&27212&7102&26.10&520&4350&61.25&56&1.29\\ \hline
9261&27162&6627&24.40&520&4185&63.15&234&5.59\\ \hline
9262&27543&8174&29.68&520&4766&58.31&229&4.80\\ \hline
9263&28000&9524&34.01&520&5801&60.91&46&0.79\\ \hline
9264&28000&7964&28.44&520&4895&61.46&91&1.86\\ \hline
9265&27759&7471&26.91&520&4677&62.60&196&4.19\\ \hline
9266&27964&7627&27.27&520&4565&59.85&499&10.93\\ \hline
9267&27122&7701&28.39&520&4697&60.99&35&0.75\\ \hline
9268&27140&8661&31.91&520&4924&56.85&39&0.79\\ \hline
9269&25085&8122&32.38&520&4505&55.47&139&3.09\\ \hline
9270&25070&7454&29.73&520&4543&60.95&273&6.01\\ \hline
9271&25040&9929&39.65&520&5478&55.17&101&1.84\\ \hline
9272&26000&9073&34.90&520&5268&58.06&115&2.18\\ \hline
9273&25905&8515&32.87&520&4816&56.56&139&2.89\\ \hline
9274&25167&6410&25.47&520&3847&60.02&487&12.66\\ \hline
9275&25641&7192&28.05&520&4550&63.28&471&10.35\\ \hline
9276&25940&8995&34.68&520&4905&54.53&29&0.59\\ \hline
9277&25068&7749&30.91&520&4589&59.22&40&0.87\\ \hline
9278&25059&7242&28.90&520&4337&59.89&40&0.92\\ \hline

    \end{tabular}
   } 
  } 
\end{table*}

An important need in many situations involving video collections
(archive video search/reuse, personal video organization/search,
surveillance, law enforcement, protection of brand/logo use) is to find more video segments of a certain specific person, object, or place, given one or more visual examples of the specific item. Building on work from previous years in the concept detection task \cite{awad2016trecvid} the instance search task seeks to address some of these needs. For six years (2010-2015) the instance search task tested systems on retrieving specific instances of individual objects, persons and locations. From 2016 to 2018, a new query type, to retrieve specific persons in specific locations had been introduced. From 2019, a new query type has been introduced to retrieve instances of named persons doing named actions.

\subsection{Instance Search Data}
The task was run for three years starting in 2010 to explore task definition and evaluation issues using data of three sorts: Sound and Vision (2010), British Broadcasting Corporation (BBC) rushes (2011), and Flickr (2012). Finding realistic test data, which contains sufficient recurrences of various specific objects/persons/locations under varying conditions has been difficult.

In 2013 the task embarked on a multi-year effort using 464 h of the BBC soap opera EastEnders. 244 weekly ``omnibus'' files were divided by the BBC into 471\,523 video clips to be used as the unit of retrieval. The videos present a ``small world'' with a slowly changing set of recurring people (several dozen), locales (homes, workplaces, pubs, cafes, restaurants, open-air market, clubs, etc.), objects (clothes, cars, household goods, personal possessions, pets, etc.), and views (various camera positions, times of year, times of day).

\subsection{System task}

The instance search task for the systems was as follows. Given a collection of test videos, a master shot reference, a set of known action example videos, and a collection of topics (queries) that delimit a specific person performing a specific action, locate for each topic up to the 1000 clips most likely to contain a recognizable instance of the person performing one of the named actions. 

Each query consisted of a set of:
\begin{itemize}
\item{The name of the target person}
\item{The name of the target action}
\item{4 example frame images drawn at intervals from videos
  containing the person of interest. For each frame image:}
\begin{itemize}
  \item{a binary mask covering one instance of the target person}
  \item{the ID of the shot from which the image was taken}
\end{itemize}
\item{4 - 6 short sample video clips of the target action}
\item{A text description of the target action}
\end{itemize}

Information about the use of the examples was reported by participants
with each submission. The possible categories for use of examples
were as follows:
\begin{enumerate} \itemsep0pt \parskip0pt
\item[A] one or more provided images - no video used 
\item[E] video examples (+ optional image examples) 
\end{enumerate}

Each run was also required to state the source of the training data used. This year participants were allowed to use training data from an external source, instead of, or in addition to the NIST provided training data. The following are the options of training data to be used:
\begin{enumerate} \itemsep0pt \parskip0pt
\item[A] Only sample video 0
\item[B] Other external data
\item[C] Only provided images/videos in the query
\item[D] Sample video 0 AND provided images/videos in the query (A+C)
\item[E] External data AND NIST provided data (sample video 0 OR query images/videos)
\end{enumerate}

\subsection{Topics}

NIST viewed a sample of test videos and developed a list of recurring
actions and the persons performing these actions. In order to test the effect of persons or actions on the performance of a given query, the topics tested different target persons performing the same actions. In total, this year we provided 30 unique queries to be evaluated this year, in addition to 20 common queries which will be stored and evaluated in later years and used to measure teams progress year-on-year (10 will be evaluated in 2020 to measure 2019-2020 progress, 10 remaining queries will be evaluated in 2021 to measure 2019-2021 progress). 12 progress runs were submitted by 3 separate teams in 2019. The 30 unique queries provided for this years task comprised of 10 individual persons and 12 specific actions. The 20 common queries which will be evaluated in later years comprised of 9 individual persons and 10 specific actions (Appendix \ref{appendixB}).

The guidelines for the task allowed the use of metadata assembled by
the EastEnders fan community as long as its use was documented by
participants and shared with other teams.

\subsection{Evaluation}

Each group was allowed to submit up to 4 runs (8 if submitting pairs
that differ only in the sorts of examples used). In total, 6 groups
submitted 26 automatic and 2 interactive runs (using only the first
21 topics). Each interactive search was limited to 5 minutes.

The submissions were pooled and then divided into strata based on the
rank of the result items.  For a given topic\footnote{Please refer to Appendix \ref{appendixB} for query descriptions.}, the submissions for that
topic were judged by a NIST assessor who played each submitted shot
and determined if the topic target was present.  The assessor started
with the highest ranked stratum and worked his/her way down until too
few relevant clips were being found or time ran out. 
In general, submissions were pooled and judged down to at least rank 100, resulting
in 141\,599 judged shots including 6\,592 total relevant shots (4.66\%).
Table \ref{searchstats} presents information about the pooling and judging.

\subsection{Measures}
This task was treated as a form of search, and evaluated accordingly with average precision for each query in each run and per-run mean average precision (MAP) over all queries. While speed and location accuracy
were also of interest here, of these two, only speed was reported.

\subsection{Instance Search Results}
\begin{figure}[htbp]
\begin{center}
\includegraphics[height=2.5in,width=3in,angle=0]{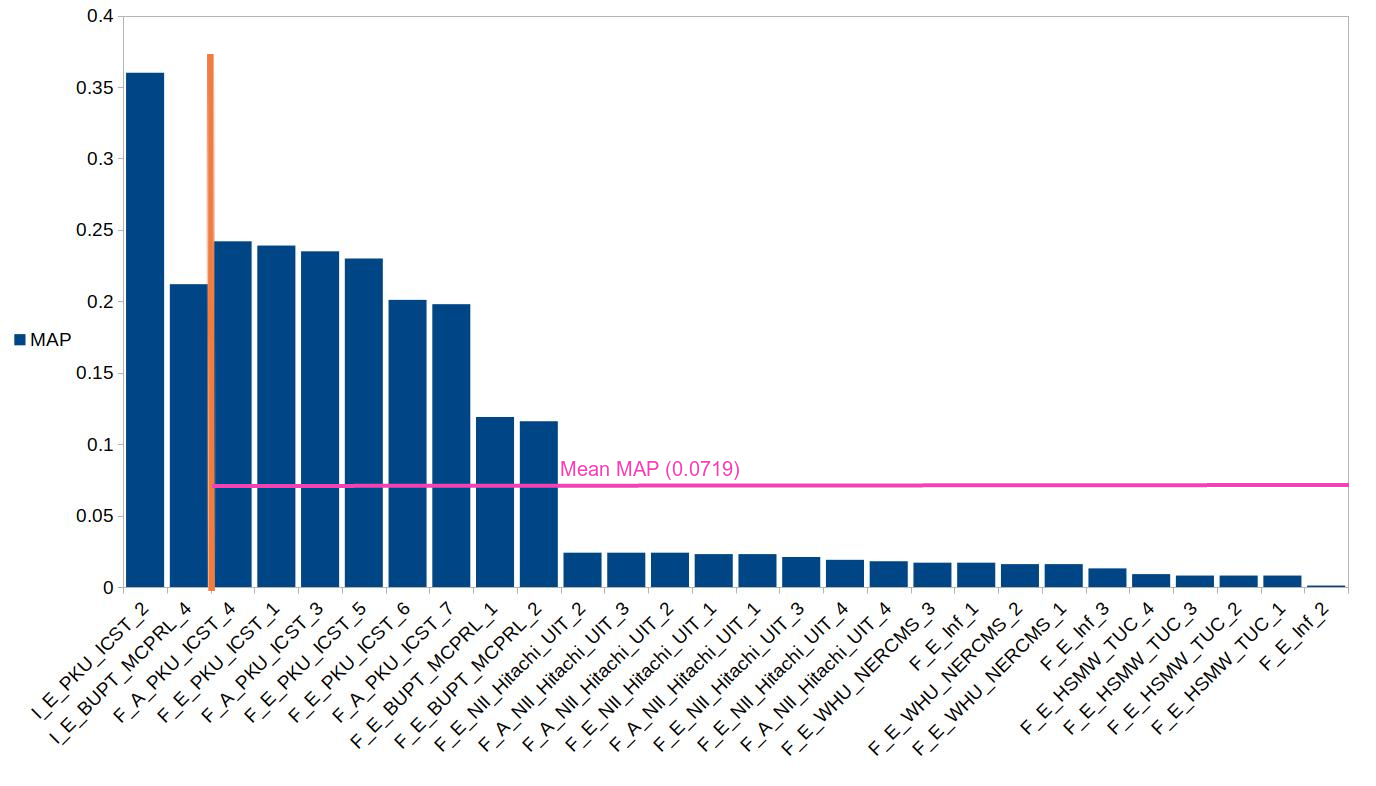}
\caption{INS: Mean average precision scores for automatic and interactive systems}
\label{ins.auto.scores}
\end{center}
\end{figure}

Figure \ref{ins.auto.scores} shows the sorted scores
of runs for both automatic and interactive systems. With only two interactive runs submitted this year these runs have been included in the automatic runs chart. Results show a big decrease from those recorded on the INS task over the previous years, however, the INS task has been completely changed this year and results can not be compared in any way to previous years. In subsequent years we can compare results using the set of common queries.

\begin{figure}
\begin{center}
\includegraphics[height=3.5in,width=3in,angle=0]{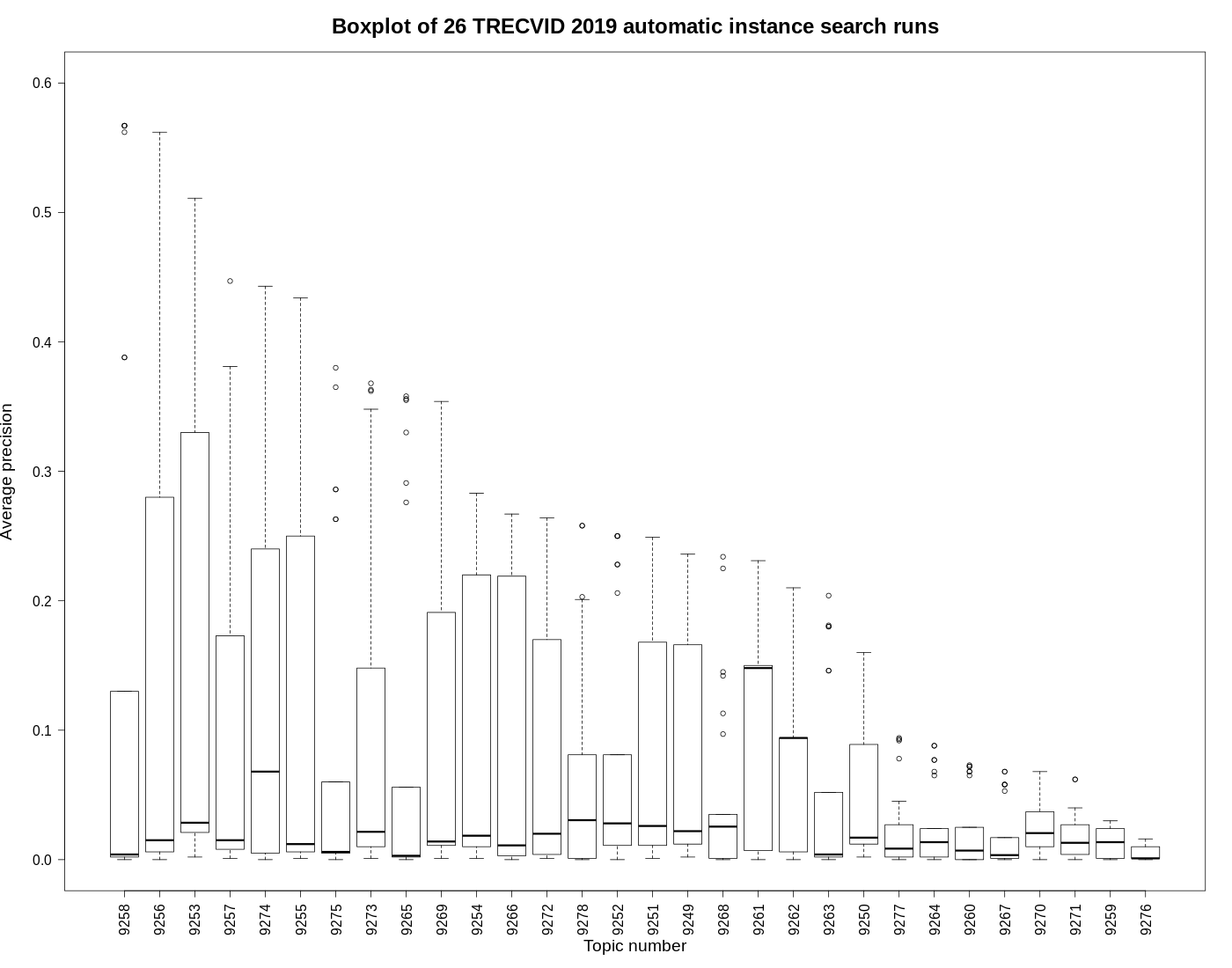}
\caption{INS: Boxplot of average precision by topic for automatic runs.}
\label{ins.auto.bp.topic.v.ap}
\end{center}
\end{figure}

Figure \ref{ins.auto.bp.topic.v.ap} shows the distribution of
automatic run scores (average precision) by topic as a box plot. The
topics are sorted by the maximum score with the best performing topic
on the left. Median scores vary from 0.148 down to 0.001. 
The main factor affecting topic difficulty this year is the target action.

One thing of interest in this figure are the topics 9261, 9262, and 9274: Max, Phil and Jack shouting. These topics do not score among the highest for maximum scores, but do have the highest median scores.

\begin{figure}
\begin{center}
\includegraphics[height=3.5in,width=3in,angle=0]{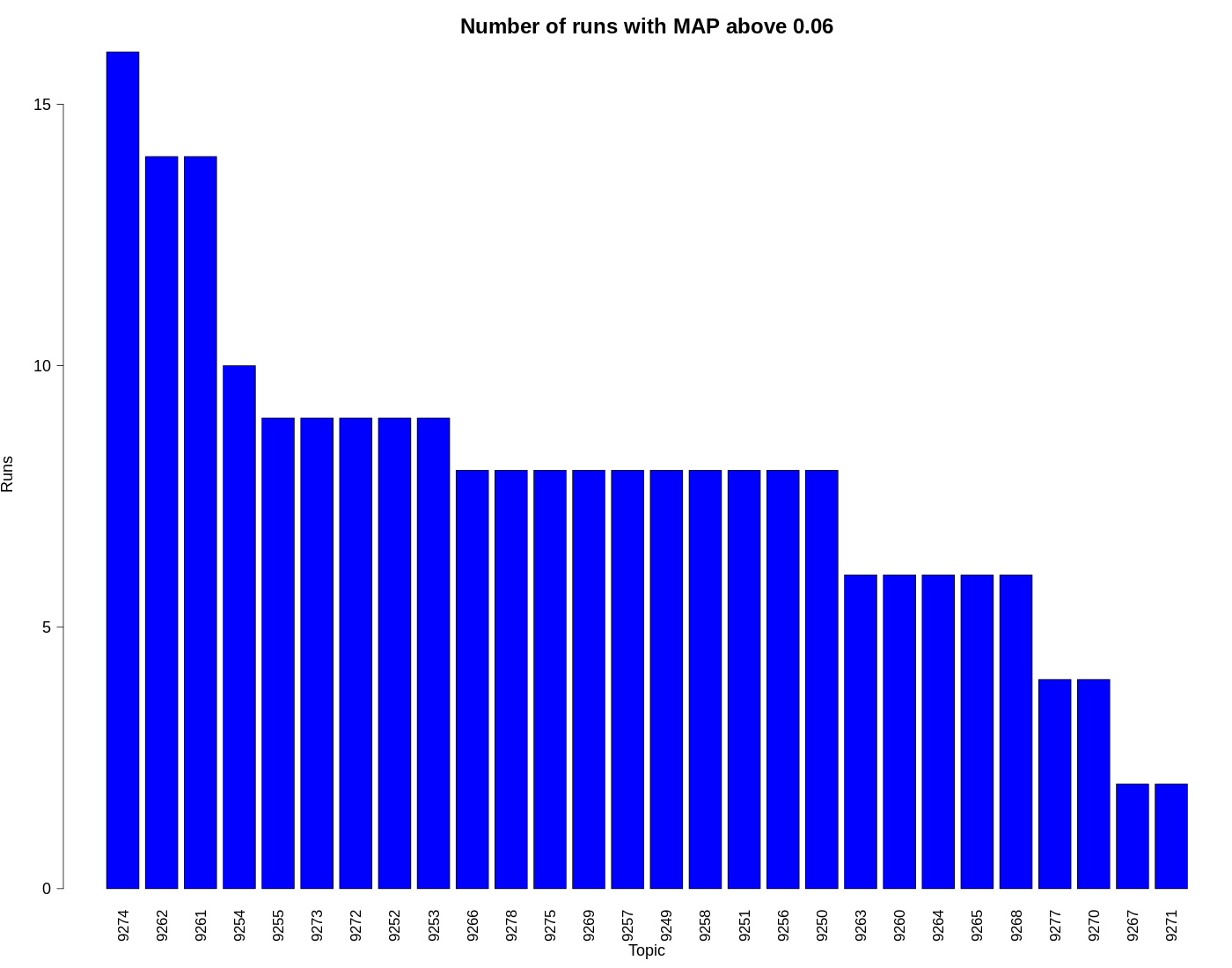}
\caption{INS: Easiest topics for automatic systems}
\label{ins.easy.topics}
\end{center}
\end{figure}

\begin{figure}
\begin{center}
\includegraphics[height=3.5in,width=3in,angle=0]{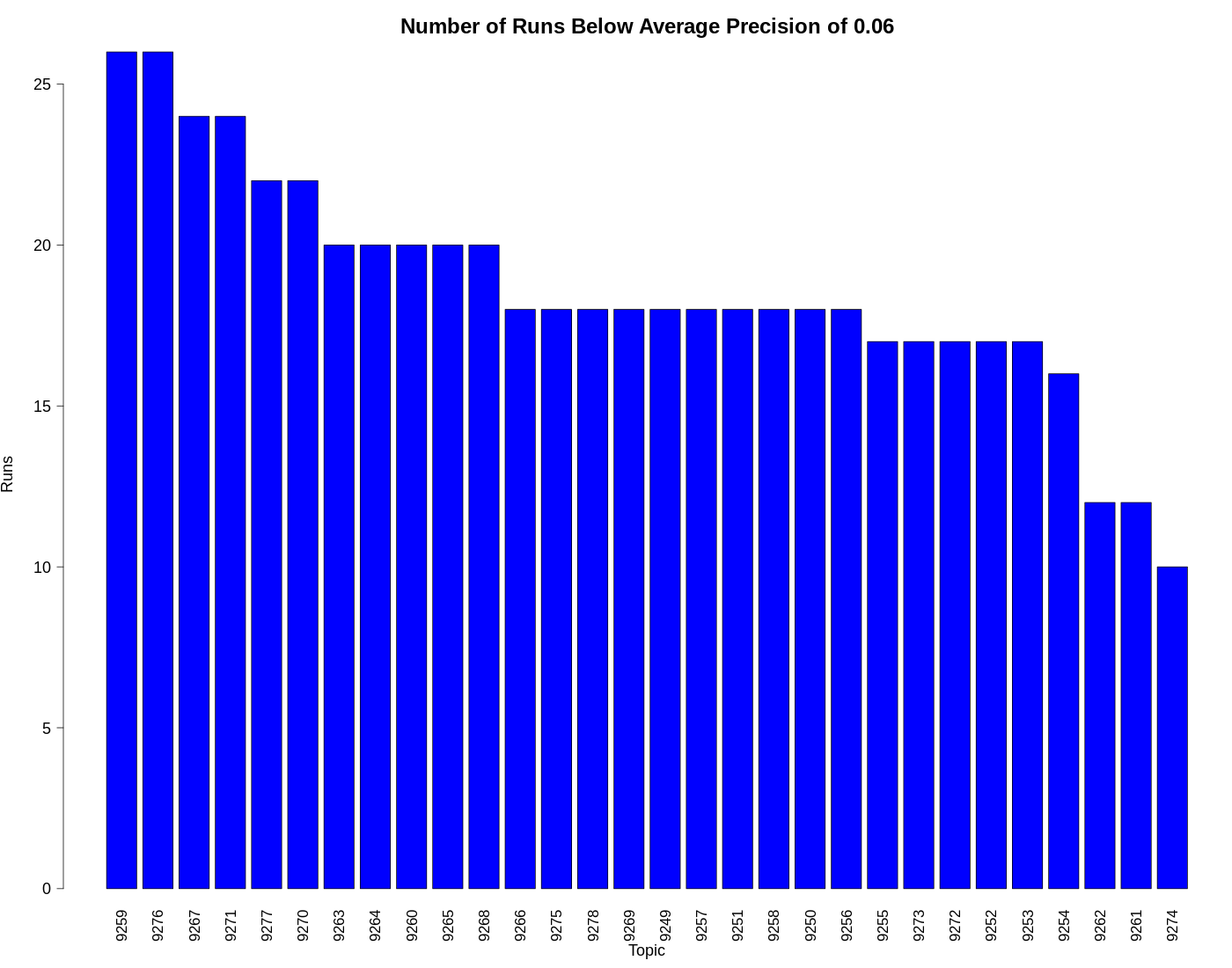}
\caption{INS: Hardest topics for automatic systems}
\label{ins.hard.topics}
\end{center}
\end{figure}

Figures \ref{ins.easy.topics} and \ref{ins.hard.topics} show the easiest and hardest topics, calculated by the number of runs which scored average precision above 0.06 and below 0.06 respectively. These figures show that Shouting was the easiest action to find, these figures also show drinking, sitting on couch, and holding phone to be among the easiest topics to find. Open door \& leave, open door \& enter, and carrying bag are shown to be among the hardest topics to find.

\begin{figure}[htbp]
\begin{center}
\includegraphics[height=2.5in,width=3in,angle=0]{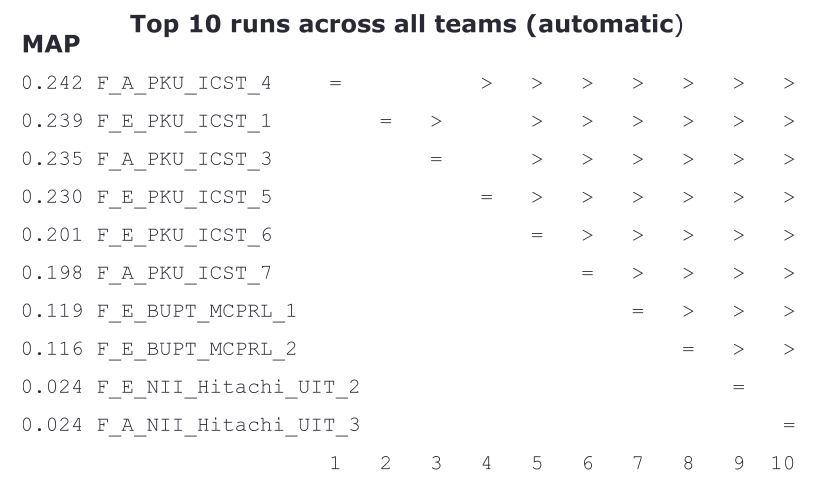}
\caption{INS: Randomization test results for top automatic runs. "E":runs used video examples. "A":runs used image examples only.}
\label{ins.auto.random.test}
\end{center}
\end{figure}

Figure \ref{ins.auto.random.test} documents the raw scores of the top 10
automatic runs and the results of a partial randomization test \cite{manly97} and sheds some light on which differences in ranking are
likely to be statistically significant. One angled bracket indicates p
$<$ 0.05. There are little significant differences between the top runs this year.

\begin{figure}[htbp]
\begin{center}
\includegraphics[height=2.5in,width=3in,angle=0]{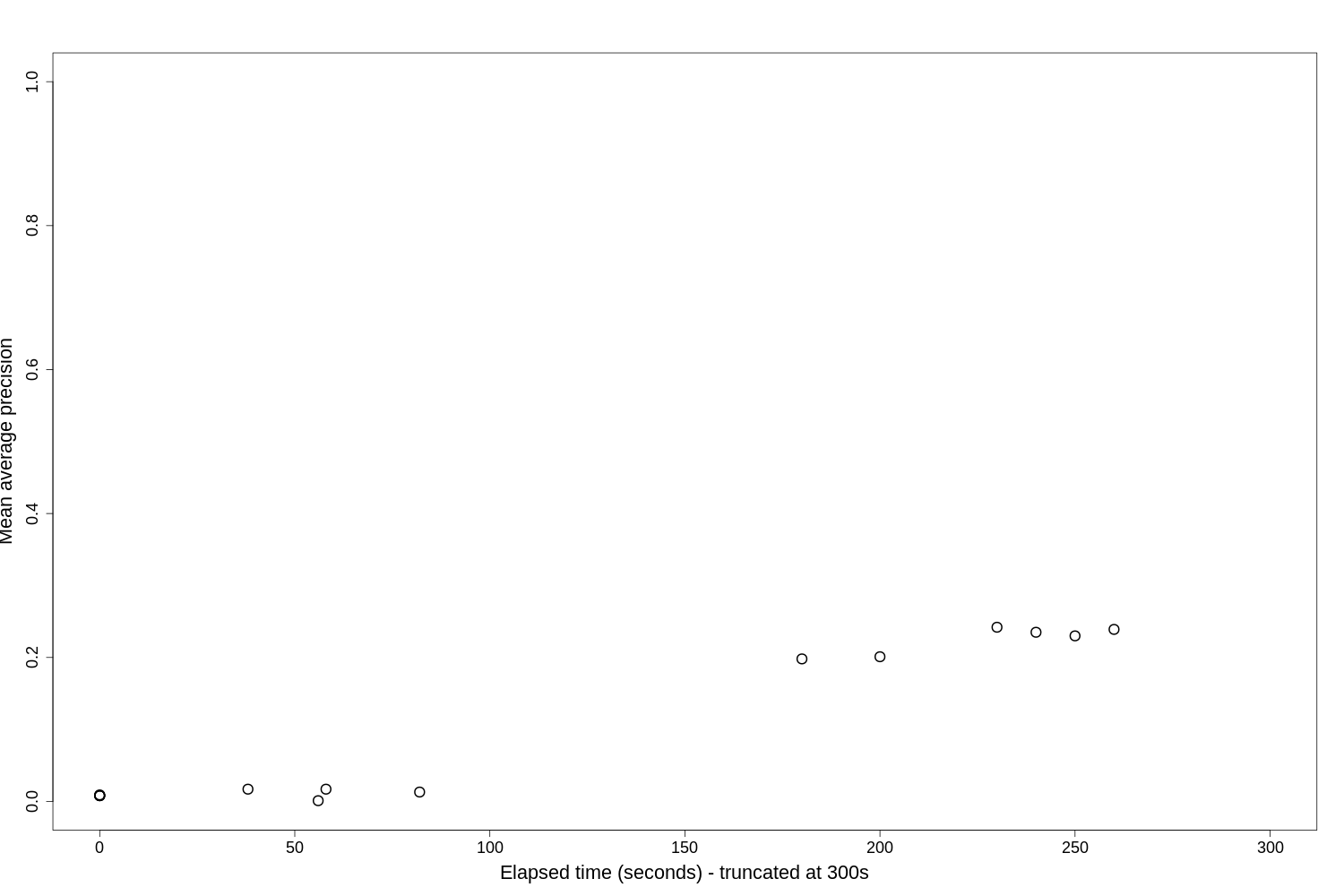}
\caption{INS: Mean average precision versus time for fastest runs}
\label{ins.map.vs.fastest}
\end{center}
\end{figure}

The relationship between the two main measures --- effectiveness (mean average precision)
versus elapsed processing time is depicted in Figure \ref{ins.map.vs.fastest} for the automatic runs with elapsed times less than or equal to 300s. Of those reported times below 300s, we can see that the most accurate systems take longer processing times.

\begin{figure}[htbp]
\begin{center}
\includegraphics[height=2.5in,width=3in,angle=0]{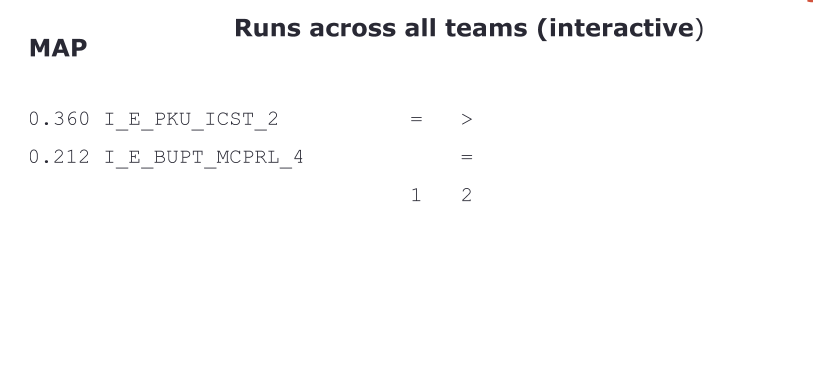}
\caption{INS: Randomization test results for the two interactive runs. "E":runs used video examples. "A":runs used image examples only.}
\label{ins.inter.random.test}
\end{center}
\end{figure}

Figure \ref{ins.inter.random.test} shows the results of a partial randomization test for the 2 submitted interactive runs. Again, one angled bracket indicates p $<$ 0.05 (the probability the result could have been achieved under the null hypothesis, i.e., could be due to chance). This shows much more evidence for a significant difference between the interactive runs than for the top 10 automatic runs.

\begin{figure}[htbp]
\begin{center}
\includegraphics[height=2.5in,width=3in,angle=0]{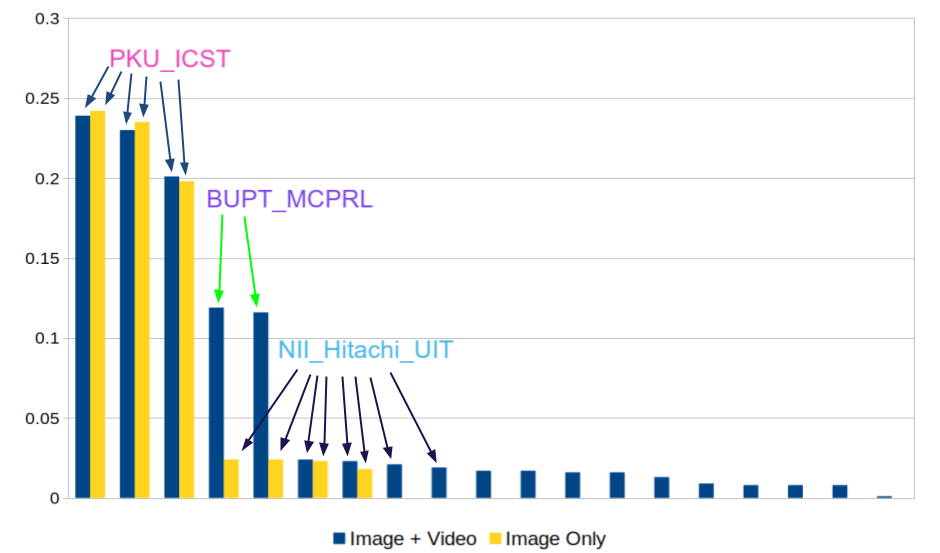}
\caption{INS: Effect of number of topic example images used}
\label{ins.effect.numimages}
\end{center}
\end{figure}

Figure \ref {ins.effect.numimages} shows the relationship between the
two category of runs (images only for training OR video and images) and 
the effectiveness of the runs. These show that far more runs make use of video and image examples than just image examples. Comparing results however for systems making use of both show that there was actually very little difference between results for systems which differed only in the category of runs (images only for training OR video and images).

\begin{figure}[htbp]
\begin{center}
\includegraphics[height=2.5in,width=3in,angle=0]{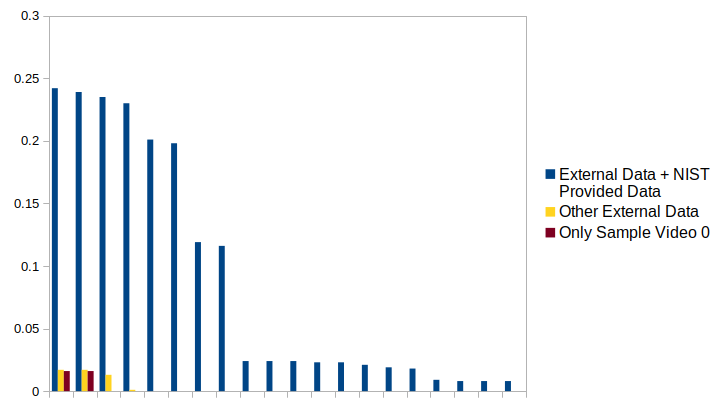}
\caption{INS: Effect of data source used}
\label{ins.effect.datasource}  
\end{center}
\end{figure}

Figure \ref{ins.effect.datasource} shows the effect of the data source used for training, with participants being able to use an external data source instead of or in addition to the NIST provided training data. The use of external data in addition to the NIST provided data gives by far the best results. The use of external data in addition to the NIST provided data is used by the vast majority of participating teams. Results for other external data only and sample video '0' only are similar, however these are way below results for teams which use external data in addition to the NIST provided data, and very few teams use these data sources.

\subsection{Instance Search Observations}
This is the first year the task is using the new query type of person+action. It is the fourth year using the Eastenders dataset. Although there was a slight decrease in number of participants who signed up for the task and the number of finishers, there was a slight increase in the percentage of finishers. 

We should also note that this year a time consuming process was spent trying to get the data agreement set with the donor (BBC) which happened but may have affected number of teams who did not get enough time to work on and finish the task.

The task guidelines were updated for the new updated INS task. This is the first year Human Activity Recognition has been a part of TRECVid. Once again participating teams could use external data instead or in addition to NIST provided data. Results have shown that the use of external data in addition to the NIST provided data consistently gives far better results. However, results also show that the use of external data instead of the NIST provided data, or NIST provided data only, gives quite poor results. Teams could also again make use of video examples or image only examples. Many more teams used video examples in this new task, however results from runs which differed only in the examples used showed very little difference between video examples and image examples only. 

We now summarize the main approaches taken by the different teams. NII\_Hitachi\_UIT used VGGFace2 for face representation. Face is then matched and reranked using cosine similarity. For finding actions, they used VGGish \cite{hershey2017cnn} for audio representation for audio types (laughing, shouting, crying), and for visual types used C3D and semantic features extracted from VGG-1K \cite{simonyan2014very}. Similar to person search, matching and reranking are then applied using cosine similarity. Two computed similarity scores of person and action are then fused for final rank list. 

PKU\_ICST employed four aspects for action specific recognition: frame-level action recognition, video-level action recognition (trained using Kinetics-400), object detection (pre-trained on MS-COCO) and facial expression recognition. They finally computed the average value of the prediction scores of a shot as the final prediction score \textit{ActScore}. For person specific recognition they used query augmentation by super resolution, face recognition with two deep models and top N query expansion. They then employed a score fusion strategy to mine common information from action specific and person specific recognition.

WHU\_NERCMS used two schemes. For the first scheme they retrieved person and action respectively first and then fused them together. For person retrieval they adopted a face recognition model to get person score. For action they adopted 3D convolutional networks to extract spatiotemporal features from videos and measured similarity with queries to get action scores. For fusion they exploited weighting based and person identity based filter to combine results. For the second scheme they retrieved and track specific people and then retrieve their actions. They adopted face recognition to determine the face ID of all characters and bind track ID of the track detected by the object tracking first. They then adopted action recognition of consecutively tracked person target frames to get Action ID, so that each action of each character in all clips has been identified and recorded. Finally, they used specific action of the specific character that the task needs to get final results.

The approach of the Inf teams was as follows: For person search they used MTCNN model to detect faces from frames. Cropped faces are then fed to face recognizer VGG-Face2 for feature selection. They used cosine similarity to measure similarity between queries and retrieved samples. For action search they used Faster-RCNN model pre-trained on MSCOCO dataset for person detection. Proposals were expanded 15\% to the periphery to include actions and objects completely. Tracklets for each person were generated by DeepSort. Fine tune RGB benchmark of I3D model on the combination Charades dataset and offered video shots to extract the features of tracklets. Cosine similarity was then used for action ranking. They used three re-ranking methods: Person search based: Use person ranking to re-rank the action search rank list
Action-based search, Fusion-based: re-ranked by the average similarities between person search and action search. 

BUPT\_MCPRL used the following scheme: They adopted a multi task CNN model, extracted face features based on dlib to get 128-dim face representation and conducted cosine distance between queries and detected persons. For instance retrieval they divided instances into three categories: emotion related, human object interactions and general actions. Emotion Related: They used crying, laughing and shouting as sad, happy or angry - emotion recognition models based on VGG-19 networks taking FER-2013 and CK+ as main training set. For Human-object interactions, they explored dependencies between semantic objects and human keypoints using object detection and pose estimation models. Human bounding boxes were fed into HRNet to estimate human pose. They calculated distance between object location and target persons interactive keypoint. This was used for holding glass, holding phone, carrying bag etc. For general action retrieval: kissing, walking, hugging, they used action detection models: spatio-temporal networks to extract video representation, use ECO as basic network for feature extraction, feed videos in parallel in different frame rates into ECO to extract video representation. Also they used pose-based action detection models to extract video features. They proposed new pose representation by using both absolute and relative positions of pose, they encoded into two feature maps, and they constructed light CNN trained on JHMDB datasets to classify pose representations.

HSMW\_TUC extended a heterogeneous system that enables the identification performance for the recognition and localization of individuals and their activities by heuristically combining several state-of-the-art activity recognition, object recognition and classification frameworks. In their first approach, which deals with the recognition of complex activities of persons or objects, they also integrated state-of-the-art neural network object recognition and classification frames to extract boundary frames from prominent regions or objects that can be used for further processing. Basic tracking of objects detected by bounding boxes requires special algorithmic or feature-driven handling to include statistical correlations between frames.

For detailed information about the approaches and results for individual teams' performance and runs, the reader should see the various site reports \cite{tv19pubs} in the online workshop notebook proceedings.

\subsection{Instance Search Conclusions}
This was the first year of the updated Instance Search task in which queries comprised of a specific person doing a specific action. The action recognition part of the task made this task a much more difficult problem than before, with maximum and average results far below those of previous years for the specific person in a specific location queries. 

There were a total of 6 finishers out of 12 participating teams in this years task. All 6 finishers submitted notebook papers. 3 teams submitted runs for the progress queries to be evaluated in subsequent years in order to measure the progress teams make in the task over the next 3 years.

\section{Activities in Extended Video}
In 2018, NIST TRECVID Activities in Extended Video (ActEV) series was initiated to support the Intelligence Advanced Research Projects Activity (IARPA) Deep Intermodal Video Analytics (DIVA) Program. ActEV is an extension of the TRECVID Surveillance Event Detection (SED) \cite{TrecVIDSed} evaluations where systems only detected and temporally localized activities. The ActEV series are designed to accelerate development of robust automatic activity detection in multi\-camera views for forensic and real-time alerting applications in mind. The previous TRECVID 2018 ActEV (ActEV18) evaluated system detection performance on 12 activities for the self-reported evaluation and 19 activities for the leaderboard evaluation using the VIRAT V1 dataset \cite{TrecVIDActev18}. 
For the self-reported evaluation, the participants ran their software on their hardware and configurations and submitted the system output
with the defined format to the NIST scoring server. For the leaderboard evaluation, the participants submitted their runnable system to the NIST scoring server, which was independently evaluated on the sequestered data using the NIST hardware.
\par The ActEV18 evaluation addressed the two different tasks: 1) identify a target activity along with the time span of the activity (AD: activity detection), 2) detect objects associated with the activity occurrence (AOD: activity and object detection). 
\par For the TRECVID 2019 ActEV (ActEV19) evaluation, we primarily focused on the 18 activities and increased the number of instances for each activity. ActEV19 included the test set from both VIRAT V1 and V2 datasets and the systems were evaluated on the activity detection (AD) task only. 
\par
\begin{figure}[htb]
\begin{centering}
\includegraphics[width=3.16667in,height=1.60000in]{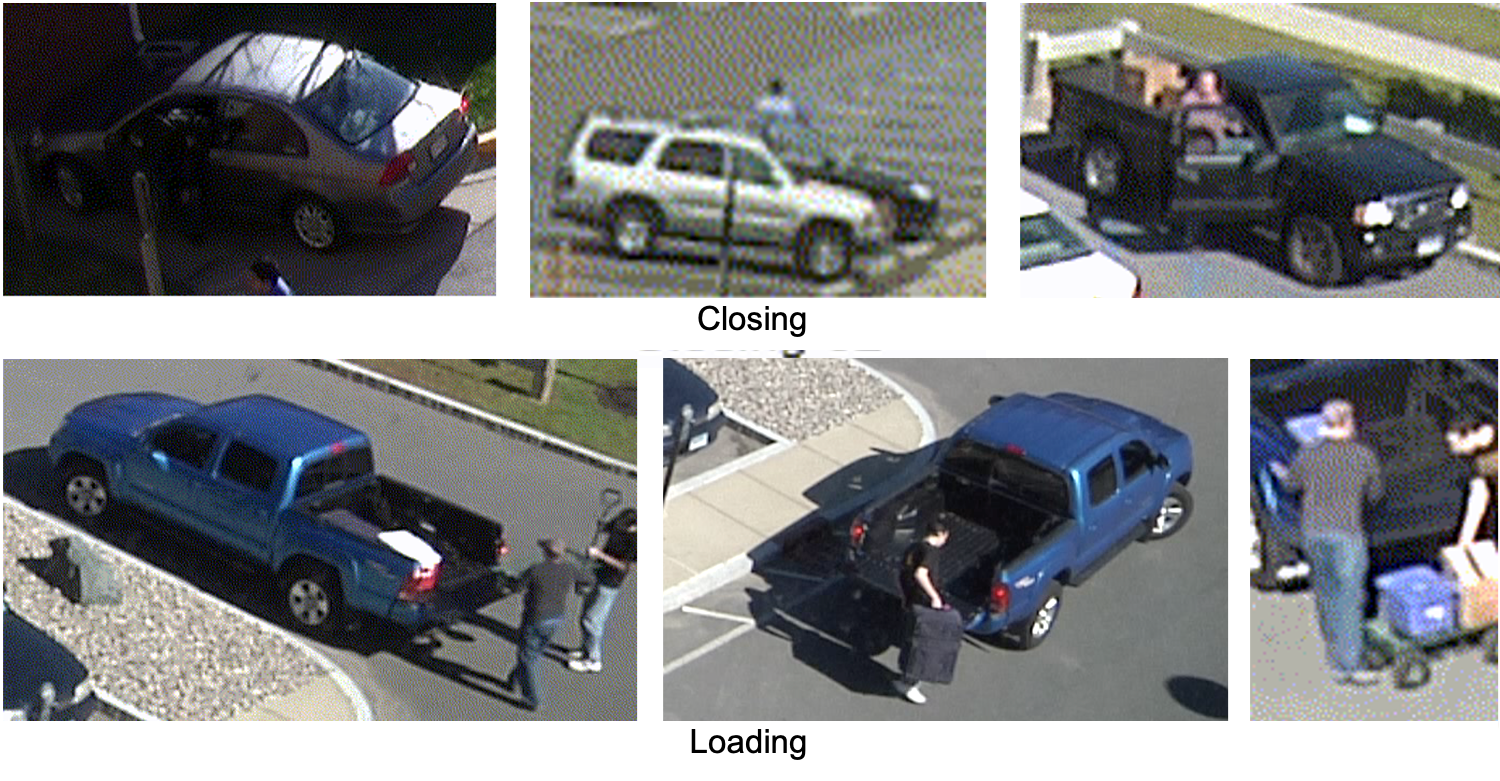}
\caption{Example of activities for ActEV series. IRB (Institutional Review Board): 00000755}
\label{fig_actev:7}
\end{centering}
\end{figure}

 Figure \ref{fig_actev:7} illustrates an example of representative activities that were used in the ActEV series. The evaluation primarily targeted on the forensic analysis that processes the full corpus prior to returning a list of detected activity instances. A total of 9 different organizations participated in this year evaluation (ActEV19) and over 256 different algorithms were submitted.

\par In this paper, we first discuss task and dataset used and introduce a new metric to evaluate algorithm performance. In addition, we present the results for the TRECVID19 ActEV submissions and discuss observations and conclusions.

\subsection {Task and Dataset}
\par In the ActEV19 leaderboard evaluation, we addressed activity detection (AD) task for detecting and localizing activities; a system required to automatically detects and temporally localizes all instances of the activity. For a system-identified activity instance to be evaluated as correct, the type of activity should be correct, and the temporal overlap should fall within a minimal requirement. The type of the ActEV19 challenge was called an open leaderboard evaluation; the challenge participants should run their software on their systems and configurations and submit the defined system output to the NIST Scoring Server. The leaderboard evaluation should submit a system to report activities that visibly occur in a single-camera video by identifying the video file, the frame span (the start and end frames) of the activity instance, and the presence confidence value indicating the system’s “confidence score” how likely the activity is present.

\begin{table}[htb]
\caption{A list of 18 activities on the VIRAT dataset and their associated number of instances for the train and validation sets}
\label{table_actev:1}
\begin{centering}
\begin{tabular}{ m{4.2cm} | m{0.8cm}| m{1.5cm} } 
\hline
Activity Type & Train & Validation \\ \hline
Closing & 126 & 132 \\ 
Closing\_trunk & 31 & 21 \\ 
Entering & 70 & 71 \\ 
Exiting & 72 & 65 \\ 
Loading & 38 & 37 \\ 
Open\_Trunk & 35 & 22 \\ 
Opening & 125 & 127 \\ 
Transport\_HeavyCarry & 45 & 31 \\ 
Unloading & 44 & 32 \\ 
Vehicle\_turning\_left & 152 & 133 \\ 
Vehicle\_turning\_right & 165 & 137 \\ 
Vehicle\_u\_turn & 13 & 8 \\ 
Pull & 21 & 22 \\
Riding & 21 & 22 \\
Talking & 67 & 41 \\
Activity\_carrying & 364 & 237 \\
Specialized\_talking\_phone & 16 & 17 \\
Specialized\_texting\_phone & 20 & 5 \\
\hline
\end{tabular}
\end{centering}
\end{table}

\par For this evaluation, we used 18 activities from the VIRAT dataset \cite{oh2011large} and the activities were annotated by Kitware, Inc. The VIRAT dataset consisted of 29 video hours and more than 23 activity types. A total of 10 video hours were annotated for the test set across 18 activities. The detailed definition of each activity and evaluation requirments are described in the evaluation plan \cite{TrecVIDActev19}. 
\par Table \ref{table_actev:1} lists the number of instances for each activity for the train and validation sets. Due to ongoing evaluations, the test sets are not included in the table. The numbers of instances are not balanced across activities, which may affect the system performance results. 

\subsection {Measures}
\begin{figure*}[htbp]
\begin{centering}
\includegraphics[width=5.56667in,height=1.2000in]{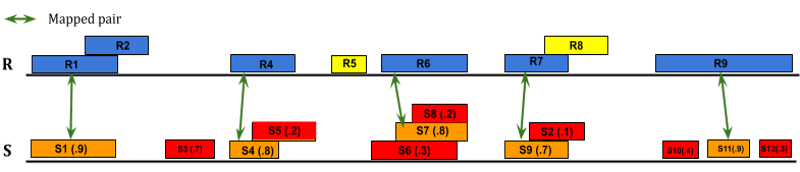}
\caption{Illustration of activity instance alignment and \(P_{miss}\) calculation (\(R\) is the reference instances and \(S\) is the system instances. In \(S\), the first number indicates instance id and the second indicates presence confidence score. For example, \(S1 (.9)\) represents the instance \(S1\) with corresponding confidence score .9. Green arrows indicate aligned instances between \(R\) and \(S\))}
\label{fig_actev:1}
\end{centering}
\end{figure*}

In this evaluation, an activity is defined as "one or more people performing a specified movement or interacting with an object or group of objects (including driving and flying)", while an instance indicates an occurrence (time span of the start and end frames) in associated with the activity.
For the past year TRECVID ActEV18, the primary metric was instance-based measures for both missed detections and false alarms (as illustrated in Figure \ref{fig_actev:1}. The metric evaluated how accurately the system detected the instance occurrences of the activity. 

 As shown in Figure \ref{fig_actev:1}, the detection confusion matrix are calculated with alignment between reference and system output on the target activity instances; Correct Detection (\(CD\)) indicates that the reference and system output instances are correctly mapped (instances marked in blue). Missed Detection (\(MD\)) indicates that an instance in the reference has no correspondence in the system output (instances marked in yellow) while False Alarm (\(FA\)) indicates that an instance in the system output has no correspondence in the reference (instances marked in red). After calculating the confusion matrix, we summarize system performance: for each instance, a system output provides a confidence score that indicates how likely the instance is associated with the target activity. The confidence score can be used as a decision threshold.
 \par In the last year evaluation, a probability of missed detections (\(P_{\text{miss}}\)) and a rate of false alarms (\(R_{\text{FA}})\) were used and computed at a given decision threshold:
\[P_{\text{miss}}(\tau)\  = \frac{N_{\text{MD}}(\tau)}{N_{\text{TrueInstance}}}\]
\[\text{R}_{\text{FA}}(\tau)\  = \frac{N_{\text{FA}}(\tau)}{\text{VideoDurInMinutes}}\]\

where \(N_{\text{MD\ }}(\tau)\) is the number of missed detections at the threshold \(\tau\) , \(N_{\text{FA\ }}(\tau)\) is the number of false alarms, and \emph{VideoDurInMinutes} is number of minutes of video. \(N_{\text{TrueInstance}}\) is the number of reference instances annotated in the sequence. Lastly, the Detection Error Tradeoff (DET) curve \cite{martinDET} is used to visualize system performance. For the TRECVID ActEV18 challenges last year, we evaluated algorithm performance on the operating points;
\(P_{\text{miss}}\text{\ at\ }R_{\text{FA}} = 0.15\) and
\(P_{\text{miss}}\text{\ at\ }R_{\text{FA}} = 1\).

To understand system performance better and to be more relevant to the user cases, for ActEV19, we used the normalized, partial area under the DET curve (\(nAUDC\)) from 0 to a fixed time-based false alarm (\(T_{fa}\)) to evaluate algorithm performance.
The partial area under DET curve is computed separately for each activity over all videos in the test collection and then is normalized to the range [0, 1] by dividing by the maximum partial area \(nAUDC_a=0\) is a perfect score. The \(nAUDC_a\) is defined as:

\[nAUDC_{a} = \frac{1}{a}\int_{x=0}^{a} P_{miss}(x)dx,  x=T_{fa}\]

where \(x\) is integrated over the set of \(T_{fa}\) values. The instance-based probability of missed detections \(P_{miss}\) is defined as:

\[P_{miss} (x) = \frac{N_{md}(x)}{N_{TrueInstance}}\]

where \(N_{md}(x)\) is the number of missed detections at the presence confidence threshold that result in \(T_{fa}=x\) (see the below equation for the details). \(N_{TrueInstance}\) is the number of true instances in the sequence of reference.
\par
\begin{figure*}[htbp]
\begin{centering}
\includegraphics[width=5.56667in,height=2.70000in]{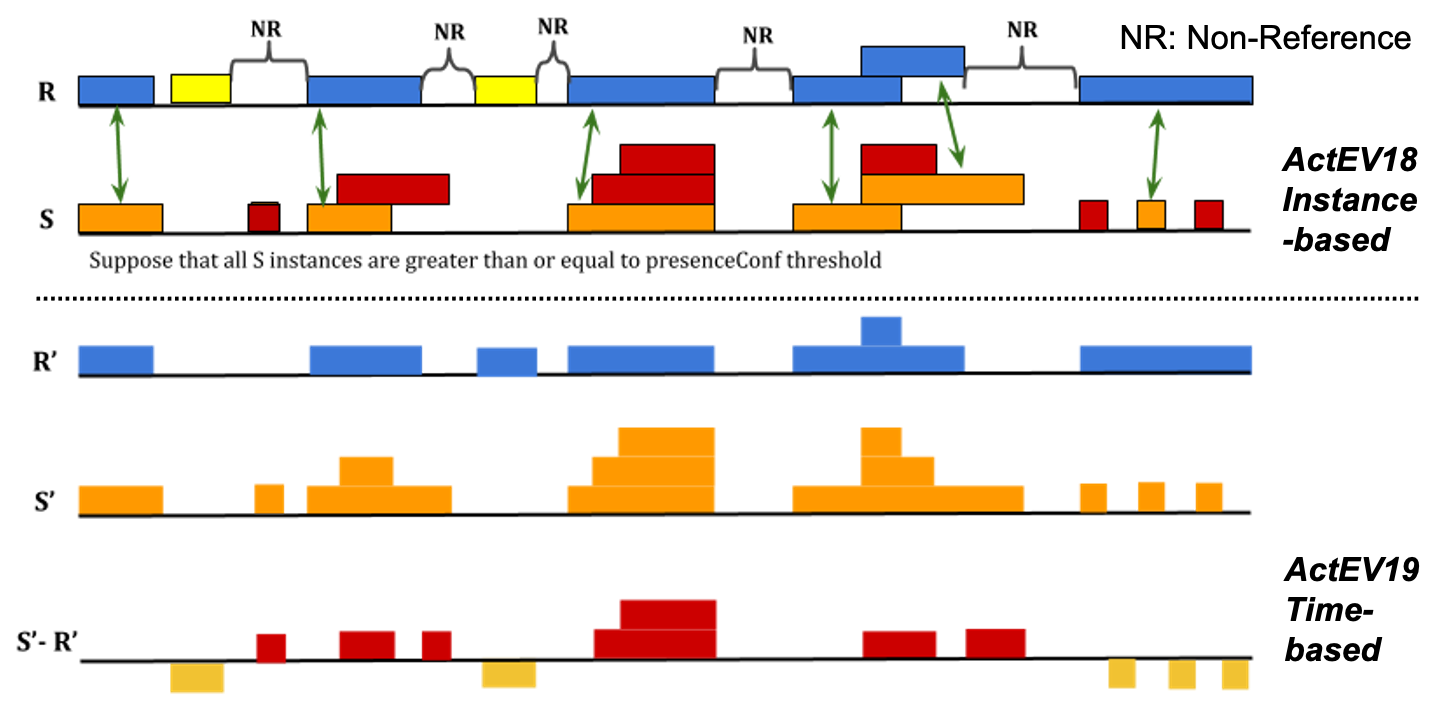}
\caption{Comparison of instance-based and time-based false alarms. \(R\) is the reference instances and \(S\) is the system instances. \(R^\prime\) is the histogram of the count of reference instances and \(S^\prime\) is the histogram of the count of system instances for the target activity. \(S\) shows a depiction of instance-based false alarms while \({S^\prime}-{R^\prime}\) illustrates time-based false alarms as marked in red.}
\label{fig_actev:2}
\end{centering}
\end{figure*}

 The time-based false alarm \(T_{fa}\) is defined as: 
\[T_{fa} = \frac{1}{NR} {\sum_{i=1}^{N_{frames}}} {\max(0, {S_i^\prime}-{R_i^\prime})}\]

where \(N_{frames}\) is the duration of the video and \(NR\) is the non-reference duration; the duration of the video without the target activity occurring. \({S_i^\prime}\) is the total count of system instances for frame \(i\) while \({R_i^\prime}\) is the total count of reference instances for frame \(i\). The detailed calculation of \(T_{fa}\) is illustrated in Figure \ref{fig_actev:2}.
\par The non-reference duration (NR) of the video where no target activities occurs is computed by constructing a time signal composed of the complement of the union of the reference instances duration. \(R\) is the reference instances and \(S\) is the system instances. \(R^\prime\) is the histogram of the count of reference instances and \(S^\prime\) is the histogram of the count of system instances for the target activity. \(R^\prime\) and \(S^\prime\) both have \(N_{frames}\) bins, thus \(R_i^\prime\) is the value of the \(i^{th}\) bin \(R^\prime\) while \(S_i^\prime\) is the value of the \(i^{th}\) bin \(S^\prime\).
\(S^\prime\) is the total count of system instances in frame \(i\) and \(R^\prime\) is the total count of reference instances in frame \(i\).  
False alarm time is computed by summing over positive difference of \({S^\prime}-{R^\prime}\)(shown in red in Figure \ref{fig_actev:2}); that is the duration of falsely detected system instances. This value is normalized by the non-reference duration of the video to provide the \(T_fa\) value in Equation above.

\begin{figure}[hbt]
\begin{centering}
\includegraphics[width=3.16667in,height=1.60000in]{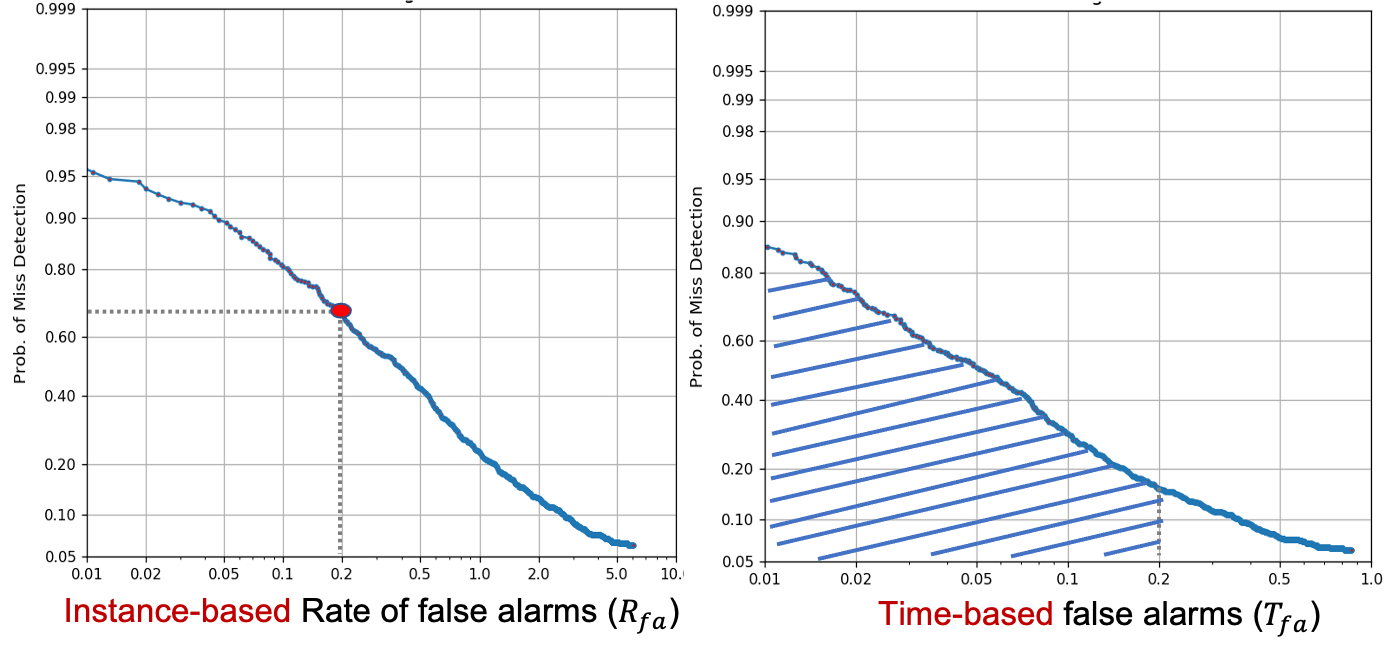}
\caption{Comparison of ActEV18 (\(R_{fa}\)) and ActEV19 (\(T_{fa}\)) measures using the Detection Error Tradeoff (DET) curves}
\label{fig_actev:3}
\end{centering}
\end{figure}

Figure \ref{fig_actev:3} shows visual representations of the major differences between the ActEV18 and ActEV19 metrics. For the ActEV18 metric, we used Instance-based Rate of false alarms and system performance was evaluated at the specific operating point as illustrated in the left DET. For the ActEV19 metric, we used Time-based false alarms and calculated \(nAUDC\) from \(T_{fa}\) 0 to 0.2.

\subsection{ActEV Results}

A total of 9 teams from academia and industry participated in the ActEV19 evaluation. Each participant was allowed to submit multiple system outputs. From the 9 teams, we have a total of 256 submissions as of the deadline November 1, 2019. Table \ref{table_actev:4} lists the participants and their system performance measure \(nAUDC\) which was identified as the best system per team.

\begin{table*}[t]
\caption{Summary of participants information and their \(nAUDC\) values. Each team was allowed to have multiple submissions. Table below lists the best system result per site from multiple submissions.}
\label{table_actev:4} 
\begin{centering}
\begin{tabular}{ m{3cm} | m{10cm}| m{1.2cm} } 
\hline
Team               & Organization     & nAUDC \\
\hline
BUPT-MCPRL         & Beijing University of Posts and Telecommunications, China                                                   & 0.524 \\
Fraunhofer IOSB    & ĘFraunhofer Institute, Germany                                                                              & 0.827 \\
HSMW\_TUC          & University of Applied Sciences Mittweida and Chemnitz University of Technology, Germany                     & 0.941 \\
MKLab (ITI\_CERTH) & Information Technologies Institute, Greece                                                                  & 0.964 \\
MUDSML             & Monash University, Australia and Carnegie Mellon University, USA                                            & 0.484 \\
NII\_Hitachi\_UIT  & National Institute of Informatics, Japan Hitachi, Ltd., Japan University of Information Technology, Vietnam & 0.599 \\
NTT\_CQUPT         & NTT company \& Chongqing University of Posts and Telecommunications, China                                  & 0.601 \\
UCF                & University of Central Florida, USA                                                                          & 0.491 \\
vireoJD-MM         & City University of Hong Kong and JD AI Research, China  & 0.601 \\
\hline
\end{tabular}
\end{centering}
\end{table*}

\begin{figure}[hbt]
\begin{centering}
\includegraphics[width=3.16667in,height=2.00000in]{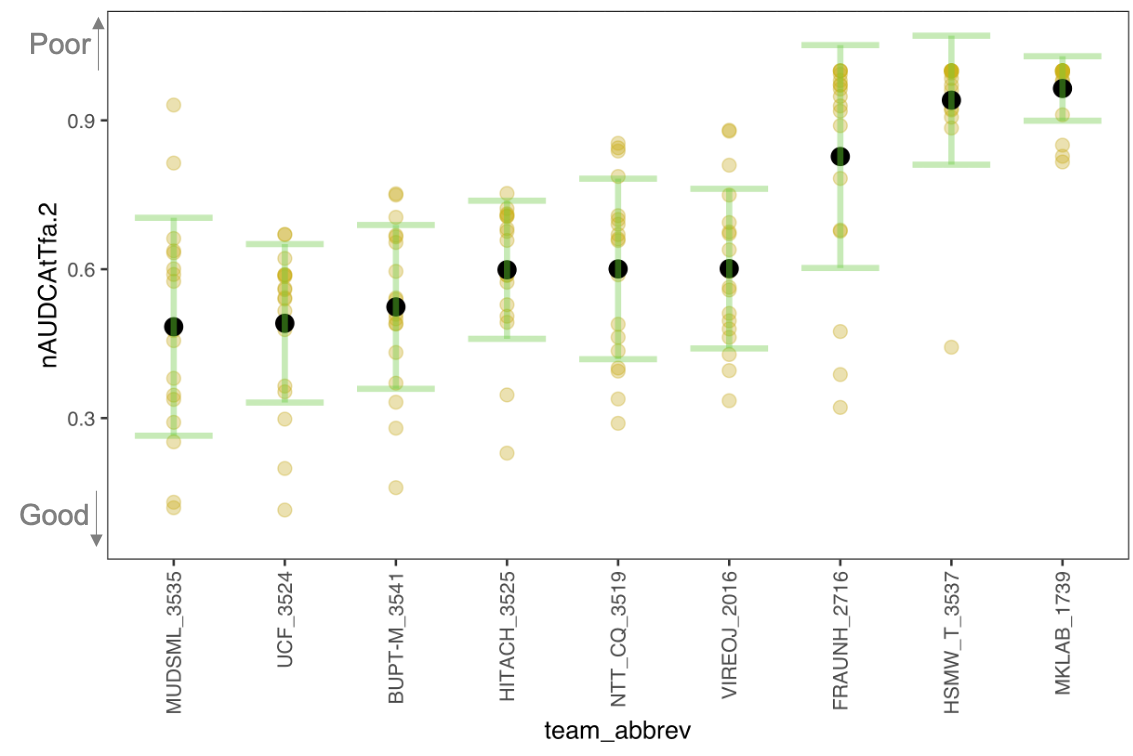}
\caption{Comparison of system performance across teams. The x-axis is the team name and the y-axis is \(nAUDC\) value and a lower value is considered as a better performance. The green dots represent average performance of the 18 different activities. The black dot indicates the mean value across the 18 activities and the horizontal bar represents standard deviation.}
\label{fig_actev:4}
\end{centering}
\end{figure}

Figure \ref{fig_actev:4} illustrates the ranking of the 9 systems ordered by  \(nAUDC\) values. The result shows that MUDSML achieved the lowest error rate (\(nAUDC\): 0.484) followed by UCF (\(nAUDC\): 0.491). We also observe that some systems have a larger error bar across the 18 different activities.
For comparison purpose, Table \ref{table_actev:6} summarizes the leaderboard evaluation results from both ActEV18 and ActEV19. Out of the 9 teams in current year participants, only 4 teams participated in both ActEV18 and ActEV19 evaluations. Note that, for this comparison, we had a slightly different dataset and the number of activities, while using the same scoring protocol and performance measure (namely, PR.15: \(P_{miss} at R_{FA} = 0.15\). 
\par We took out the activity "interact" in the ActEV19 evaluation due to its activity definition ambiguity.

\begin{table}[htb]
\caption{Comparison of the ActEV18 and ActEV19 results. Since \(P_{miss} at R_{FA} = 0.15\) was a primary measure for ActEV18, the ActEV19 column lists both \(P_{miss}\) at \(R_{FA} = 0.15\) (PR.15) and \(nAUDC\) for comparison purpose.}
\label{table_actev:6}
\begin{centering}
\begin{tabular}{ m{2.8cm} | m{1.2cm}| m{1cm} | m{1cm}} 
\hline
\multicolumn{1}{c}{\multirow{3}{*}{Team}} & \multicolumn{1}{|c|}{ActEV18} & \multicolumn{2}{c}{ActEV19}                            \\ \cline{2-4} 
\multicolumn{1}{c}{}                      & \multicolumn{1}{|c|}{LB (19)} & \multicolumn{2}{c}{LB (18)}                            \\ \cline{2-4} 
\multicolumn{1}{c}{}                      & \multicolumn{1}{|c|}{PR.15}   & \multicolumn{1}{c|}{PR.15} & \multicolumn{1}{c}{nAUDC} \\ \hline
UCF                & 0.733   & 0.68    & 0.491 \\
MUDSML (INF)       & 0.844   & 0.789   & 0.484 \\
HSMW\_TUC          & x       & 0.951   & 0.941 \\
BUPT-MCPRL         & 0.749   & 0.736   & 0.524 \\
MKLab              & x       & 0.968   & 0.964 \\
NII\_Hitachi\_UIT  & 0.925   & 0.819   & 0.599 \\
Fraunhofer IOSB    & x       & 0.849   & 0.827 \\
NTT\_CQUPT         & x       & 0.878   & 0.601 \\
vireoJD-MM         & x       & 0.714   & 0.601 \\
\hline
\end{tabular}
\end{centering}
\end{table}

\begin{figure}[htb]
\begin{centering}
\includegraphics[width=3.16667in,height=2.00000in]{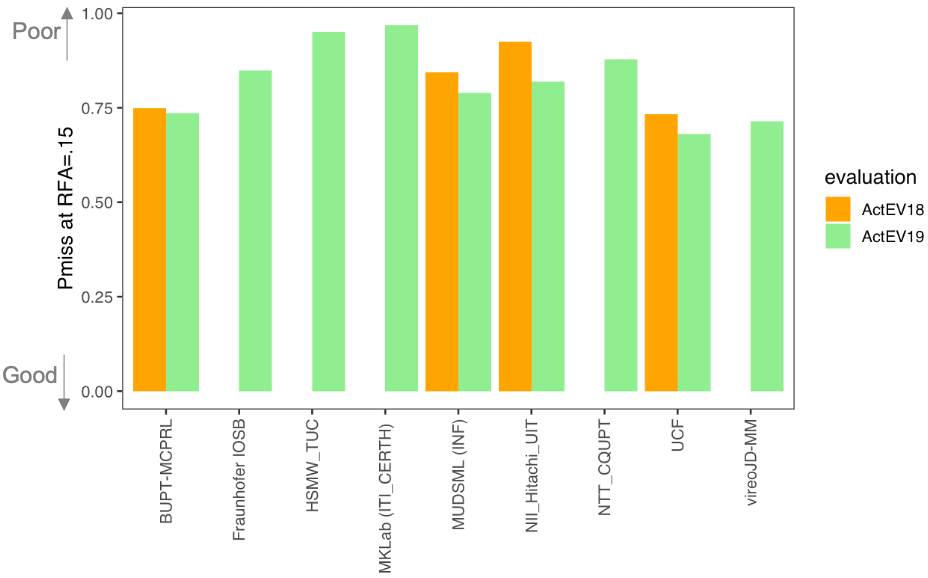}
\caption{Comparison of the ActEV18 and ActEV19 results. The x-axis is the team name and the y-axis is \(P_{miss} at R_{FA} = 0.15\) value and a lower value is considered as a better performance. The green bars represent performance of the ActEV19 leaderboard results while the orange bars indicate the leaderboard results from ActEV18.}
\label{fig_actev:16}
\end{centering}
\end{figure}

Figure \ref{fig_actev:16} shows that all the 4 participants improved their system performance from last year for the leaderboard evaluations. The relative error rates were reduced \textasciitilde12\% for NII\_Hitachi\_UIT and \textasciitilde7\% for UCF and MUDSML.

\par
\begin{figure}[htb]
\begin{centering}
\includegraphics[width=3.16667in,height=2.80000in]{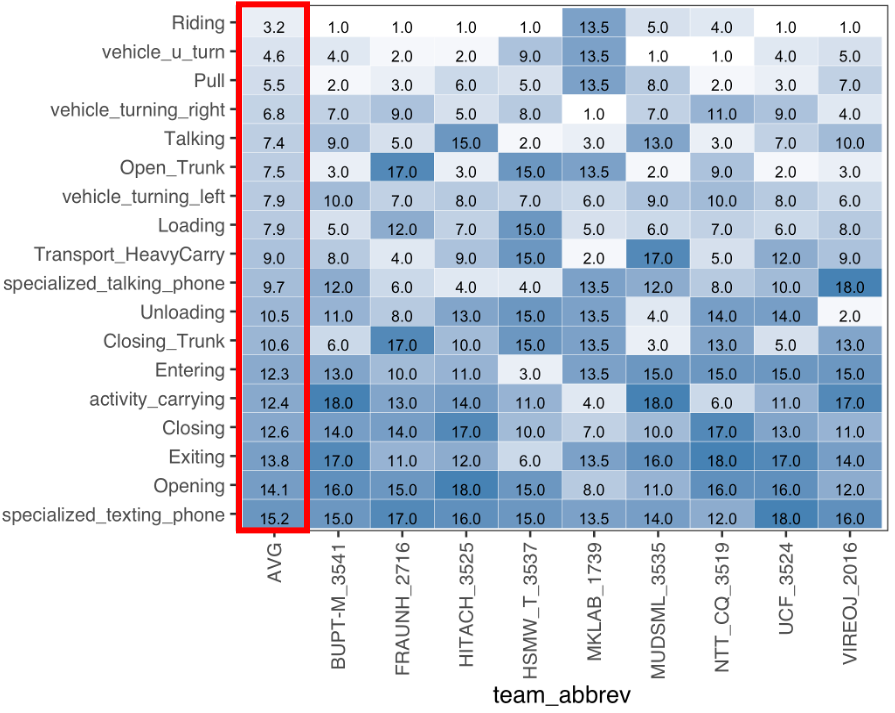}
\caption{Summary of activity detection difficulty. The x-axis denotes systems and their average ranking (AVG). The y-axis indicates the 18 activities. The numbers on the matrix represent the ranking of 18 activities per system. The AVG column marked in red is the average value of system performance across the 9 teams.}
\label{fig_actev:5}
\end{centering}
\end{figure}

\begin{figure}[htb]
\begin{centering}
\includegraphics[width=3.16667in,height=2.50000in]{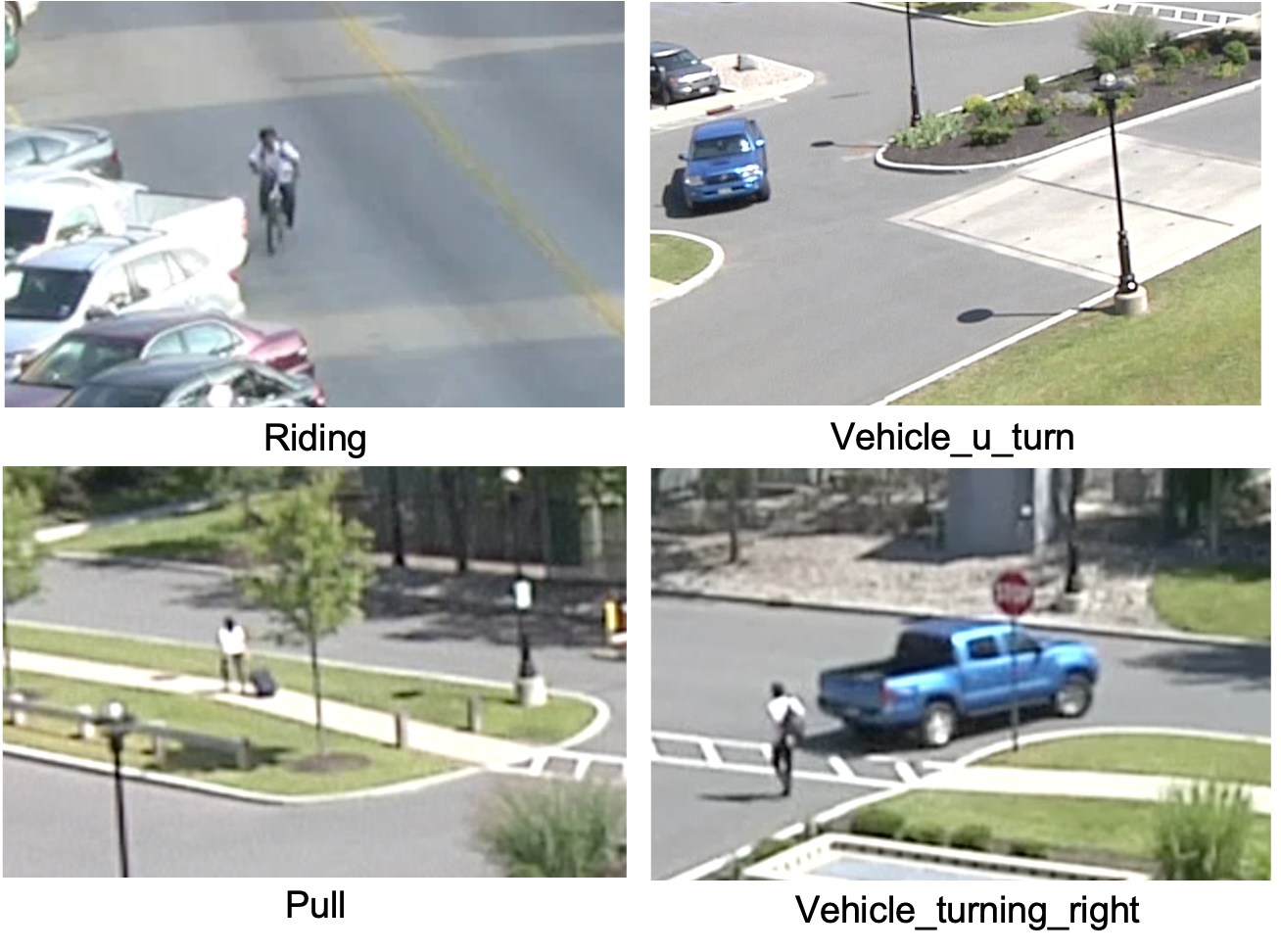}
\caption{Example of the four activities that are easier to detect. IRB \#: 00000755}
\label{fig_actev:6}
\end{centering}
\end{figure}

 To determine activity detection difficulty, the activities are characterized by performance of system outputs. In Figure \ref{fig_actev:5}, we observe that riding, vehicle\_u\_turn, pull, and vehicle\_turn\_right activities are easier to detect compared to the rest of the other activities. Figure \ref{fig_actev:6} shows examples of those top-performed activities.

\subsection{Summary}\label{ActEV_Conclusion}
In this section, we presented the TRECVID ActEV19 evaluation task, new performance metric and results for human activity detection. We primarily focused on the activity detection task only and the time-base false alarms were used to have a better understanding of system's behavior and to be more relevant to the user cases. The proposed metric was compared to the instance-based false alarms that were used in the last year evaluation (ActEV18).
Nine teams participated in the ActEV19 evaluation and a total of the 256 systems were submitted. We provided a ranked list of system performance and examined the level of activity difficulty in detection using the submissions selected as the best\-performed system per team.


\section{Video to Text Description}
\begin{table*}[htbp] 
\begin{center}

\begin{tabular}[htbp]{|c|c|c|}
\hline
&Matching \& Ranking (11 Runs) & Description Generation (30 Runs)\\ \hline\hline
IMFD\_IMPRESEE& X & X\\ \hline
KSLAB& X & X\\ \hline
RUCMM& X & X\\ \hline
RUC\_AIM3& X & X\\ \hline
EURECOM\_MeMAD&  & X\\ \hline
FDU& & X\\ \hline
INSIGHT\_DCU& & X\\ \hline
KU\_ISPL&  & X\\ \hline
PICSOM&  & X\\ \hline
UTS\_ISA&  & X\\ \hline

\end{tabular}
\caption{VTT: List of teams participating in each of the subtasks. Description Generation was a core subtask in 2019.}
\label{tab:vtt_participants}
\end{center}
\end{table*}

Automatic annotation of videos using natural language text descriptions has been a long-standing goal of computer vision. The task involves understanding many concepts such as objects, actions, scenes, person-object relations, the temporal order of events throughout the video, and many others. In recent years there have been major advances in computer vision techniques which enabled researchers to start practical work on solving the challenges posed in automatic video captioning. 

There are many use-case application scenarios which can greatly benefit from the technology, such as video summarization in the form of natural language,  facilitating the searching and browsing of video archives using such descriptions, describing videos as an assistive technology, etc. In addition, learning video interpretation and temporal relations among events in a video will likely contribute to other computer vision tasks, such as prediction of future events from the video. 

The ``Video to Text Description'' (VTT) task was introduced in TRECVID 2016 as a pilot. Since then, there have been substantial improvements in the dataset and evaluation.

For this year, 10 teams participated in the VTT task. There were a total of 11 runs submitted by 4 teams for the matching and ranking subtask, and 30 runs submitted by 10 teams for the description generation subtask. A summary of participating teams is shown in Table~\ref{tab:vtt_participants}.

\subsection{Data}

The VTT data for 2019 consisted of two video sources. 
\begin{itemize}
    \item \textbf{Twitter Vine}: Since the inception of the VTT task, the testing data has comprised of Vine videos. Over 50k Twitter Vine videos have been collected automatically, and each video has a total duration of about 6 seconds. We selected 1044 Vine videos for this year's task. 
    \item \textbf{Flickr}: Flickr video was collected under the Creative Commons License. Videos from this dataset have previously been used for the Instance Search Task at TRECVID. A set of 91 videos was collected, which was divided into 74\,958 segments of about 10 seconds each. A subset of 1010 segments was used for this year's VTT task.
\end{itemize}

A total of 2054 videos were selected and annotated manually by multiple annotators.
An attempt was made to create a diverse dataset by removing any duplicates or similar videos as a preprocessing step. 

\subsubsection{Data Cleaning}
We carried out data preprocessing before the annotation process to ensure a usable dataset. Firstly, we clustered videos based on visual similarity. We used a tool called SOTU \cite{ngo2012sotu}, which uses visual bag of words, to cluster videos with 60\,\% similarity for at least 3 frames. This allowed us to remove any duplicate videos, as well as videos which were very similar visually (e.g., soccer games). However, we learned from previous experience that this automated procedure is not sufficient to create a clean and diverse dataset. For this reason, we manually went through a large set of videos. We used a list of commonly appearing topics that was collected from previous years' data, and filtered videos to ensure that the dataset was not dominated by certain topics. We also removed the following types of videos:
\begin{itemize}
    \item Videos with multiple, unrelated segments that are hard to describe, even for humans.
    \item Any animated videos.
    \item Other videos that may be considered inappropriate or offensive.
\end{itemize}

\begin{table}[htbp]
    \centering
    
    \begin{tabular}{|c|c|}
    \hline
    Annotator  & Avg. Length\\
    \hline\hline
       1  & 12.83 \\
       \hline
       2  & 16.07 \\
       \hline
       3  & 16.49 \\
       \hline
       4  & 17.72 \\
       \hline
       5  & 18.76 \\
       \hline
       6  & 19.55 \\
       \hline
       7  & 20.42 \\
       \hline
       8  & 21.16 \\
       \hline
       9  & 21.73 \\
       \hline
       10  & 22.07\\
       \hline
    \end{tabular}
    \caption{VTT: Average number of words per sentence for all 10 annotators. A large variation is observed between average sentence lengths for the different annotators.}
    \label{tab:avg_gt_length}
\end{table}

\subsubsection{Annotation Process}

The videos were divided amongst 10 annotators, with each video being annotated by exactly 5 people. The annotators were asked to include and combine into 1 sentence, if appropriate and available, four facets of the video they are describing:

\begin{itemize}
\item{\textbf{Who} is the video showing (e.g., concrete objects and beings, kinds of persons, animals, or things)}?
\item{\textbf{What} are the objects and beings doing (generic actions, conditions/state or events)}?
\item{\textbf{Where} is the video taken (e.g., locale, site, place, geographic location, architectural)}?
\item{\textbf{When} is the video taken (e.g., time of day, season)}?
\end{itemize}

Different annotators provide varying amount of detail when describing videos. Some people try to incorporate as much information as possible about the video, whereas others may write more compact sentences. Table~\ref{tab:avg_gt_length} shows the average number of words per sentence for each of the 10 annotators. The average sentence length varies from 12.83 words to 22.07 words, emphasizing the difference in descriptions provided by the annotators. 

Furthermore, the annotators were also asked the following questions for each video:
\begin{itemize}
\item Please rate how difficult it was to describe the video.
\begin{enumerate}
    \item Very Easy
    \item Easy
    \item Medium
    \item Hard
    \item Very Hard
\end{enumerate}
\item How likely is it that other assessors will write similar descriptions for the video?
\begin{enumerate}
    \item Not Likely
    \item Somewhat Likely
    \item Very Likely
\end{enumerate}
\end{itemize}

The average score for the first question was 2.03 (on a scale of 1 to 5), showing that in general the annotators thought the videos were on the easier side to describe. The average score for the second question was 2.51 (on a scale of 1 to 3), meaning that they thought that other people would write a similar description as them for most videos. The two scores are negatively correlated as annotators are more likely to think that other people will come up with similar descriptions for easier videos. The correlation score between the two questions is -0.72.

\begin{figure}[htbp]
  \centering
  \includegraphics[width=1.0\linewidth]{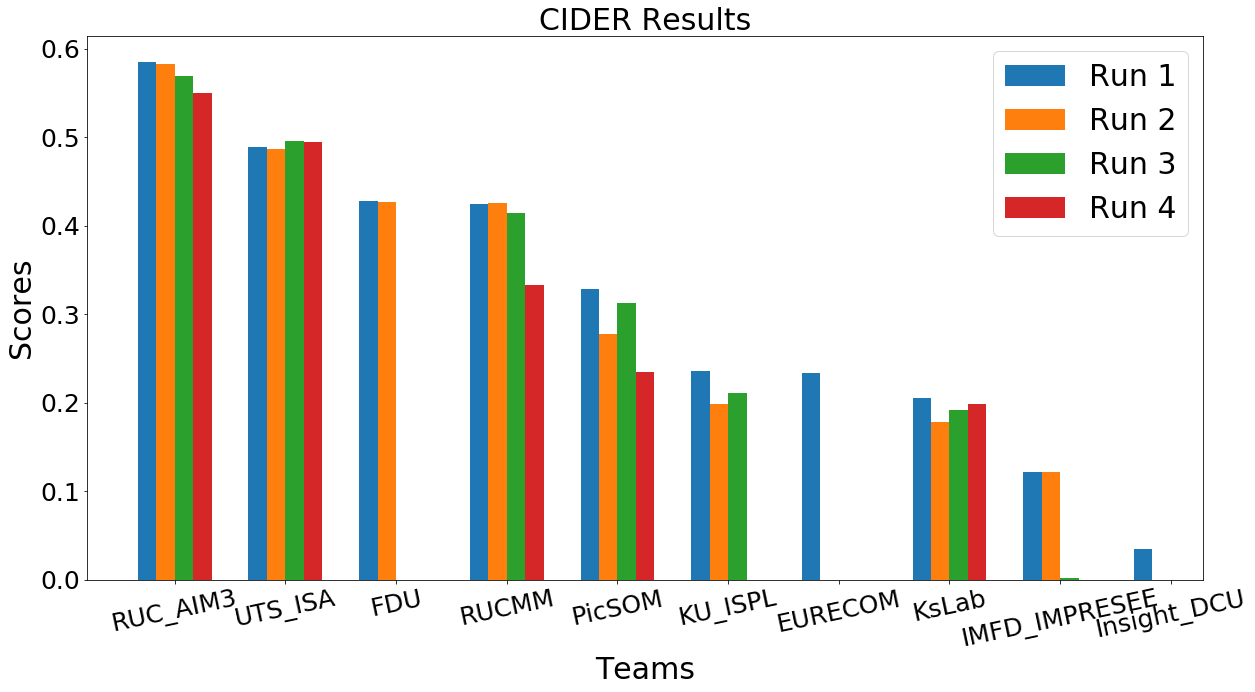}
  \caption{VTT: Comparison of all runs using the CIDEr metric.}
  \label{fig:vtt.cider.results}
\end{figure}

\begin{figure}[htbp]
  \centering
  \includegraphics[width=1.0\linewidth]{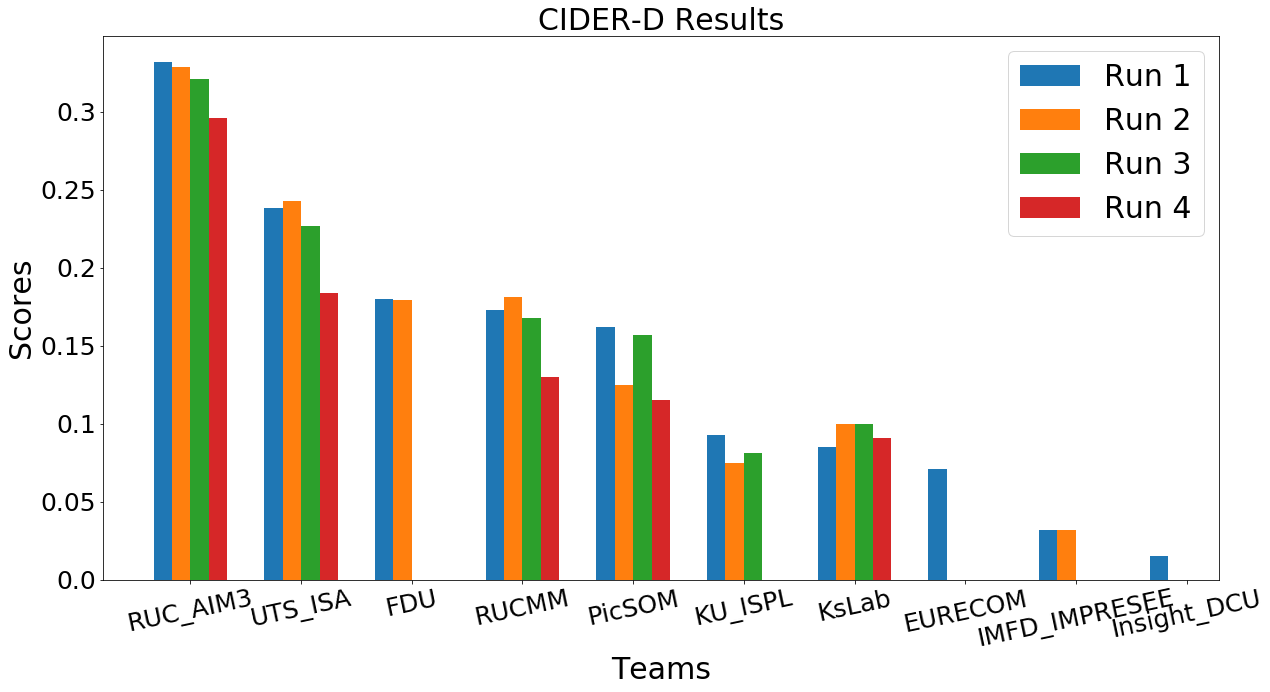}
  \caption{VTT: Comparison of all runs using the CIDEr-D metric.}
  \label{fig:vtt.ciderd.results}
\end{figure}

\begin{figure}[htbp]
  \centering
  \includegraphics[width=1.0\linewidth]{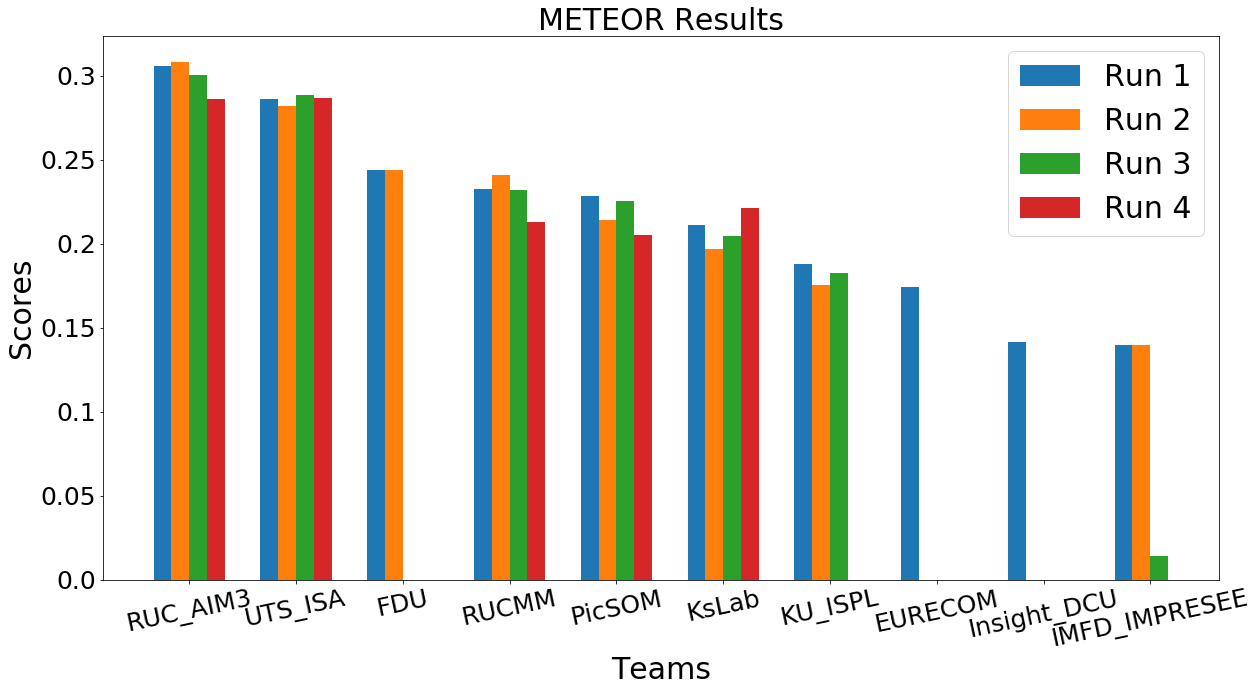}
  \caption{VTT: Comparison of all runs using the METEOR metric.}
  \label{fig:vtt.meteor.results}
\end{figure}

\begin{figure}[htbp]
  \centering
  \includegraphics[width=1.0\linewidth]{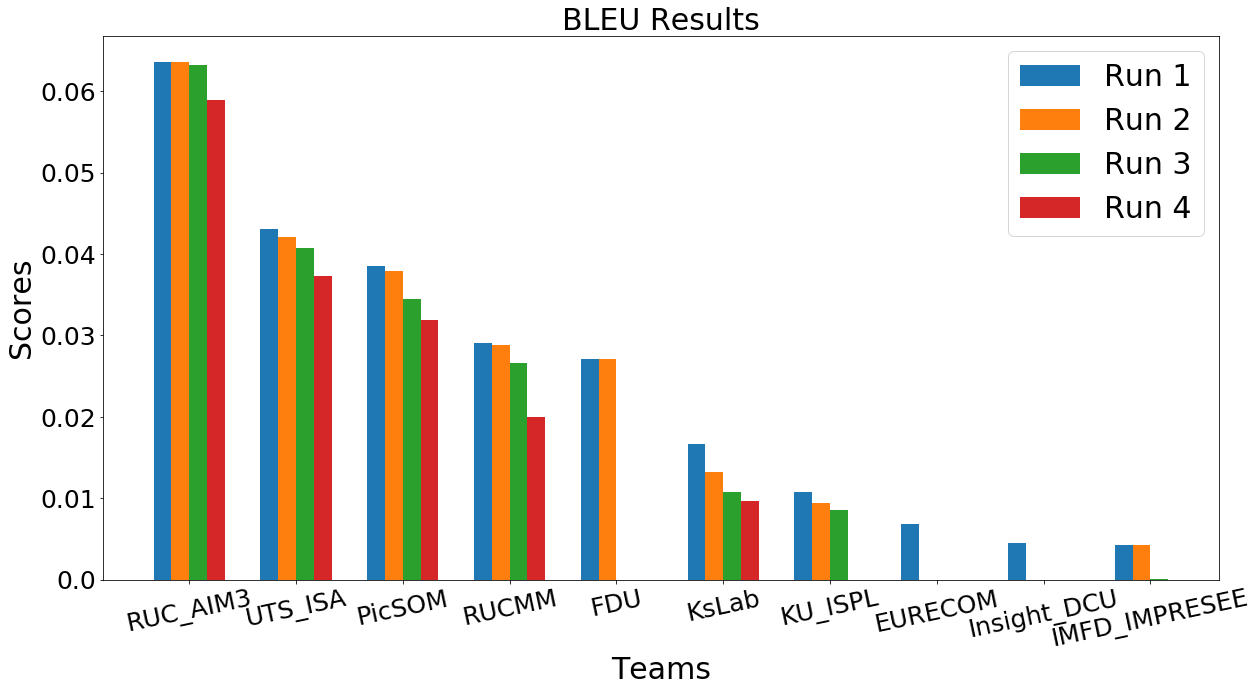}
  \caption{VTT: Comparison of all runs using the BLEU metric.}
  \label{fig:vtt.bleu.results}
\end{figure}

\subsection{System task}

The VTT task is divided into two subtasks:
\begin{itemize}
    \item Description Generation Subtask
    \item Matching and Ranking Subtask
\end{itemize}

Starting in 2019, the description generation subtask has been designated as core/mandatory, which means that teams participating in the VTT task must submit at least one run to this subtask. The matching and ranking subtask is optional for the participants. 
Details of the two subtasks are as follows:

\begin{itemize}
\item \textbf{Description Generation} (Core): For each video, automatically generate a text description of 1 sentence independently and without taking into consideration the existence of any annotated descriptions for the videos.
\item \textbf{Matching and Ranking} (Optional): In this subtask, 5 sets of text descriptions are provided along with the videos. Each set contains a description for each video in the dataset, but the order of descriptions is randomized. The goal of the subtask is to return for each video a ranked list of the most likely text description that corresponds (was annotated) to that video from each of the 5 sets.
\end{itemize}
Up to 4 runs were allowed per team for each of the subtasks.

This year, systems were also required to choose between three run types based on the type of training data they used:
\begin{itemize}
    \item Run type `I' : Training using image captioning datasets only.
    \item Run type `V' : Training using video captioning datasets only.
    \item Run type `B' : Training using both image and video captioning datasets.
\end{itemize}

\begin{figure*}[htbp]
    \centering
    \subfloat[STS 1]{\includegraphics[width=0.5\linewidth]{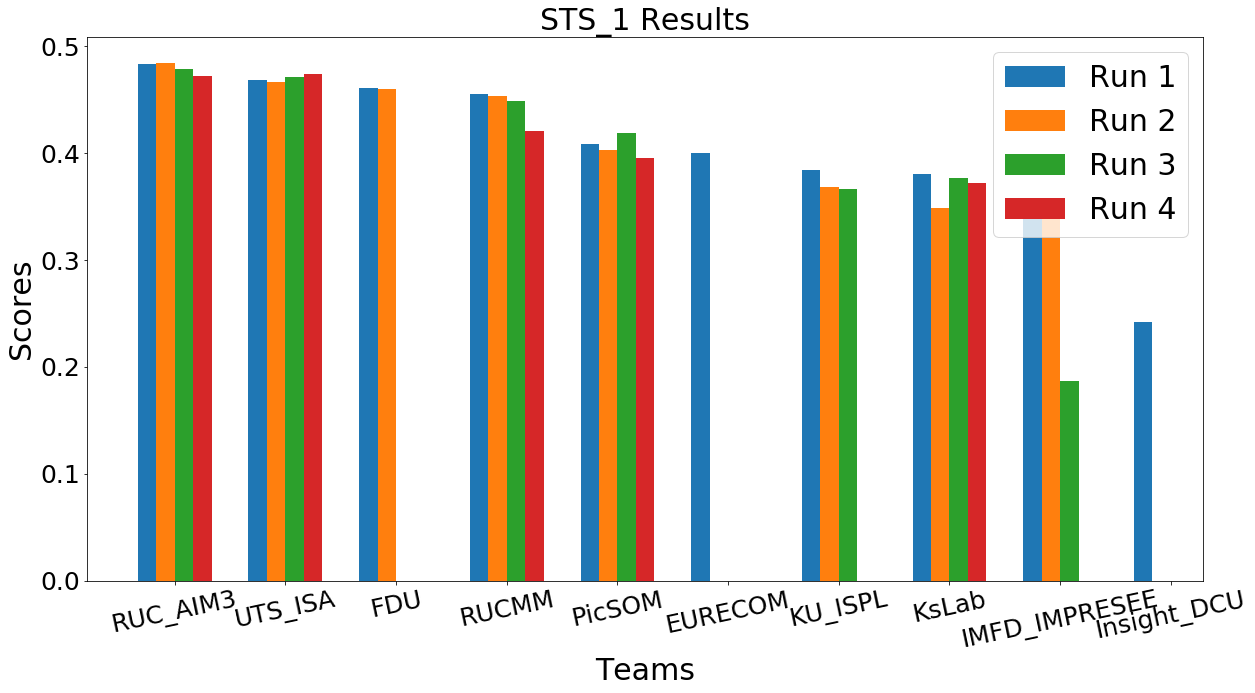}}
    \subfloat[STS 2]{\includegraphics[width=0.5\linewidth]{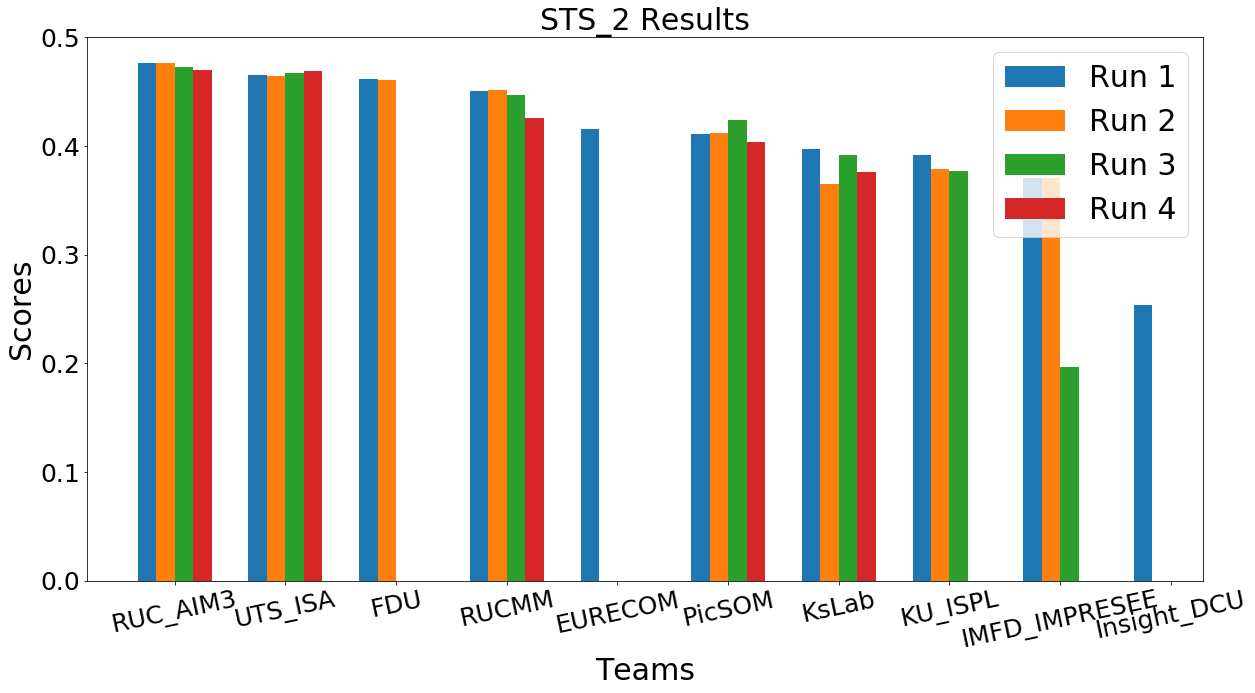}}\\
    \subfloat[STS 3]{\includegraphics[width=0.5\linewidth]{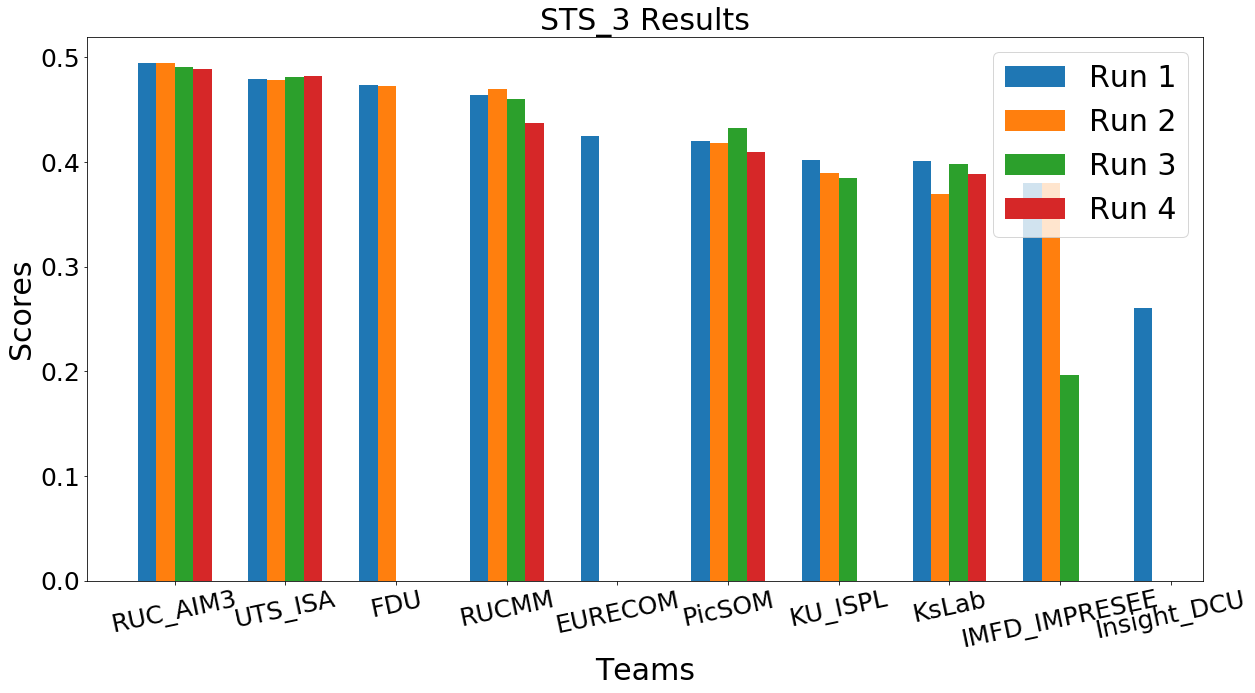}}
    \subfloat[STS 4]{\includegraphics[width=0.5\linewidth]{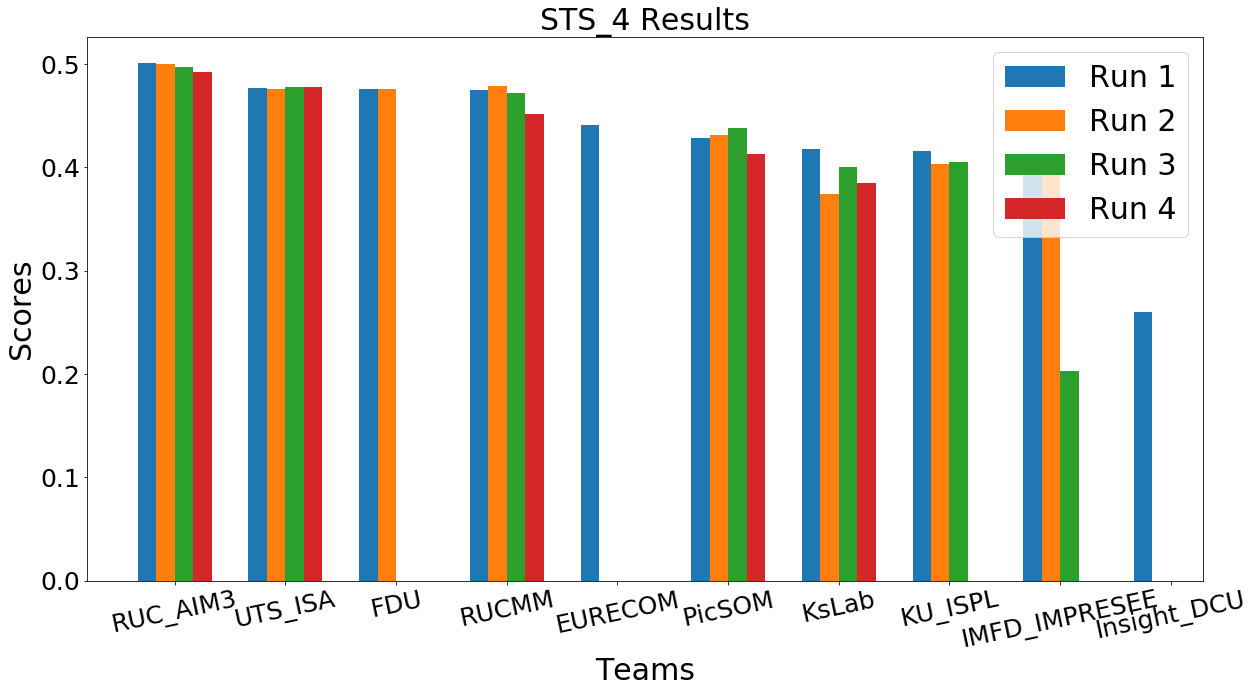}}\\
    \subfloat[STS 5]{\includegraphics[width=0.5\linewidth]{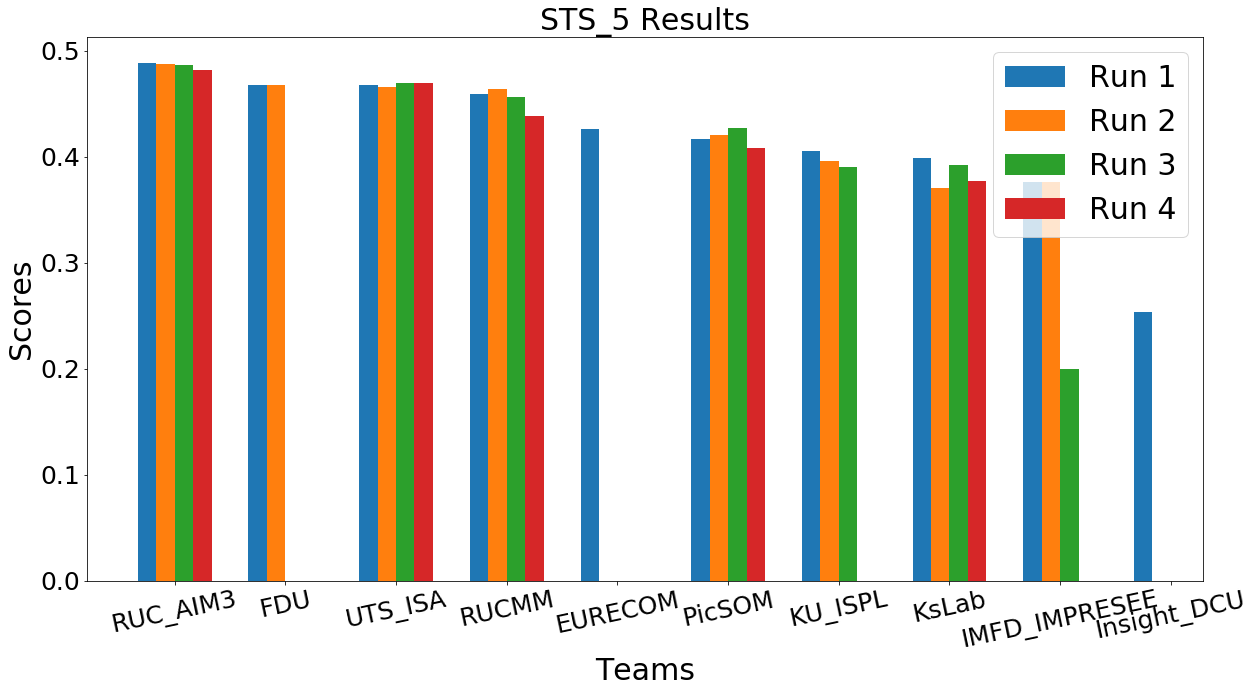}}
    \caption{VTT: Comparison of all runs using the STS metric.}
    \label{fig:vtt.sts.results}
\end{figure*}

\subsection{Evaluation}

The matching and ranking subtask scoring was done automatically against the ground truth using mean inverted rank at which the annotated item is found. The description generation subtask scoring was done automatically using a number of metrics. We also used a human evaluation metric on selected runs to compare with the automatic metrics. 

METEOR (Metric for Evaluation of Translation with Explicit ORdering) \cite{banerjee2005meteor} and BLEU (BiLingual Evaluation Understudy) \cite{papineni2002bleu} are standard metrics in machine translation (MT). BLEU was one of the first metrics to achieve a high correlation with human judgments of quality. It is known to perform poorly if it is used to evaluate the quality of individual sentence variations rather than sentence variations at a corpus level. In the VTT task the videos are independent and there is no corpus to work from. Thus, our expectations are lowered when it comes to evaluation by BLEU.  METEOR is based on the harmonic mean of unigram or n-gram precision and recall in terms of overlap between two input sentences. It redresses some of the shortfalls of BLEU such as better matching synonyms and stemming, though the two measures seem to be used together in evaluating MT.

The CIDEr (Consensus-based Image Description Evaluation) metric \cite{vedantam2015cider} is borrowed from image captioning. It computes TF-IDF (term frequency inverse document frequency) for each n-gram to give a sentence similarity score. The CIDEr metric has been reported to show high agreement with consensus as assessed by humans. We also report scores using CIDEr-D, which is a modification of CIDEr to prevent ``gaming the system''. 

The STS (Semantic Textual Similarity) metric \cite{han2013umbc} was also applied to the results, as in the previous years of this task. This metric measures how semantically similar the submitted description is to one of the ground truth descriptions.

In addition to automatic metrics, the description  generation task includes human evaluation of the quality of automatically generated captions.
Recent developments in Machine Translation evaluation have seen the emergence of DA (Direct Assessment), a method shown to produce highly reliable human evaluation results for MT \cite{DA}. 
DA now constitutes the official method of ranking in main MT benchmark evaluations \cite{WMT17}. With respect to DA for evaluation of video captions (as opposed to MT output), human assessors are presented with a video and a single caption. After watching the video,  assessors rate how well the caption describes what took place in the video on a 0--100 rating scale \cite{graham2018evaluation}. Large numbers of ratings are collected for captions, before ratings are combined into an overall average system rating (ranging from 0 to 100\,\%). Human assessors are recruited via Amazon's Mechanical Turk (AMT) \footnote{\url{http://www.mturk.com}}, with quality control measures applied to filter out or downgrade the weightings from workers unable to demonstrate the ability to rate good captions higher than lower quality captions. This is achieved by deliberately ``polluting'' some of the manual (and correct) captions with linguistic substitutions to generate captions whose semantics are questionable. Thus we might substitute a noun for another noun and turn the manual caption ``A man and a woman are dancing on a table" into ``A {\em horse} and a woman are dancing on a table'', where ``horse'' has been substituted for ``man''.  We expect such automatically-polluted captions to be rated poorly and when an AMT worker correctly does this, the ratings for that worker are improved.

DA was first used as an evaluation metric in TRECVID 2017. We have used this metric again this year to rate each team's primary run, as well as 4 human systems.

\subsection{Overview of Approaches}
For detailed information about the approaches and results for individual teams' performance and runs, the reader should see the various site reports \cite{tv19pubs} in the online workshop notebook proceedings. Here we present a high-level overview of the different systems. 

A large number of datasets are available and are being used by the participants to train their systems. A list of the training datasets used is as follows:
\begin{enumerate}
    \item TGIF
    \item MSR-VTT
    \item MSVD
    \item TRECVID VTT 2016 -- 2018
    \item VATEX
    \item MS-COCO (Image captioning dataset)
\end{enumerate}

\subsubsection{Description Generation}

\begin{figure}[!htbp]
    \centering
    \subfloat[CIDEr]{\includegraphics[width=0.5\linewidth]{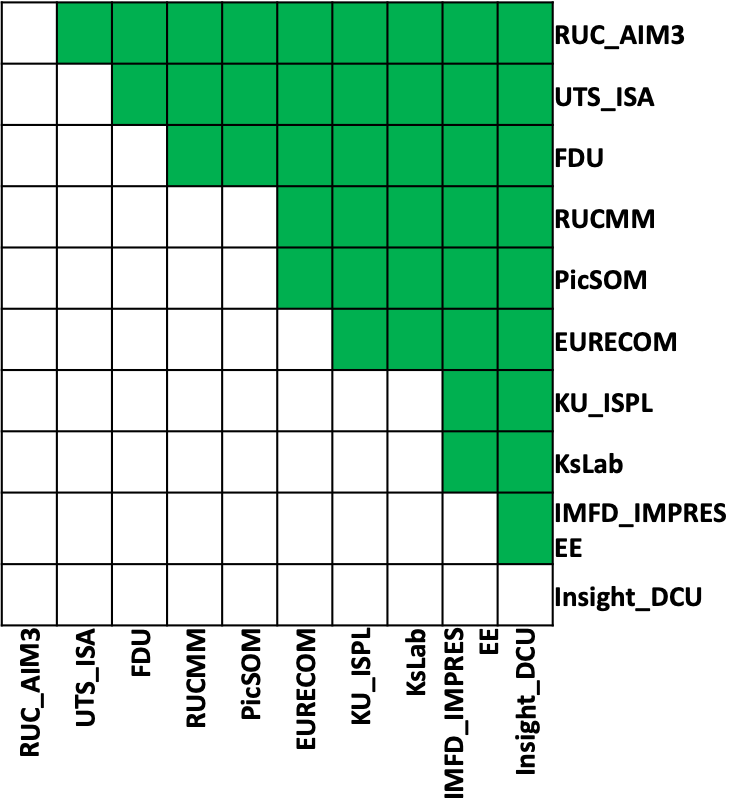}}
    \subfloat[CIDEr-D]{\includegraphics[width=0.5\linewidth]{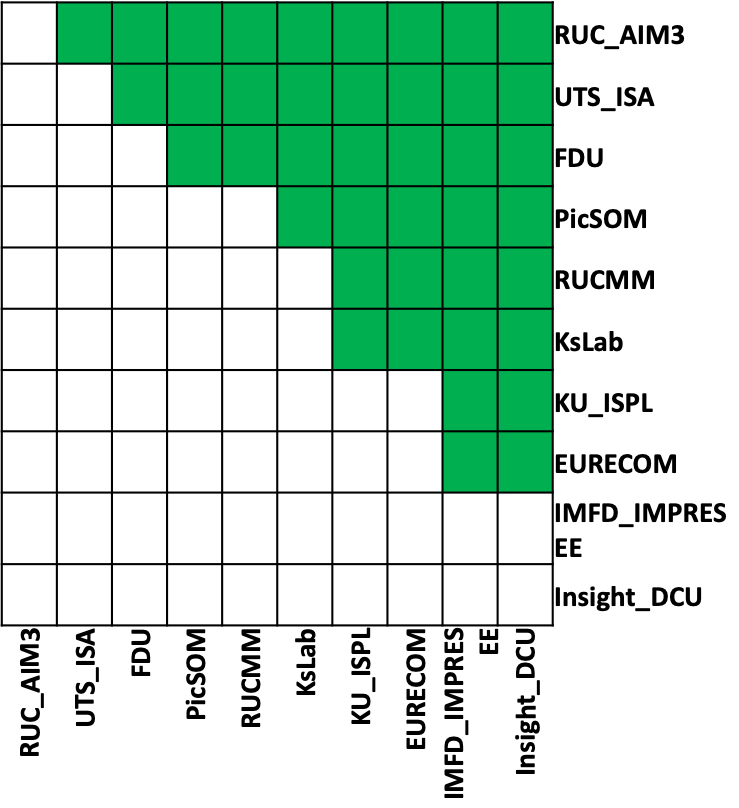}}\\
    \subfloat[METEOR]{\includegraphics[width=0.5\linewidth]{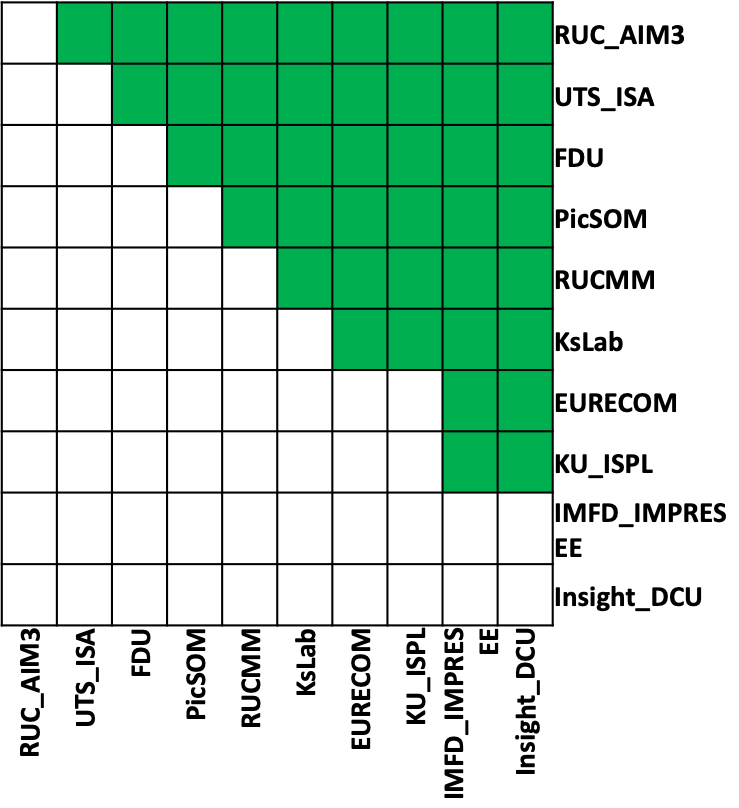}}
    \subfloat[BLEU]{\includegraphics[width=0.5\linewidth]{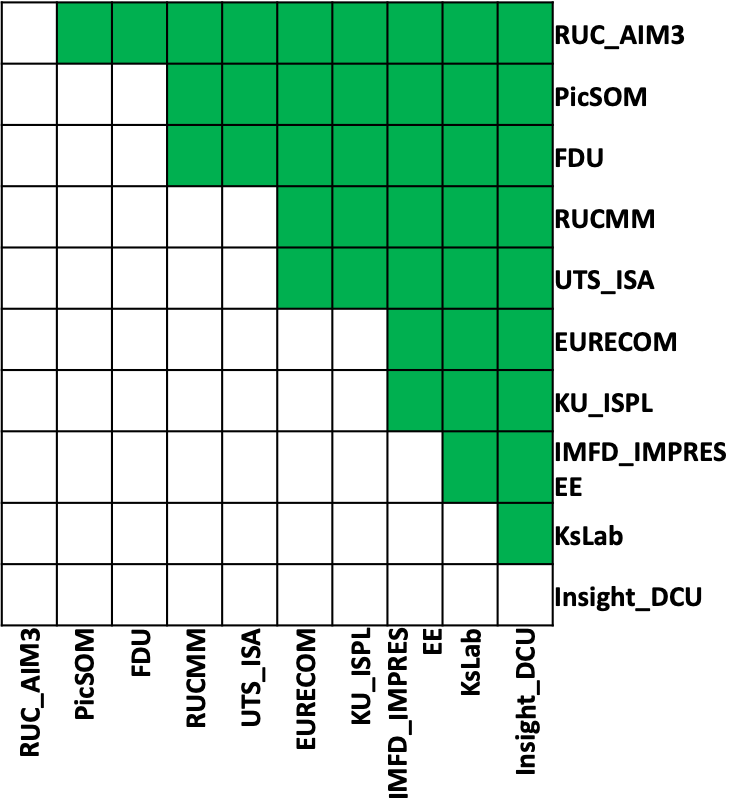}}\\
    \subfloat[STS]{\includegraphics[width=0.5\linewidth]{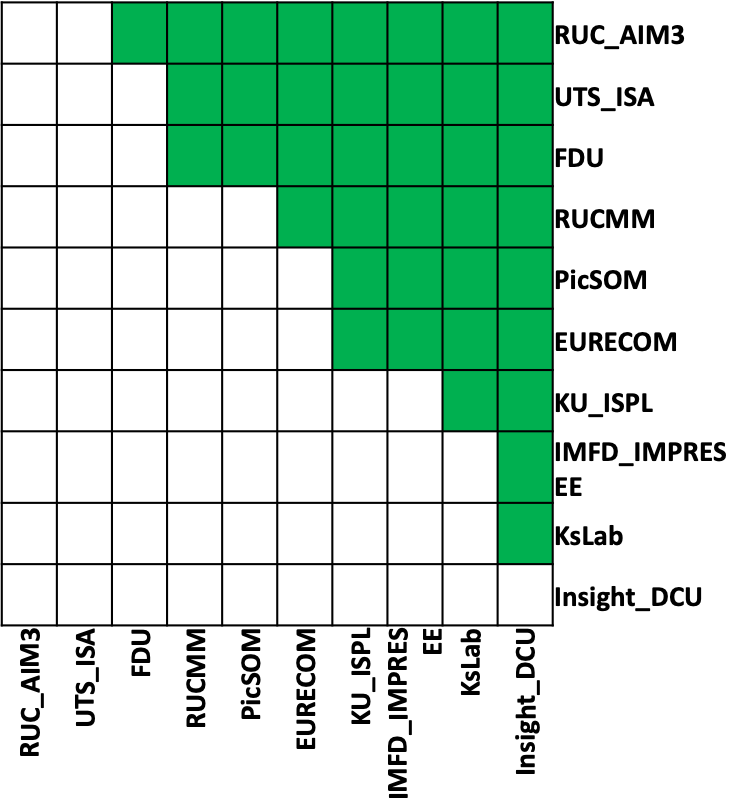}}
    \caption{VTT: Comparison of the primary runs of each team with respect to all the automatic metrics. Green squares indicate a significantly better result for the row over the column.}
    \label{fig:vtt.auto.metric.significance}
\end{figure}

RUC\_AIM3 outperformed the other systems on all metrics. They used video semantic encoding to extract video features in temporal and semantic attention. The captioning model was fine tuned through reinforcement learning with fluency and visual relevance rewards. A pre-trained language model was used for fluency, and for visual relevance they used the matching and ranking model such that the embedding vectors should be close in the joint space. The various caption modules were ensembled to rerank captions.

The UTS\_ISA framework consisted of three parts:
\begin{enumerate}
    \item Extraction of high level visual and action features. ResnetXT-WSL and EfficientNet were used for visual features, whereas Kinect-i3d features were used for action and temporal information.
    \item An LSTM based encoder-decoder framework was used to handle fusion and learning.
    \item Finally, an expandable ensemble module was used, and a controllable beam search strategy generated sentences of different lengths.
\end{enumerate}

RUCMM based their system on the classical encoder-decoder framework. They utilized the video-side multi-level encoding branch of dual encoding framework instead of common mean pooling.

DCU used the commonly used bidirectional LSTM network. They used C3D as input followed by soft attention, which was fed again to a final LSTM. A beam search method was used to find the sentences with the highest probability.

IMFD\_IMPRESSEE used a semantic compositional network (SCN) to understand effectively the individual semantic concepts for videos. Then, a recurrent encoder based on a bidirectional LSTM was used.

FDU used the Inception-Resnet-V2 CNN pretrained on the ImageNet dataset for visual representation. They used concept detection to remove gap between feature representation and text domain. Finally, an LSTM network was used to generate the sentences.

KSLab attempted to decrease the processing time for the task. They achieved this by processing 5 consecutive frames from the beginning and end of the video. Each frame was converted to a 2048 feature vector through the Inception V3 network. An encoder-decoder network was constructed by two LSTM networks. It seems reasonable to assume that this approach will only work for videos where the first and last few frames are representative of the video, and no substantial information is present in the middle frames.

PicSOM compared the cross-entropy and self-critical training loss functions. They used the CIDEr-D scores as reward in reinforcement learning for the self-critical loss funciton. As expected, this worked better than cross-entropy. They also trained systems using each of the three run types, and found that using both image and video data for training improved their results. When combining the training data, they used non-informative video features for the image dataset.

EURECOM experimented with the use of Curriculum Learning in video captioning. The idea was to present data in an ascending order of difficulty during training. They translated captions into a list of indices, where a bigger index was used for less frequent words. The score of a sample was then the maximum index of its caption. Video features were extracted with an I3D neural network. Unfortunately, they did not see any benefits of this process.

\subsubsection{Matching and Ranking}
RUC\_AIM3 used the dual encoding model~\cite{dong2019dual}. Given a sequence of input features, they used 3 branches to encode global, temporal, and local information. The encoded features were then concatenated and mapped into joint embedding space. 

RUCMM used dual encoding, and included the BERT encoder to improve it. Their best results were obtained by combining models.

IMFD\_IMPRESSEE used a deep learning model based on W2VV++ (which was developed for AVS). They extended it by using dense trajectories as visual embedding to encode temporal information for the video. K-means clustering was used to encode dense trajectories. They found that not using batch normalization improved their results.

\begin{figure}[htbp]
  \centering
  \includegraphics[width=1.0\linewidth]{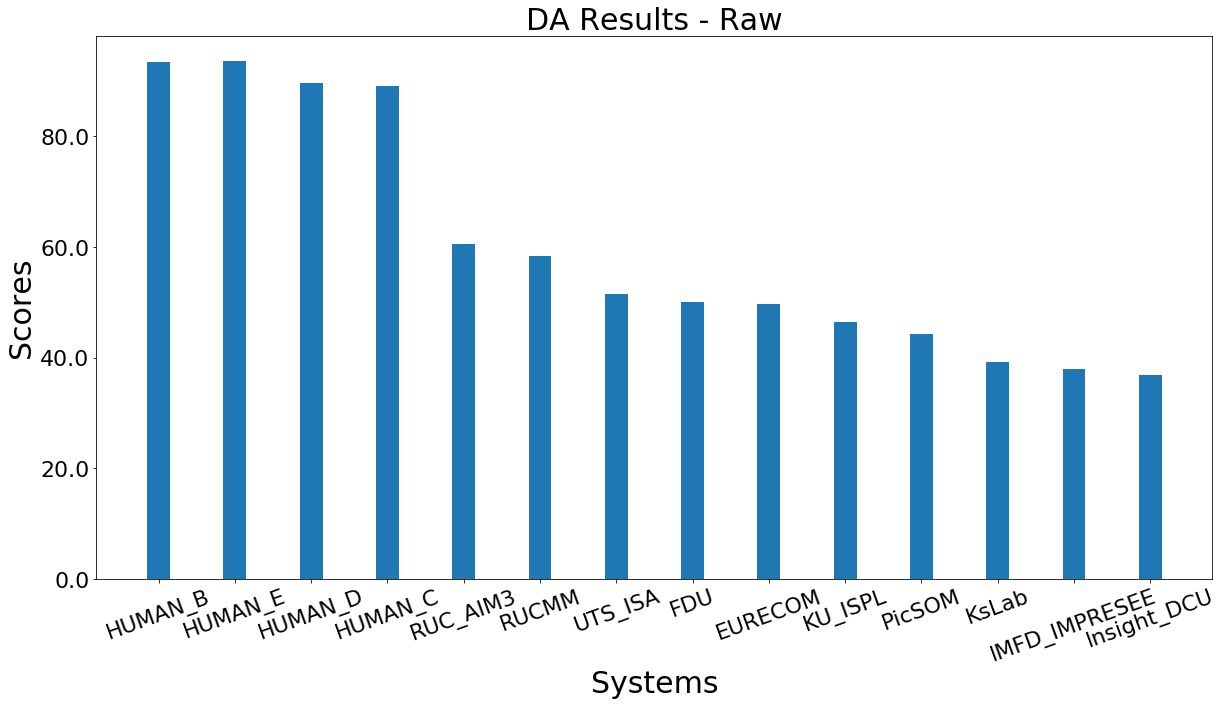}
  \caption{VTT: Average DA score for each system. The systems compared are the primary runs submitted, along with 4 manually generated system labeled as HUMAN\_B to HUMAN\_E.}
  \label{fig:vtt.da.raw.results}
\end{figure}

\begin{figure}[htbp]
  \centering
  \includegraphics[width=1.0\linewidth]{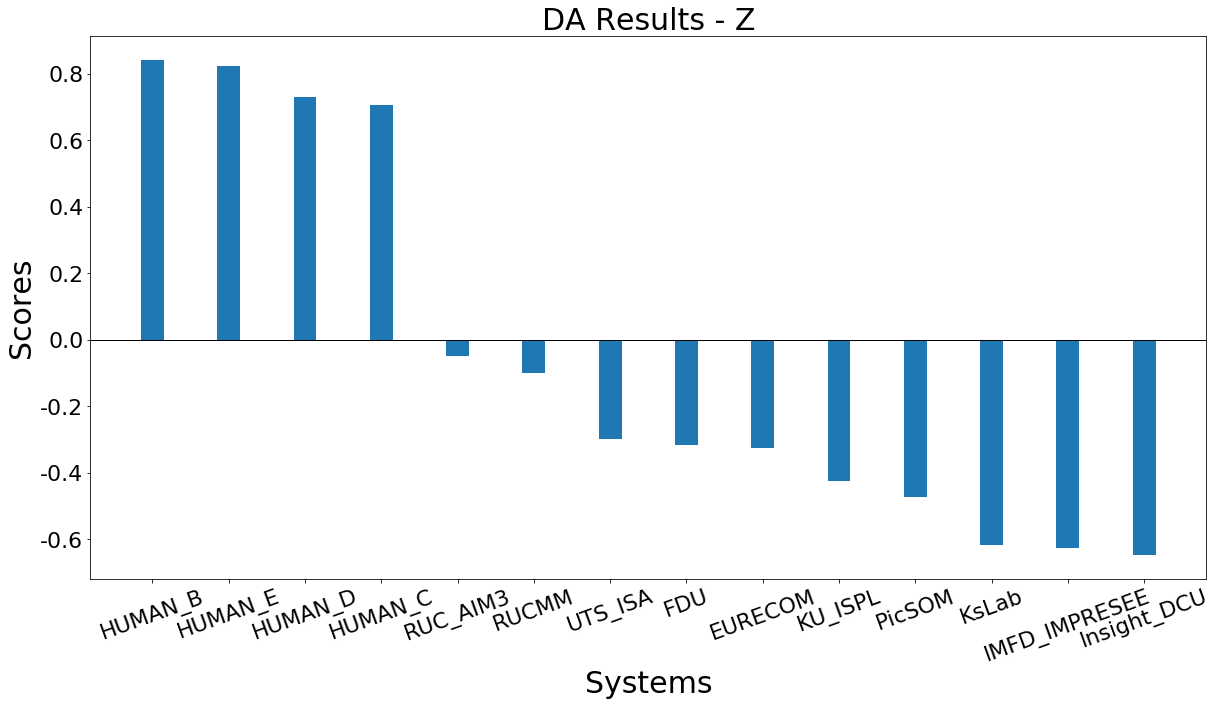}
  \caption{VTT: Average DA score per system after standardization per individual worker's mean and standard deviation score.}
  \label{fig:vtt.da.z.results}
\end{figure}

\begin{figure}[!htbp]
  \centering
  \includegraphics[width=1.0\linewidth]{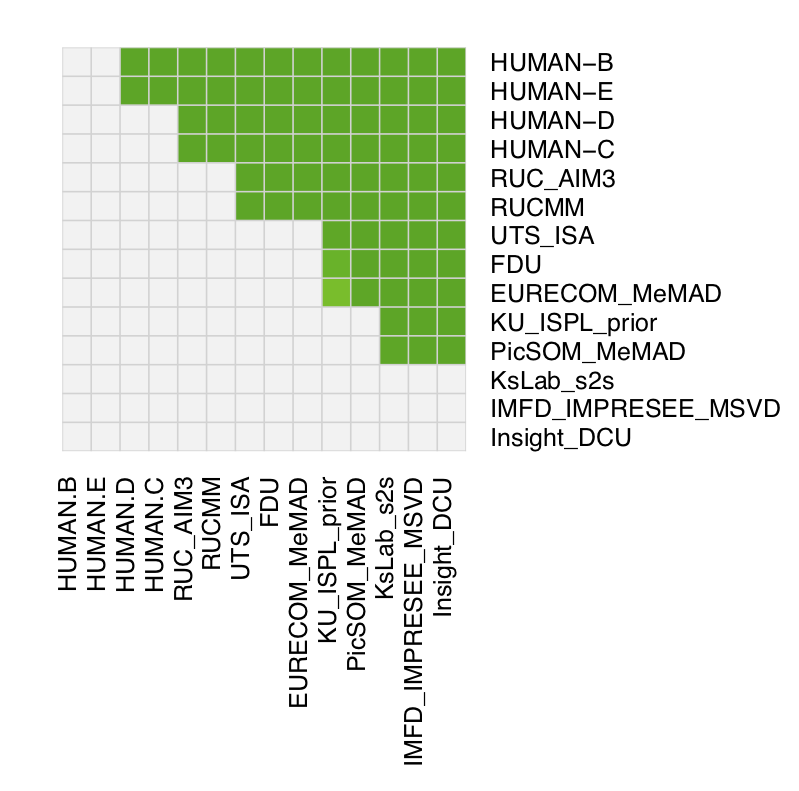}
  \caption{VTT: Comparison of systems with respect to DA. Green squares indicate a significantly better result for the row over the column.}
  \label{fig:vtt.da.heat.map}
\end{figure}

\subsection{Results}

\subsubsection{Description Generation}

\begin{table*}
\begin{tabular}{|l|rrrrrrrrr|}
\hline
 &  CIDER &  CIDER-D &  METEOR &  BLEU &  STS\_1 &  STS\_2 &  STS\_3 &  STS\_4 &  STS\_5 \\
\hline
CIDER   &        1.000 &          0.964 &         0.923 &       0.902 &  0.929 &  0.900 &  0.910 &  0.887 &  0.900 \\
CIDER-D &        0.964 &          1.000 &         0.903 &       0.958 &  0.848 &  0.815 &  0.828 &  0.800 &  0.816 \\
METEOR  &        0.923 &          0.903 &         1.000 &       0.850 &  0.928 &  0.916 &  0.921 &  0.891 &  0.904 \\
BLEU    &        0.902 &          0.958 &         0.850 &       1.000 &  0.775 &  0.742 &  0.752 &  0.724 &  0.741 \\
STS\_1         &        0.929 &          0.848 &         0.928 &       0.775 &  1.000 &  0.997 &  0.998 &  0.990 &  0.994 \\
STS\_2         &        0.900 &          0.815 &         0.916 &       0.742 &  0.997 &  1.000 &  0.999 &  0.995 &  0.997 \\
STS\_3         &        0.910 &          0.828 &         0.921 &       0.752 &  0.998 &  0.999 &  1.000 &  0.995 &  0.997 \\
STS\_4         &        0.887 &          0.800 &         0.891 &       0.724 &  0.990 &  0.995 &  0.995 &  1.000 &  0.998 \\
STS\_5         &        0.900 &          0.816 &         0.904 &       0.741 &  0.994 &  0.997 &  0.997 &  0.998 &  1.000 \\
\hline

\end{tabular}
\caption{VTT: Correlation scores between automatic metrics.}
\label{tab:vtt.auto.metric.corr}
\end{table*}

\begin{table}[]
    \centering
    \begin{tabular}{|c|c|c|}
    \hline
    Metric & 2018 & 2019 \\ \hline \hline
    CIDEr & 0.416 & 0.585\\ \hline
    CIDEr-D & 0.154 & 0.332\\ \hline
    METEOR & 0.231 & 0.306\\ \hline
    BLEU & 0.024 & 0.064\\ \hline
    STS & 0.433 & 0.484\\ \hline
    \end{tabular}
    \caption{VTT: Comparison of maximum scores for each metric in 2018 and 2019. Scores have increased across all metrics from last year.}
    \label{tab:desc.year.comparison}
\end{table}

\begin{table*}[]
    \centering
    \begin{tabular}{|l|rrrrrr|}
\hline
{} &  CIDER &  CIDER-D &  METEOR &  BLEU &  STS\_1 &   DA\_Z \\
\hline
CIDER   &        1.000 &          0.972 &         0.963 &       0.902 &  0.937 &  0.874 \\
CIDER-D &        0.972 &          1.000 &         0.967 &       0.969 &  0.852 &  0.832 \\
METEOR  &        0.963 &          0.967 &         1.000 &       0.936 &  0.863 &  0.763 \\
BLEU    &        0.902 &          0.969 &         0.936 &       1.000 &  0.750 &  0.711 \\
STS\_1         &        0.937 &          0.852 &         0.863 &       0.750 &  1.000 &  0.812 \\
DA\_Z          &        0.874 &          0.832 &         0.763 &       0.711 &  0.812 &  1.000 \\
\hline
\end{tabular}

    \caption{VTT: Correlation of scores between metrics for the primary runs of each team. `DA\_Z' is the score given by humans, whereas all other metrics are automatic.}
    \label{tab:vtt.primary.corr}
\end{table*}

\begin{figure}
    \centering
    \includegraphics[width=1\linewidth]{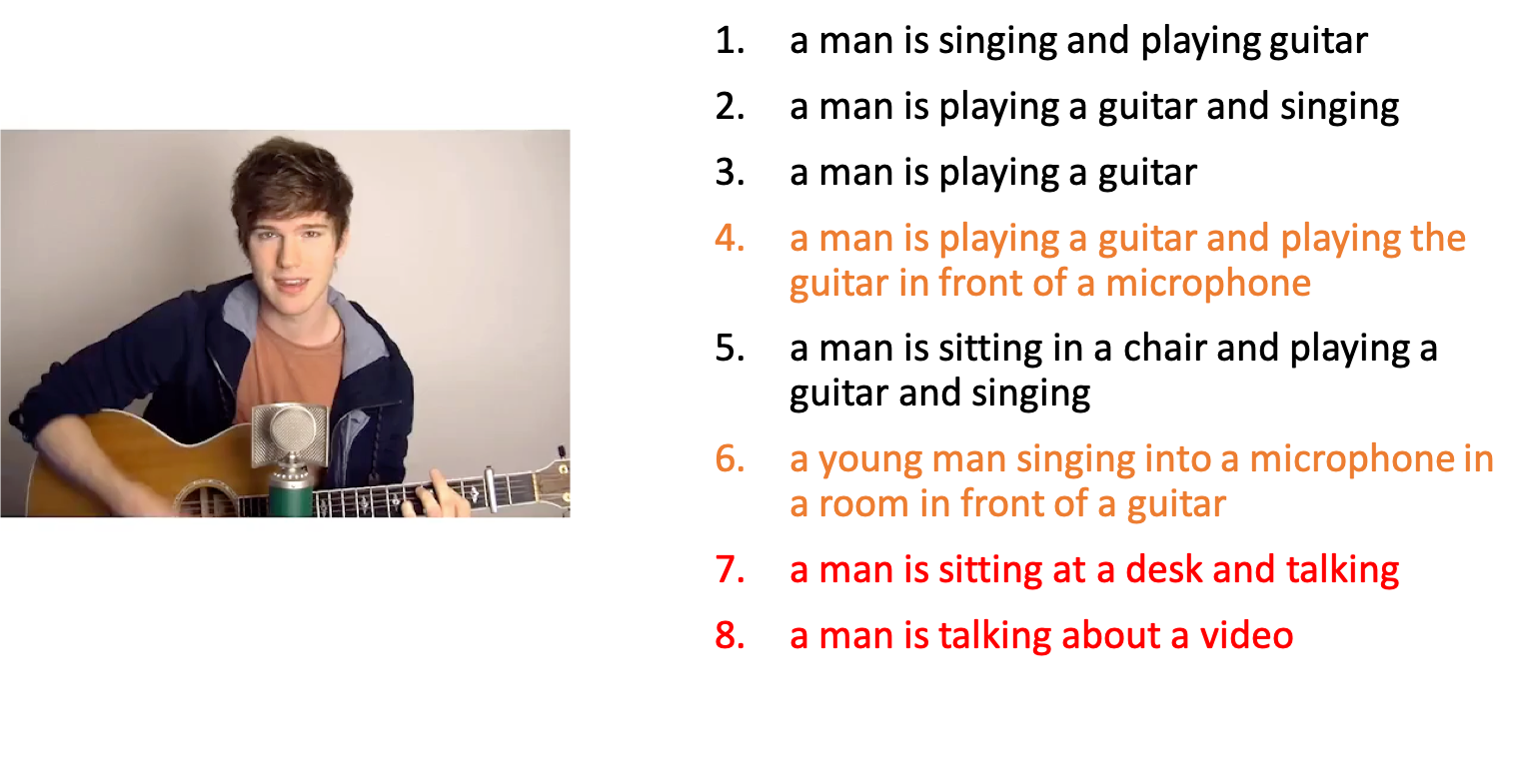}
    \caption{System captions for a video. In general, systems scored high on this video. Captions 7 and 8 are obviously wrong, but captions 4 and 6 may score high on automatic metrics, despite not being good natural language sentences.}
    \label{fig:vtt.example.captions}
\end{figure}

\begin{figure}[htbp]
  \centering
  \includegraphics[width=1.0\linewidth]{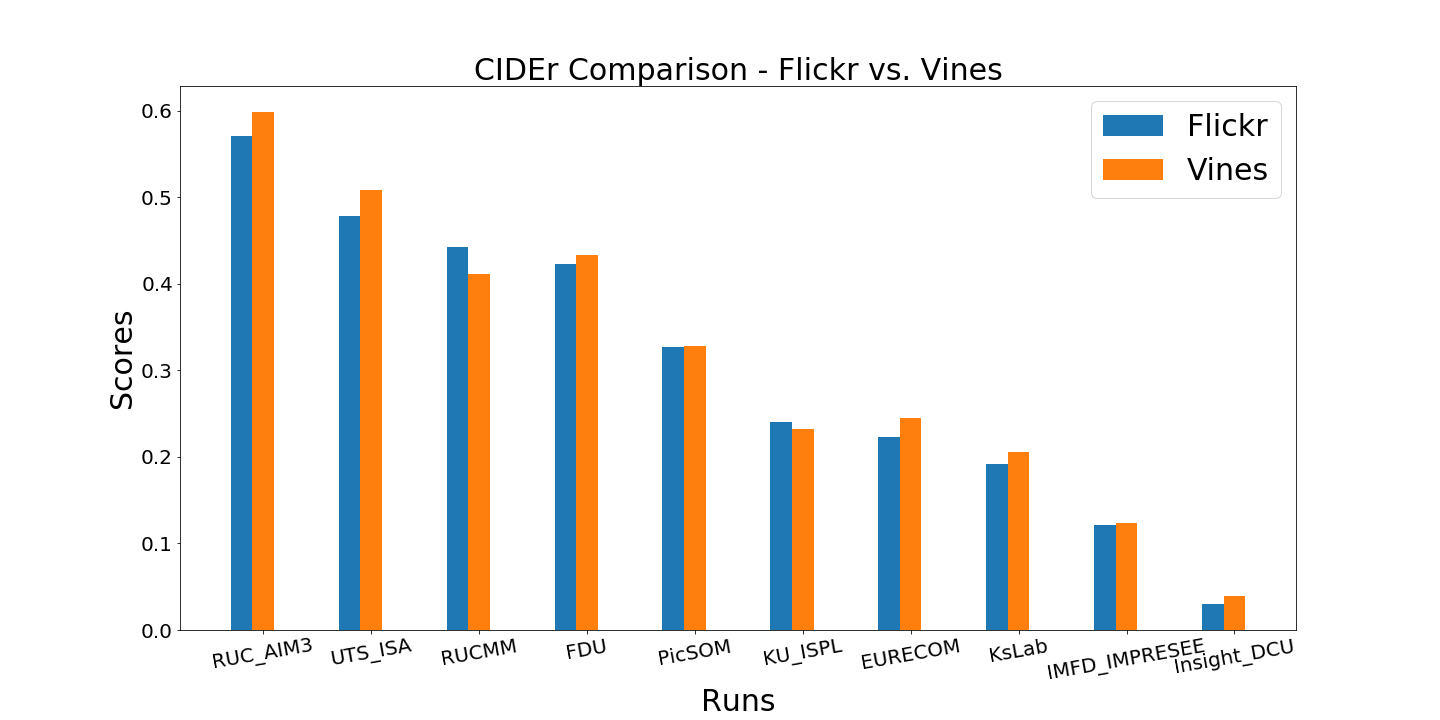}
  \caption{VTT: Comparison of Flickr and Vine videos using the CIDEr metric.}
  \label{fig:vtt.flickr.vines.cider}
\end{figure}

\begin{figure}[htbp]
  \centering
  \includegraphics[width=1.0\linewidth]{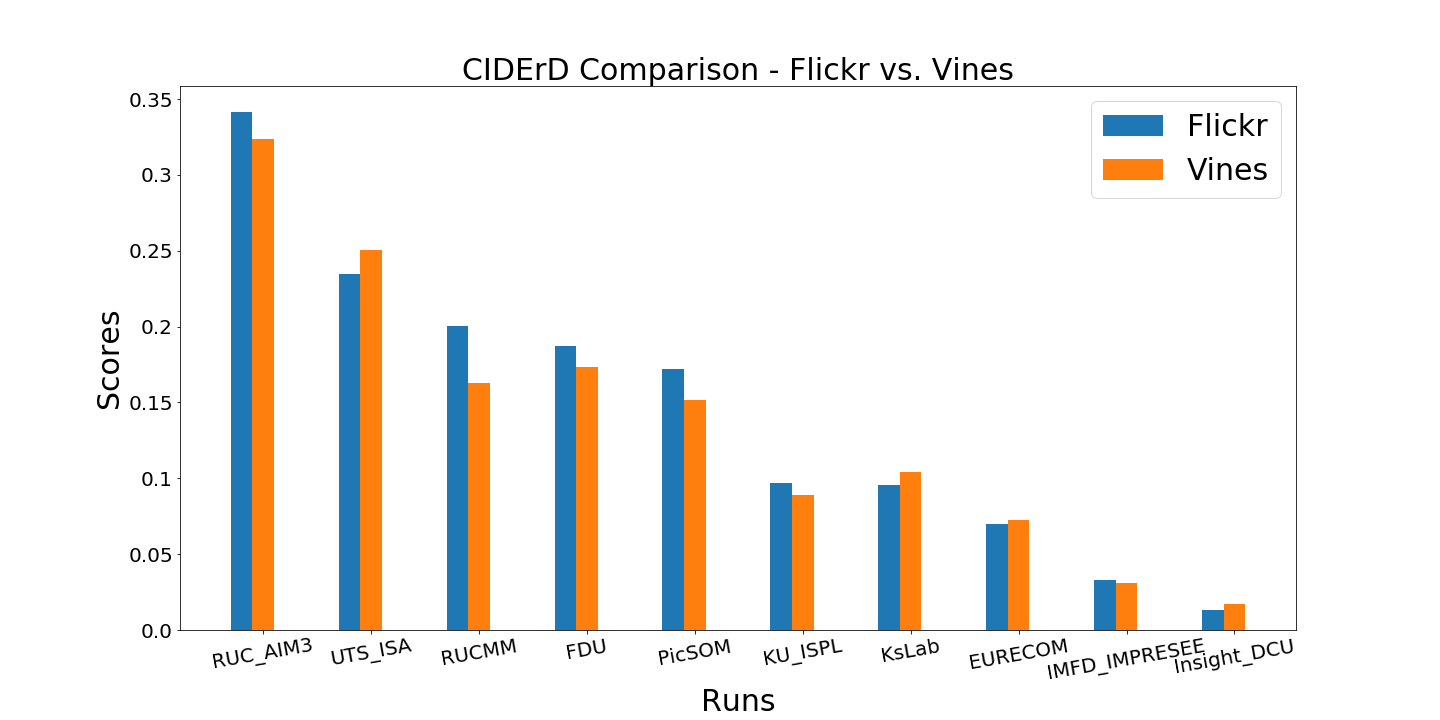}
  \caption{VTT: Comparison of Flickr and Vine videos using the CIDEr-D metric.}
  \label{fig:vtt.flickr.vines.ciderd}
\end{figure}

\begin{figure}[htbp]
  \centering
  \includegraphics[width=1.0\linewidth]{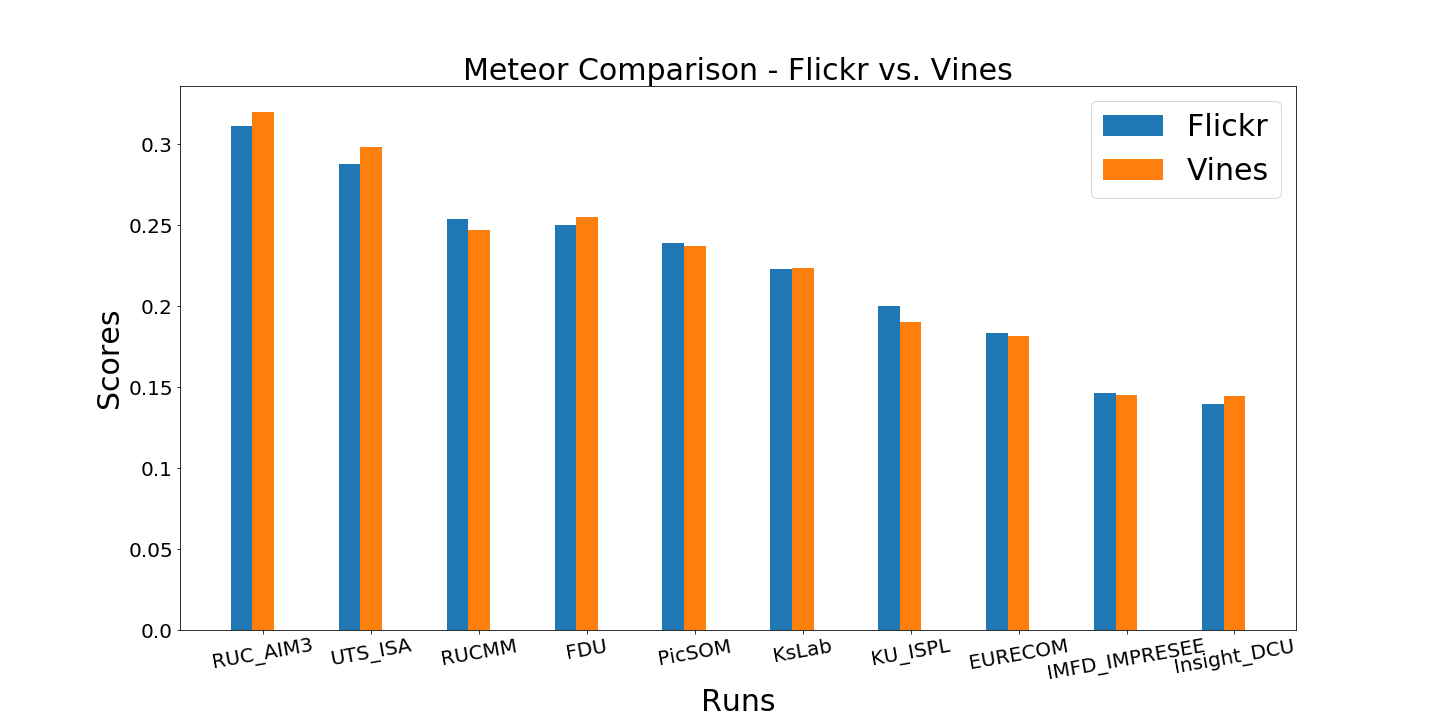}
  \caption{VTT: Comparison of Flickr and Vine videos using the METEOR metric.}
  \label{fig:vtt.flickr.vines.meteor}
\end{figure}

\begin{figure}[htbp]
  \centering
  \includegraphics[width=1.0\linewidth]{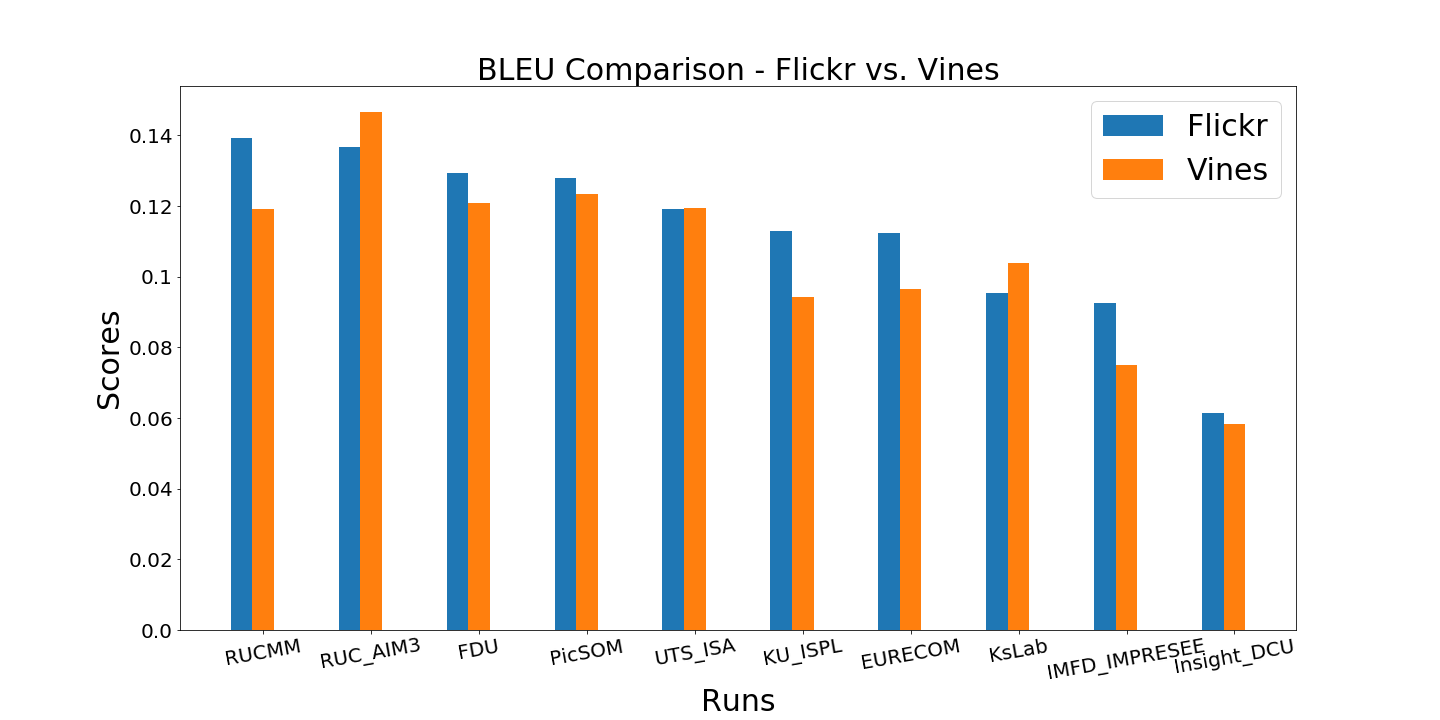}
  \caption{VTT: Comparison of Flickr and Vine videos using the BLEU metric.}
  \label{fig:vtt.flickr.vines.bleu}
\end{figure}

\begin{figure}[htbp]
  \centering
  \includegraphics[width=1.0\linewidth]{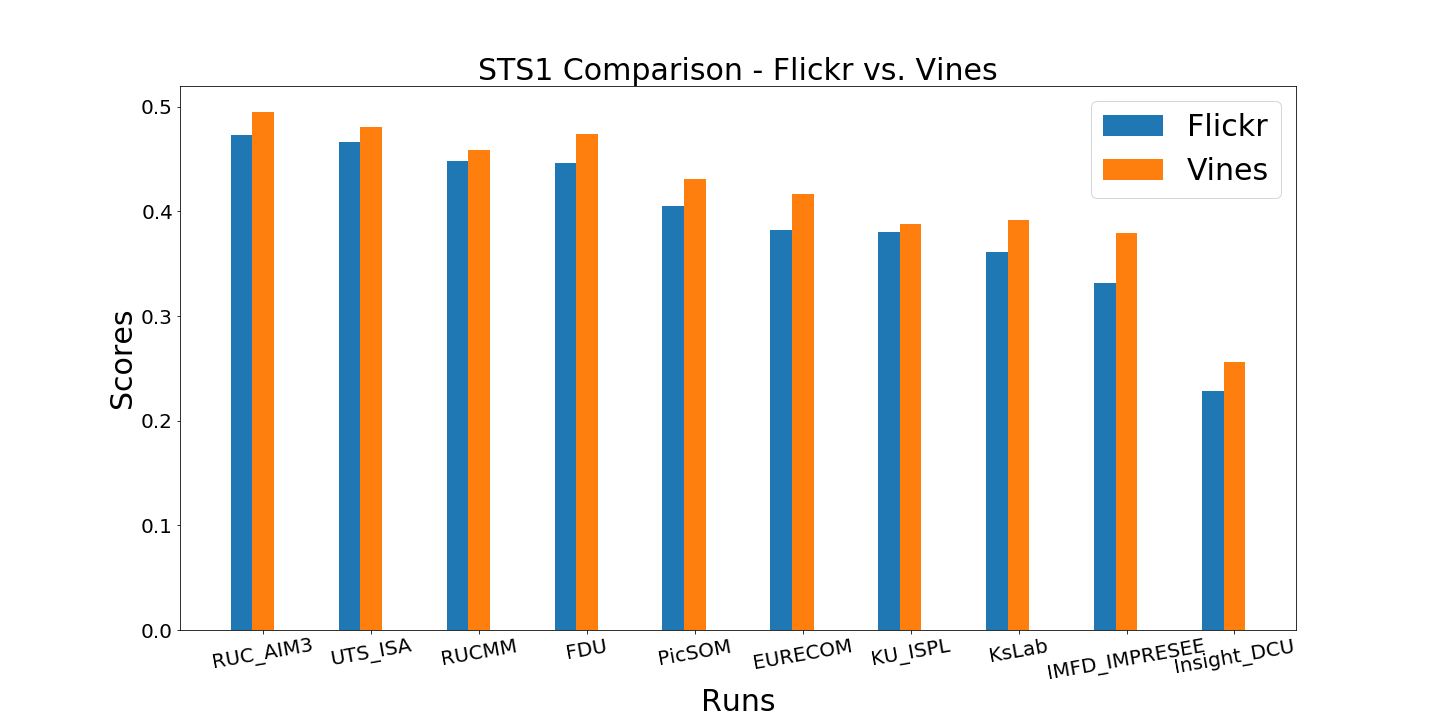}
  \caption{VTT: Comparison of Flickr and Vine videos using the STS metric.}
  \label{fig:vtt.flickr.vines.sts}
\end{figure}

\begin{figure*}[!htb]
  
  \subfloat[Video \#1439]{\includegraphics[width=0.3\textwidth]{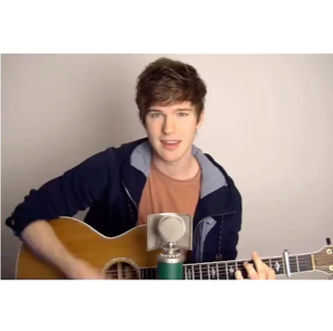}} \hfill
  \subfloat[Video \#1080]{\includegraphics[width=0.3\textwidth]{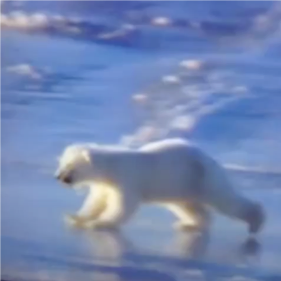}} \hfill
  \subfloat[Video \#826]{\includegraphics[width=0.3\textwidth]{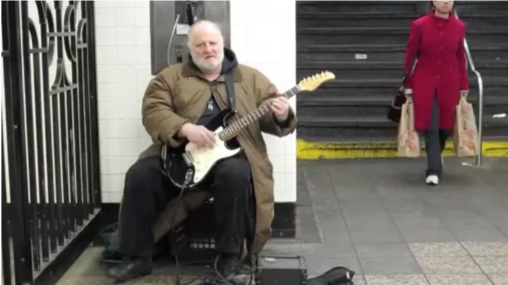}}
 
  \caption{VTT: The top 3 videos for the description generation subtask with their video IDs. All videos focus on a single object, and there are not much movement or variations between frames.}
  \label{fig:desc.top3.results}
  
\end{figure*}

\begin{figure*}[!htb]

  \subfloat[Video \#688]{\includegraphics[width=0.3\linewidth]{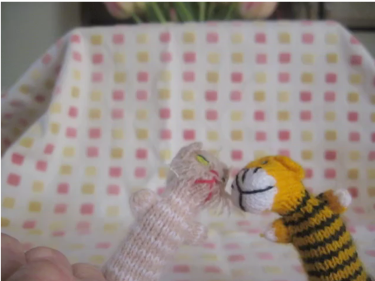}} \hfill
  \subfloat[Video \#1330]{\includegraphics[width=0.3\linewidth]{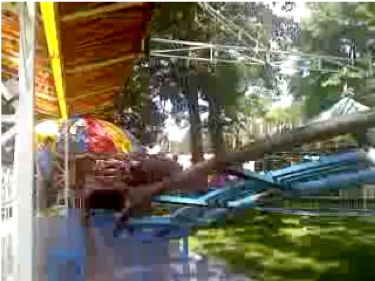}} \hfill
  \subfloat[Video \#913]{\includegraphics[width=0.3\linewidth]{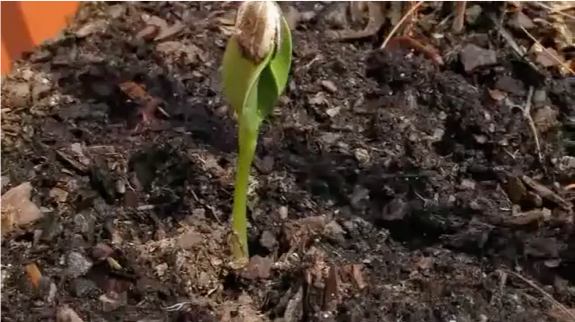}}

  \caption{VTT: The bottom 3 videos for the description generation subtask with their video IDs. These videos mostly have uncommon objects and actions, and in some cases there is a lot of activity in the video.}
  \label{fig:desc.bottom3.results}
\end{figure*}

The distribution of runs for this subtask is as follows:
\begin{itemize}
    \item Type `I': 1 run
    \item Type `B': 3 runs
    \item Type `V': 26 runs
\end{itemize}
It is, therefore, not possible to make any meaningful comparison between the performance of these different run types.

Each team identified one run as their `primary' run. Interestingly, the primary run was not necessarily the best run for each team according to the metrics.

The description generation subtask scoring was done using popular automatic metrics that compare the system generation captions with groundtruth captions as provided by assessors. We also continued the use of Direct Assessment, which was introduced in TRECVID 2017, to compare the submitted runs. 

Figure~\ref{fig:vtt.cider.results} shows the comparison of all teams using the CIDEr metric. All runs submitted by each team are shown in the graph. Figure~\ref{fig:vtt.ciderd.results} shows the scores for the CIDEr-D metric, which is a modification of CIDEr. Figures~\ref{fig:vtt.meteor.results} and~\ref{fig:vtt.bleu.results} show the scores for METEOR and BLEU metrics respectively. The STS metric allows comparison between two sentences. For this reason, the captions are compared to a single groundtruth description at a time, resulting in 5 STS scores. Figure~\ref{fig:vtt.sts.results} shows all the STS scores. It can be seen that all 5 graphs are very similar to each other. For further comparison purposes, we will use STS\_1 to represent the STS scores.

Table~\ref{tab:vtt.auto.metric.corr} shows the correlation between the average scores for all runs for the automatic metrics. The correlation between each of the STS scores is above 0.99, proving our hypothesis that there is not much to differentiate between them. In general, the metrics seem to correlate well. CIDEr-D has a high correlation with CIDEr, but comparatively lower correlation with STS. BLEU seems to be least correlated with STS, as well METEOR. 

Figure~\ref{fig:vtt.auto.metric.significance} shows how the systems compare according to each of the metrics. The green squares indicate that the system in the row is significantly better (p < 0.05) than the system shown in the column. The figure shows that RUC\_AIM3 outperforms all other systems according to most metrics.

Scores have increased across all metrics from last year. Table~\ref{tab:desc.year.comparison} shows the maximum scores for all metrics in 2018 and 2019. The testing datasets were different, which makes a direct comparison of scores difficult. However, the selection process of the videos was similar between the two years, and we expect that the score increase may, at least partially, be due to the improvement in systems.  

Figure~\ref{fig:vtt.da.raw.results} shows the average DA score [$0 - 100$] for each system. The score is micro-averaged per caption, and then averaged over all videos. Figure~\ref{fig:vtt.da.z.results} shows the average DA score per system after it is standardized per individual AMT worker's mean and standard deviation score. The HUMAN systems represent manual captions provided by assessors. As expected, captions written by assessors outperform the automatic systems. Figure~\ref{fig:vtt.da.heat.map} shows how the systems compare according to DA. The green squares indicate that the system in the row is significantly better than the system shown in the column (p < 0.05). The figure shows that no system reaches the level of the human performance. Among the systems, RUC\_AIM3 and RUCMM outperform the other systems. An interesting observation is that HUMAN\_B and HUMAN\_E statistically perform better than HUMAN\_C and HUMAN\_D. This may not be important since each `HUMAN' system contains multiple annotators. One possible reason could be due to the difference in average sentence lengths in the different sets of annotations. 

Table~\ref{tab:vtt.primary.corr} shows the correlation between different metrics for the primary runs of all teams. The `DA\_Z' metric is the score generated by humans. It can be observed that this score seems to be the least correlated to other metrics. There could be multiple reasons for this. One possibility is that while most automatic metrics make use of the words in sentences, they may miss semantic information that is obvious to humans. For example, Figure~\ref{fig:vtt.example.captions} shows eight system captions for a video\footnote{All figures are in the public domain and permissible under HSPO \#ITL-17-0025} on which most systems scored high. Some captions (such as 4 and 6) may score well since they have all the relevant words, but may not be judged to be good sentences by people. 

We also compared how the Flickr and Vines videos compared in their level of difficulty. Figures~\ref{fig:vtt.flickr.vines.cider}-~\ref{fig:vtt.flickr.vines.sts} show the comparison of scores for the Flickr and Vines on different metrics for all the teams. There was no evidence that systems performed better on one source than the other. 

Figure~\ref{fig:desc.top3.results} shows the top 3 videos for this subtask. Most systems were able to describe these videos well. Figure~\ref{fig:desc.bottom3.results} shows the bottom 3 videos for this subtask. The systems failed to provide acceptable descriptions for these videos.

\begin{figure*}[htb]
  \centering
  \includegraphics[width=1\linewidth]{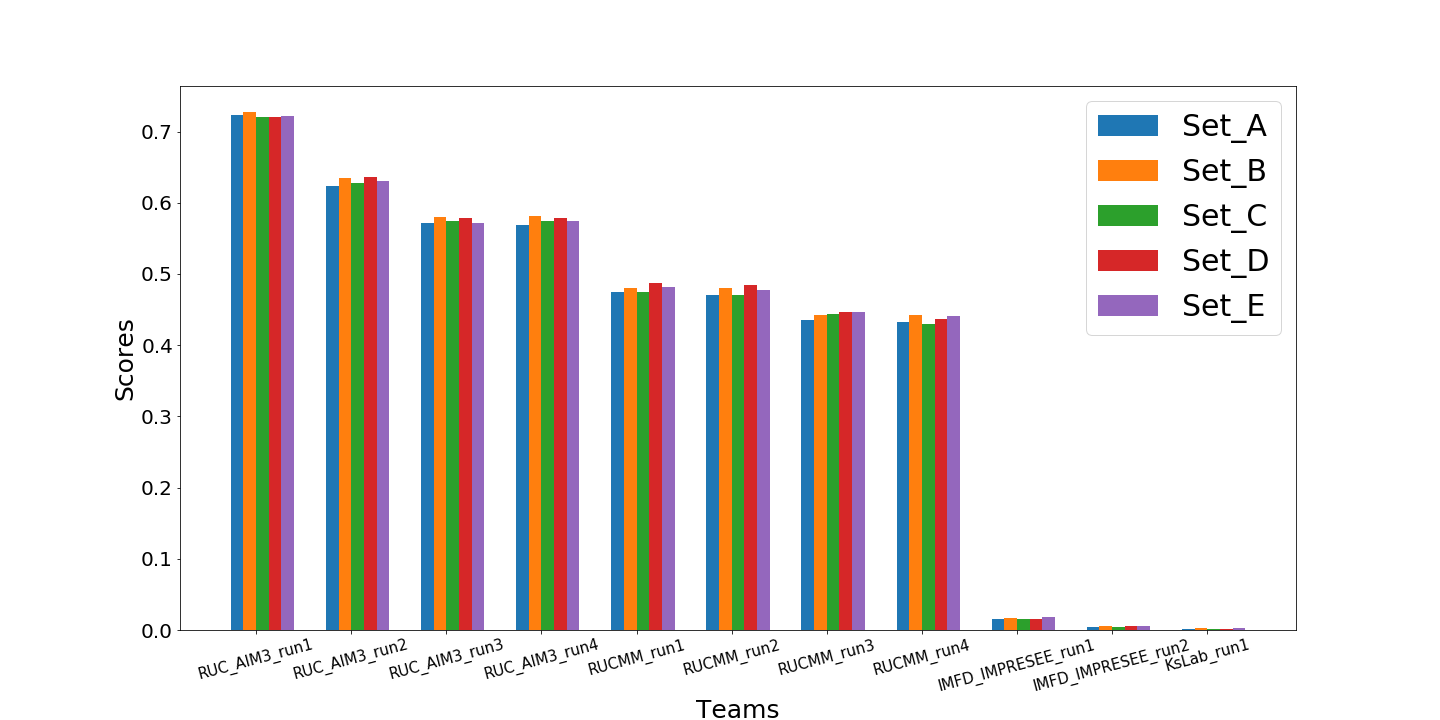}
  \caption{VTT: Matching and Ranking results across all runs for all sets.}
  \label{fig:vtt.match.rank.all.sets}
\end{figure*}

\begin{figure*}[!htb]

  \subfloat[Video \#13]{\includegraphics[width=0.3\linewidth]{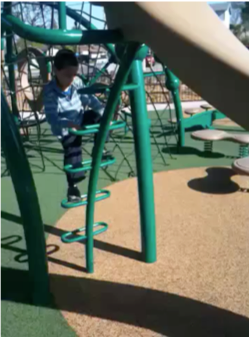}} \hfill
  \subfloat[Video \#455]{\includegraphics[width=0.3\linewidth]{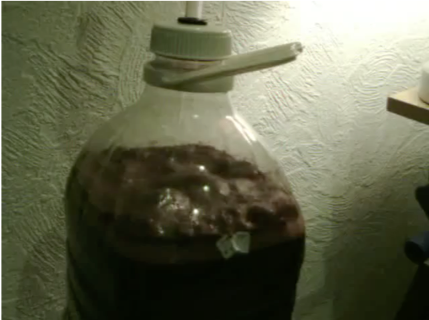}} \hfill
  \subfloat[Video \#32]{\includegraphics[width=0.3\linewidth]{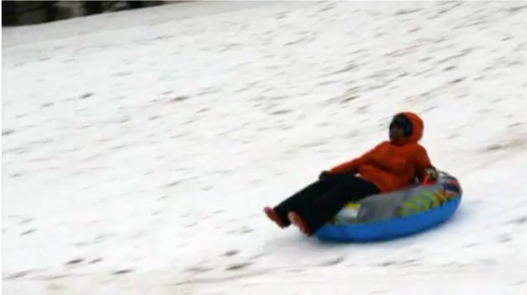}}
 
  \caption{VTT: The top 3 videos for the matching and ranking subtask with their video IDs. All the videos have easy to recognize objects and actions, and are unique enough to not cause much ambiguity with matching. }
  \label{fig:matching.ranking.top3.results}
  
\end{figure*}

\begin{figure*}[!htb]

  \subfloat[Video \#1704]{\includegraphics[width=0.3\linewidth]{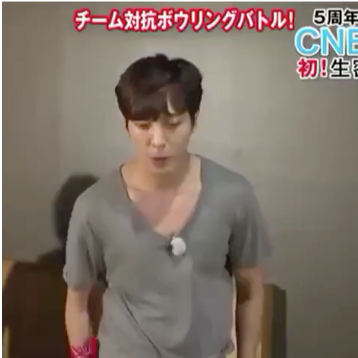}} \hfill
  \subfloat[Video \#1822]{\includegraphics[width=0.3\linewidth]{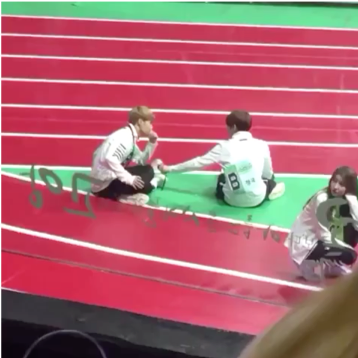}} \hfill
  \subfloat[Video \#205]{\includegraphics[width=0.3\linewidth]{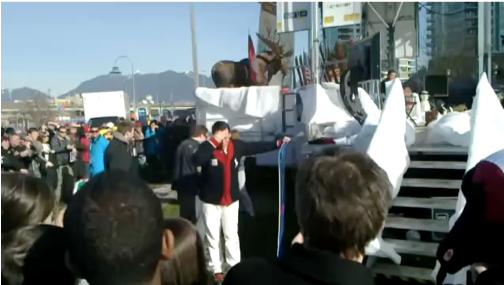}}

  \caption{VTT: The bottom 3 videos for the matching and ranking subtask with their video IDs. These videos are not necessarily hard to describe, but many of the objects and actions may also be common with other videos.}
  \label{fig:matching.ranking.bottom3.results}
\end{figure*}

\subsubsection{Matching and Ranking}

All runs submitted to the matching and ranking subtask were of type `V'. 

The results for the subtask are shown for each of the 5 sets (A-E) in Figure~\ref{fig:vtt.match.rank.all.sets}. The graph shows the mean inverted rank scores for all runs submitted by the teams for each of the description sets. RUC\_AIM3 outperformed the other systems in this subtask. 

The maximum mean inverted rank score increased from 0.516 in 2018 to 0.727 in 2019. 

Figure~\ref{fig:matching.ranking.top3.results} shows the top 3 videos for this subtask. These videos are matched correctly in a consistent manner among runs. 
Figure~\ref{fig:matching.ranking.bottom3.results} shows 3 videos that systems were generally unable to match with the correct descriptions. 

\subsection{Observations and Conclusion}

The VTT task continues to have healthy participation. Given the challenging nature of the task, and the increasing interest in video captioning in the computer vision community, we hope to see improvements in performance. 

This year we used two video sources in the testing dataset, Flickr and Vines. However, we plan to change the dataset for the coming years. With increasing interest in video captioning, participants have a number of open datasets available to train their systems.

We observed an increase in scores for all metrics from 2018 to 2019 for the description generation subtask. The mean inverted rank score for matching and ranking also increased this year. While it may not be a fair comparison due to different datasets, this year's testing dataset collection process was similar to the last year. We, therefore, believe that the score increase, at least partially, may be due to system improvements.

Systems were divided into three run types based on how they were trained. However, given that most runs were of the same type, this information did not provide us much insight (PicSOM attempted to compare between these different run types). For the next year's task we will deliberate on what information could be helpful with useful analysis. 

\section{Summing up and moving on}

This overview to TRECVID 2019 has provided basic information on the goals, data, evaluation mechanisms, and metrics used. 
Further details about each particular group's approach and performance for each task can be found in that group's site report. The raw results
for each submitted run can be found at the online proceeding of the workshop \cite{tv19pubs}.

\section{Authors' note}
TRECVID would not have happened in 2019 without support from the
National Institute of Standards and Technology (NIST). The research
community is very grateful for this. Beyond that, various individuals
and groups deserve special thanks:
\begin{itemize}

\item{Koichi Shinoda of the TokyoTech team agreed to host a copy of 
IACC.2 data.}

\item{Georges Qu\'{e}not provided the master shot reference for the
IACC.3 videos.}

\item{The LIMSI Spoken Language Processing Group and Vocapia Research
provided ASR for the IACC.3 videos.}

\item{Luca Rossetto of University of Basel for providing the V3C dataset collection.}

\item{Noel O'Connor and Kevin McGuinness at Dublin City University
  along with Robin Aly at the University of Twente worked with NIST
  and Andy O'Dwyer plus William Hayes at the BBC to make the BBC
  EastEnders video available for use in TRECVID. Finally, Rob Cooper at BBC
  facilitated the copyright licence agreement for the Eastenders data.}
 
\end{itemize}

Finally we want to thank all the participants and other contributors
on the mailing list for their energy and perseverance.

\section{Acknowledgments} 
The ActEV NIST work was supported by the Intelligence Advanced Research Projects Activity (IARPA), agreement~IARPA-16002, order R18-774-0017. The authors would like to thank Kitware, Inc. for annotating the dataset. 
The Video-to-Text work has been partially supported by Science Foundation Ireland (SFI) as a part of the Insight Centre at DCU (12/RC/2289).
We would like to thank Tim Finin and Lushan Han of University of Maryland, Baltimore County for providing access to the semantic similarity metric.



\bibliography{video}

\clearpage
\onecolumn

\appendix
\section{Ad-hoc query topics} 
\label{appendixA}
\begin{description}\itemsep0pt \parskip0pt

\item[611] Find shots of a drone flying
\item[612] Find shots of a truck being driven in the daytime
\item[613] Find shots of a door being opened by someone
\item[614] Find shots of a woman riding or holding a bike outdoors
\item[615] Find shots of a person smoking a cigarette outdoors
\item[616] Find shots of a woman wearing a red dress outside in the daytime
\item[617] Find shots of one or more picnic tables outdoors
\item[618] Find shots of coral reef underwater
\item[619] Find shots of one or more art pieces on a wall
\item[620] Find shots of a person with a painted face or mask
\item[621] Find shots of person in front of a graffiti painted on a wall
\item[622] Find shots of a person in a tent
\item[623] Find shots of a person wearing shorts outdoors
\item[624] Find shots of a person in front of a curtain indoors
\item[625] Find shots of a person wearing a backpack
\item[626] Find shots of a race car driver racing a car
\item[627] Find shots of a person holding a tool and cutting something
\item[628] Find shots of a man and a woman holding hands
\item[629] Find shots of a black man singing
\item[630] Find shots of a man and a woman hugging each other
\item[631] Find shots of a man and a woman dancing together indoors
\item[632] Find shots of a person running in the woods
\item[633] Find shots of a group of people walking on the beach
\item[634] Find shots of a woman and a little boy both visible during daytime
\item[635] Find shots of a bald man
\item[636] Find shots of a man and a baby both visible
\item[637] Find shots of a shirtless man standing up or walking outdoors
\item[638] Find shots of one or more birds in a tree
\item[639] Find shots for inside views of a small airplane flying
\item[640] Find shots of a red hat or cap

\end{description}
    
\section{Instance search topics - 30 unique} 
\label{appendixB}
\begin{description} \itemsep0pt

\item[9249] Find Max Holding a glass
\item[9250] Find Ian Holding a glass
\item[9251] Find Pat Holding a glass
\item[9252] Find Denise Holding a glass
\item[9253] Find Pat Sitting on a couch
\item[9254] Find Denise Sitting on a couch
\item[9255] Find Ian Holding phone
\item[9256] Find Phil Holding phone
\item[9257] Find Jane Holding phone
\item[9258] Find Pat Drinking
\item[9259] Find Ian Opening door and entering room / building
\item[9260] Find Dot Opening door and entering room / building
\item[9261] Find Max Shouting
\item[9262] Find Phil Shouting
\item[9263] Find Ian Eating
\item[9264] Find Dot Eating
\item[9265] Find Max Crying
\item[9266] Find Jane Laughing
\item[9267] Find Dot Opening door and leaving room / building
\item[9268] Find Phil Going up or down stairs
\item[9269] Find Jack Sitting on a couch
\item[9270] Find Stacey Carrying a bag
\item[9271] Find Bradley Carrying a bag
\item[9272] Find Stacey Drinking
\item[9273] Find Jack Drinking
\item[9274] Find Jack Shouting
\item[9275] Find Stacey Crying
\item[9276] Find Bradley Laughing
\item[9277] Find Jack Opening door and leaving room / building
\item[9278] Find Stacey Going up or down stairs

\section*{Instance search topics - 20 common}
\item[9279] Find Phil Sitting on a couch
\item[9280] Find Heather Sitting on a couch
\item[9281] Find Jack Holding phone
\item[9282] Find Heather Holding phone
\item[9283] Find Phil Drinking
\item[9284] Find Shirley Drinking
\item[9285] Find Jack Kissing
\item[9286] Find Denise Kissing
\item[9287] Find Phil Opening door and entering room / building
\item[9288] Find Sean Opening door and entering room / building
\item[9289] Find Shirley Shouting
\item[9290] Find Sean Shouting
\item[9291] Find Stacey Hugging
\item[9292] Find Denise Hugging
\item[9293] Find Max Opening door and leaving room / building
\item[9294] Find Stacey Opening door and leaving room / building
\item[9295] Find Max Standing and talking at door
\item[9296] Find Dot Standing and talking at door
\item[9297] Find Jack Closing door without leaving
\item[9298] Find Dot Closing door without leaving
    
\end{description}

\end{document}